\documentclass[twoside,11pt]{article}

\usepackage{jmlr2e}

\newtheorem{assumption}[theorem]{Assumption}
\usepackage{thm-restate}

\usepackage{amsmath,amsfonts,amssymb,stackengine,comment}
\usepackage{color,xcolor}
\usepackage{algorithm,algpseudocode} 
\usepackage{enumerate}
\usepackage{caption,subcaption,float,booktabs,multirow,lscape}
\usepackage{bbding}
\RequirePackage{caption,subcaption,booktabs,multirow,array}

\def\pdim{{\rm Pdim}}
\def\vdim{{\rm VCdim}}
\def\s{{\rm sign}}

\def\pj{\frac{\partial}{\partial x_j}}
\def\pdim{{\rm Pdim}}
\def\vdim{{\rm VCdim}}
\def\s{{\rm sign}}

\usepackage{lastpage}
\ShortHeadings{Differentiable Neural Networks with RePU Activation}{Shen,  Jiao, Lin, and Huang}

\begin{document}
\title{Differentiable Neural Networks with R\lowercase{e}PU Activation: with Applications to
Score Estimation and Isotonic Regression}

\author{\name Guohao Shen
\email guohao.shen@polyu.edu.hk \\
\addr Department of Applied Mathematics\\ The Hong Kong Polytechnic University\\
 Hong Kong SAR, China
\AND
\name Yuling Jiao
\email yulingjiaomath@whu.edu.cn \\
\addr School of Mathematics and Statistics \\ and Hubei Key Laboratory of Computational Science \\ Wuhan University\\
 Wuhan, China, 430072
\AND
\name Yuanyuan Lin
\email ylin@sta.cuhk.edu.hk\\
\addr Department of Statistics \\
The Chinese University of Hong Kong\\
Hong Kong SAR, China
\AND
\name Jian Huang
\email j.huang@polyu.edu.hk \\
\addr Department of Applied Mathematics\\
The Hong Kong Polytechnic University\\
Hong Kong SAR, China
}

\maketitle

\begin{abstract}
We study the properties of differentiable neural networks activated by rectified power unit (RePU) functions. We show that the partial derivatives of RePU neural networks can be represented by RePUs mixed-activated networks and derive upper bounds for the complexity of the function class of derivatives of RePUs networks. We establish error bounds for simultaneously approximating $C^s$ smooth functions and their derivatives using RePU-activated deep neural networks. Furthermore, we derive improved approximation error bounds when data has an approximate low-dimensional support, demonstrating the ability of RePU networks to mitigate the curse of dimensionality. To illustrate the usefulness of our results, we consider a deep score matching estimator (DSME) and propose a penalized deep isotonic regression (PDIR) using RePU networks. We establish non-asymptotic excess risk bounds for DSME and PDIR under the assumption that the target functions belong to a class of $C^s$ smooth functions. We also show that PDIR achieves the minimax optimal convergence rate and has a robustness property in the sense it is consistent with vanishing penalty parameters even when the monotonicity assumption is not satisfied. Furthermore, if the data distribution is supported on an approximate low-dimensional manifold, we show that DSME and PDIR can mitigate the curse of dimensionality.
\end{abstract}


\section{Introduction}
 In many statistical
problems, it is important to estimate the derivatives of a target function, in addition to estimating a target function itself. An  example is  the score matching method for distribution learning through
score function estimation \citep{hyvarinen2005estimation}. In this method, the objective function involves the partial derivatives of the score function. Another example is  a newly proposed penalized approach for isotonic regression described below, in which the partial derivatives are used to form a penalty function to encourage the estimated regression function to be monotonic.
Motivated by these problems, we consider  Rectified Power Unit (RePU) activated
deep neural networks for estimating differentiable target functions.
A RePU activation function
has continuous derivatives, which makes RePU networks differentiable and suitable for derivative estimation. We study the properties of RePU networks along with their derivatives, establish error bounds for using RePU networks to approximate smooth functions and their derivatives, and apply them to the problems of score estimation and isotonic regression.

\subsection{Score matching}
Score estimation is an important approach to distribution learning and
score function plays a central role in the diffusion-based generative learning  \citep{song2021scorebased, block2020generative,ho2020denoising,lee2022convergence}. 
Let $X \sim p_0$, where $p_0$ is a probability density function supported on $\mathbb{R}^d$, then $d$-dimensional score function of $p_0$ is defined as $s_0(x)=\nabla_x \log p_0(x)$, where $\nabla_x$ is the vector differential operator with respect to the input $x$.

A score matching estimator  \citep{hyvarinen2005estimation} is obtained by solving the minimization problem
  \begin{equation}
  \label{Js0}
  \min_{s \in \mathcal{F}} \frac{1}{2}  \mathbb{E}_{X}\Vert s(X)-s_0(X)\Vert^2_2,
  \end{equation}
where $\Vert \cdot \Vert_2$ denotes the Euclidean norm, $\mathcal{F}$ is a prespecified class of functions, often referred to as a hypothesis space.
However, this objective function is computationally infeasible because $s_0$ is unknown. Under some mild conditions given in Assumption \ref{assump2} in Section \ref{sec_dse}, it can be shown that \citep{hyvarinen2005estimation}
\begin{equation}\label{sm1}
  \frac{1}{2} \mathbb{E}_{X}\Vert s(X)-s_0(X)\Vert^2_2
	=J(s) +\frac{1}{2}  \mathbb{E}_X \Vert s_0(X)\Vert_2^2,
\end{equation}
with
\begin{equation}
\label{obj0}
J(s) :=\mathbb{E}_X\left[{\rm tr}(\nabla_x s(X))+\frac{1}{2} \Vert s(X)\Vert_2^2\right],
\end{equation}
where
$\nabla_x s(x)$ denotes the Jacobian matrix of $s(x)$ and ${\rm tr}(\cdot)$  the trace operator.
Since the second term on the right side of (\ref{sm1}), $\mathbb{E}_X \Vert s_0(X)\Vert_2^2/2$, does not involve $s$, it can be considered a constant.
Therefore, we can just use  $J$ in (\ref{obj0}) as an objective function for estimating the score function $s_0$. When a random sample is available,  we use a sample version of $J$ as the empirical objective function. Since $J$ involves the partial derivatives of $s(x)$, we need to compute the derivatives of the functions in  $\mathcal{F}$ during estimation. And we need to analyze the properties of $\mathcal{F}$ and their derivatives to develop the learning theories. In particular, if we take $\mathcal{F}$ to be a class of deep neural network functions, we need to study the properties of their derivatives in terms of estimation and approximation.

\subsection{Isotonic regression}
Isotonic regression is a technique that fits a regression model to observations such that the fitted
regression function is non-decreasing (or non-increasing).
It is a basic form of shape-constrained estimation
and has applications in many areas, such as  epidemiology \citep{morton2000additive}, medicine \citep{diggle1999case,jiang2011smooth}, econometrics \citep{horowitz2017nonparametric},  and biostatistics \citep{rueda2009estimation,luss2012efficient,qin2014combining}.

Consider a regression model
\begin{equation}\label{reg0}
	Y=f_0(X)+\epsilon,
\end{equation}
where  $X \in \mathcal{X}\subseteq\mathbb{R}^d,$ $Y\in \mathbb{R},$ and $\epsilon$ is an  independent noise variable
with $\mathbb{E}(\epsilon)=0$ and ${\rm Var}(\epsilon)\le\sigma^2.$
In (\ref{reg0}), $f_0$ is the underlying regression function, which is usually assumed to
belong to certain smooth function class.

In isotonic regression,  $f_0$ is assumed to satisfy a monotonicity property as follows.
Let $\preceq$ denote the binary relation ``less than" in
the partially ordered space
$\mathbb{R}^d$, i.e., $u\preceq v$ if $u_j\le v_j$ for all $j=1,\ldots,d,$ where $u=(u_1,\ldots,u_d)^\top, v=(v_1,\ldots,v_d)^\top\in\mathcal{X}\subseteq\mathbb{R}^d$.
In isotonic regression, the target regression function  $f_0$ is assumed to be coordinate-wisely non-decreasing on $\mathcal{X}$, i.e., $f_0(u)\le f_0(v)$ if $u \preceq v.$
The class of isotonic regression functions on $\mathcal{X}\subseteq\mathbb{R}^d$ is the set of coordinate-wisely non-decreasing functions
\[
\mathcal{F}_0:=\{f:\mathcal{X}\to\mathbb{R}, f(u)\le f(v) {\rm\ if \ } u\preceq v, \text{ where } \mathcal{X} \subset \mathbb{R}^d\}.
\]
The goal is to estimate the
 target regression function $f_0$ under the constraint that $f_0\in\mathcal{F}_0$ based on an observed sample $S:=\{(X_i,Y_i)\}_{i=1}^n$. For a possibly random function $f:\mathcal{X}\to\mathbb{R}$, let
the population risk be
\begin{equation}\label{risk}
	\mathcal{R}(f)=\mathbb{E}\vert Y-f(X)\vert^2,
\end{equation}
where $(X,Y)$ follows the same distribution as $(X_i,Y_i)$ and is independent of $f$.
Then the target function $f_0$ is the minimizer of the risk $\mathcal{R}(f)$ over $\mathcal{F}_0,$  i.e., \begin{equation}
\label{f0}
f_0\in\arg\min_{f \in \mathcal{F}_0} \mathcal{R}(f).
\end{equation}
The empirical version of (\ref{f0}) is a constrained minimization problem, which is generally difficult to solve directly. In light of this, we propose a penalized approach for estimating $f_0$ based on the fact that, for a smooth $f_0$, it is increasing with respect to the $j$th argument $x_j$ if and only if its partial derivative with respect to $x_j$ is nonnegative.
Let $\dot{f}_j(x) = \partial f(x)/\partial x_j $ denote the partial derivative of $f$ with respective to $x_j, j=1,\ldots, d.$
We propose the following penalized objective function
\begin{equation}
\label{iro1}
\mathcal{R}^\lambda(f)=\mathbb{E}\vert Y-f(X)\vert^2+\frac{1}{d}\sum_{j=1}^d \lambda_j \mathbb{E}\{\rho(\dot{f}_j(X))\},
\end{equation}
where $\lambda=\{\lambda_j\}_{j=1}^d$ with $\lambda_j\ge0, j=1,\ldots,d$, are tuning parameters,
$\rho(\cdot):\mathbb{R}\to[0,\infty)$ is a penalty function satisfying $\rho(x)\ge0$ for all $x$ and $\rho(x)=0$ if $x\ge0$. Feasible choices include $\rho(x)=\max\{-x,0\}$, $\rho(x)=[\max\{-x,0\}]^2$ and more generally $\rho(x)=h(\max\{-x,0\})$ for a Lipschitz function $h$ with $h(0)=0$.

The objective function (\ref{iro1}) turns the constrained isotonic regression problem into a penalized
regression problem with penalties on the partial derivatives of the regression function. Therefore, if we analyze the learning theory of estimators in
(\ref{iro1}) using neural network functions, we need to study the partial derivatives of the neural network functions in terms of their generalization and
approximation properties.

It is worth mentioning that an advantage of our penalized formulation over the hard-constrained isotonic regressions is that the resulting estimator remains consistent with proper tuning when the underlying regression function is not monotonic. Therefore, our proposed method has a robustness property against model misspecification. We will discuss this point in detail in Section \ref{misspec}.

\subsection{Differentiable neural networks}
{\color{black}
A commonality between the aforementioned two quite different problems is that they both
involve the derivatives of the target function, in addition to the target function itself. When deep neural networks are used to parameterize the hypothesis space,
the derivatives of deep neural networks must be considered. To study the statistical learning theory for these deep neural methods, it
requires the knowledge of the complexity and approximation properties of deep neural networks along with their derivatives.

Complexities of deep neural networks with ReLU and piece-wise polynomial activation functions have been studied by \cite{anthony1999} and \cite{bartlett2019nearly}. Generalization bounds in terms of the operator norm of neural networks have also been obtained by several authors \citep{neyshabur2015norm,NIPS2017_b22b257a,nagarajan2019deterministic,wei2019data}.  These generalization results are based on various complexity measures such as Rademacher complexity, VC-dimension, Pseudo-dimension, and norm of parameters.  These studies
shed light on the complexity and generalization properties of neural networks themselves, however,  the complexities of their derivatives remain unclear.

The approximation power of deep neural networks with smooth activation functions has been considered in the literature. The universality of sigmoidal deep neural networks has been established
by \cite{mhaskar1993approximation} and \cite{chui1994neural}.  In addition, the approximation properties of shallow RePU-activated networks were analyzed
by \cite{klusowski2018approximation} and \cite{siegel2022high}. The approximation rates of deep RePU neural networks
for several types of
target functions have also been investigated.
For instance,  \cite{li2019better,li2020powernet}, and \cite{ ali2021approximation} studied the approximation rates for functions in Sobolev and Besov spaces in terms of the $L_p$ norm,
\citet{duan2021convergence}, \cite{abdeljawad2022approximations} studied the approximation rates for functions in Sobolev space in terms of the Sobolev norm, and \citet{belomestny2022simultaneous} studied the approximation rates for functions in H\"older space in terms of the H\"older norm.  
Several recent papers have also studied the approximation of derivatives of smooth functions \citep{duan2021convergence,guhring2021approximation,belomestny2022simultaneous}.
We will have more detailed discussions on the related works in Section \ref{related}.

Table \ref{tab} provides a summary of the comparison between our work and the existing results for achieving the same approximation accuracy $\epsilon$ on a function with smoothness index $\beta$ in terms of the needed non-zero parameters in the network. We also summarize the results on whether the neural network approximator has an explicit architecture, where the approximation accuracy holds simultaneously for the target function and its derivative, and whether the approximation results were shown to adapt to the low-dimensional structure of the target function.

\begin{table}[H]
\caption{\label{tab}
Comparison of approximation results of RePU neural networks on a function with smoothness order $\beta>0$, within the accuracy $\epsilon$. ReQU $\sigma_{2}$ and ReCU $\sigma_{3}$ are special instances of RePU $\sigma_p$ for $p\ge2$. The Sobolev norm $W^{s,p}$ of a function $f$ refers to the mean value of $L_p$ norm of all partial derivatives of $f$ up to order $s$, and $W^{s,\infty}$  refers to the maximum value of $L_\infty$ norm of all the partial derivatives of $f$ up to order $s$. The H\"older norm $\mathcal{H}^s$ refers to $W^{s,\infty}$ when $s$ is a non-negative integer. The $C^s$ norm of a function $f$ refers to the mean value of $L_\infty$ norm of all partial derivatives of $f$ up to order $s$ when $s$ is a positive integer.}
	\resizebox{1.0\textwidth}{!}{%
		\begin{tabular}{l | >{\centering}p{1cm} >{\centering}p{1.2cm} >{\centering}p{1.7cm} >{\centering}p{1.7cm} >{\centering}p{1.9cm}  >{\centering\arraybackslash} p{1.2cm} }
			\toprule
			 & Norm   & Activation     & Non-zero parameters     & Explicit architecture   &Simultaneous approximation & Low-dim result     \\ \midrule
\cite{li2019better,li2020powernet}      & $L_2$&  RePU & $\mathcal{O}(\epsilon^{-d/\beta})~~~$ & \XSolidBrush & \XSolidBrush &  \XSolidBrush \\
	\cite{duan2021convergence}     & $W^{1,2}$ &  ReQU & $\mathcal{O}(\epsilon^{-2d/(\beta-1)})$  & \XSolidBrush & \Checkmark  &  \XSolidBrush \\
\cite{abdeljawad2022approximations}      & $W^{s,p}$ & ReCU & $\mathcal{O}(\epsilon^{-d/(\beta-s)})$ & \XSolidBrush   & \Checkmark &  \XSolidBrush  \\
\cite{belomestny2022simultaneous}      & $\mathcal{H}^s$ & ReQU & $\mathcal{O}(\epsilon^{-d/(\beta-s)})$ & \Checkmark   & \Checkmark &  \XSolidBrush \\
			This work      & $C^s$ &  RePU & $\mathcal{O}(\epsilon^{-d/(\beta-s)})$ & \Checkmark   & \Checkmark  &  \Checkmark  \\ \bottomrule
		\end{tabular}%
}
\end{table}

\subsection{Our contributions}
In this paper, motivated by the aforementioned estimation problems involving derivatives, we investigate the properties of RePU networks and their derivatives. We show that the partial derivatives of RePU neural networks can be represented by mixed-RePUs activated networks. We derive upper bounds for the complexity of the function class of the derivatives of RePU networks. This is a new result for the complexity of derivatives of RePU networks and is crucial to establish generalization error bounds for a variety of estimation problems involving derivatives, including the score matching estimation and our proposed penalized approach for isotonic regression considered in the present work.

We also derive our approximation results of the RePU network  on the smooth functions and their derivatives simultaneously. Our approximation results of the RePU network are based on its representational power on polynomials.
We construct the RePU networks with an explicit architecture, which is different from those in the existing literature. The number of hidden layers of our constructed RePU networks only depends on the degree of the target polynomial but is independent of the dimension of input.  This construction is new for studying the approximation properties of RePU networks.
}

We summarize the main contributions of this work as follows.

\begin{enumerate}

\item
We study the basic properties of RePU neural networks and their derivatives. First, we show that partial derivatives of RePU networks can be represented by mixed-RePUs activated networks.  We derive upper bounds for the complexity of the function class of partial derivatives of RePU networks in terms of pseudo dimension.
{\color{black} Second, we derive novel approximation results for simultaneously approximating $C^s$ smooth functions and their derivatives using RePU networks based on a new and efficient construction technique.} We show that the approximation can be improved when the data or target function has a low-dimensional structure, which implies that RePU networks can mitigate the curse of dimensionality.

\item We study the statistical learning theory of deep score matching
estimator (DSME) using RePU networks. We establish non-asymptotic prediction error bounds for DSME under the assumption that the target score is $C^s$ smooth. We show that DSME can mitigate the curse of dimensionality if the data has low-dimensional support.

\item We propose a penalized deep isotonic regression (PDIR) approach using RePU networks, which encourages the partial derivatives of the estimated regression function to be nonnegative. We establish non-asymptotic excess risk bounds for PDIR under the assumption that the target regression function $f_0$ is $C^s$ smooth. {\color{black} Moreover, we show that PDIR achieves the minimax optimal rate of convergence for non-parametric regression.} We also show that PDIR can mitigate the curse of dimensionality when data concentrates near a low-dimensional manifold.  Furthermore, we show that with tuning parameters tending to zero,  PDIR is consistent even when the target function is not isotonic.

	
	
\end{enumerate}

{\color{black}
The rest of the paper is organized as follows. 
In Section \ref{sec_basic} we
study basic properties of RePU neural networks.   In Section \ref{sec_approx} we establish novel approximation error bounds for approximating $C^s$ smooth functions and their derivatives using RePU networks.
In Section \ref{sec_dse}  we derive non-asymptotic error bounds for DSME.
In Section \ref{sec_dir} we propose
PDIR and establish non-asymptotic bounds for PDIR. In Section \ref{related} we discuss related works.
Concluding remarks are given in Section \ref{conclusion}. Results from simulation studies, proofs, and technical details are given in the Supplementary Material.
}

\section{Basic properties of RePU neural networks} \label{sec_basic}

In this section, we establish some basic properties of RePU networks.
We show that the partial derivatives of RePU networks can be
represented by RePUs mixed-activated networks. The width, depth, number of neurons, and size of the RePUs mixed-activated network have the same order as those of the original RePU networks. In addition, we derive upper bounds of the complexity of the function class of RePUs mixed-activated networks in terms of pseudo dimension, which leads to an upper bound of the class of partial derivatives of RePU networks.


\subsection{RePU activated neural networks}
Neural networks with nonlinear activation functions have proven to be a powerful approach for
approximating multi-dimensional functions. One of the most commonly used activation functions is the Rectified linear unit (ReLU), defined as $\sigma_1(x)=\max\{x,0\}, x \in \mathbb{R}$, due to its attractive properties in computation and optimization.
However, since partial derivatives are involved in our objective function (\ref{obj0}) and (\ref{iro1}),
it is not sensible to use networks with piecewise linear activation functions, such as ReLU and Leaky ReLU.
Neural networks activated by Sigmoid and Tanh, are smooth and differentiable but have been falling from favor due to their vanishing gradient problems in optimization. In light of these, we are particularly interested in studying the neural networks activated by RePU, which are non-saturated and differentiable. 

{\color{black}
	In Table \ref{tab:0} below we compare RePU with ReLU and Sigmoid networks in several important aspects. ReLU and RePU activation functions are continuous and non-saturated\footnote{An activation function $\sigma$ is saturating if $\lim_{\vert x\vert\to\infty}\vert\nabla \sigma(x)\vert=0$.}, which do not have ``vanishing gradients" as Sigmodal activations (e.g. Sigmoid, Tanh) in training. RePU and Sigmoid are differentiable and can approximate the gradient of a target function, but ReLU activation is not, especially for estimation involving high-order derivatives of a target function.
}

\begin{table}[H]
	\centering
	\caption{Comparison among ReLU, Sigmoid and RePU activation functions. }
	\label{tab:0}
	\resizebox{0.9\textwidth}{!}{%
		\begin{tabular}{c|cccc}
			\toprule
			Activation & Continuous        & Non-saturated             & Differentiable              & Gradient Estimation         \\ \midrule
			ReLU       & \Checkmark & \Checkmark & \XSolidBrush & \XSolidBrush \\
			Sigmoid   & \Checkmark &  \XSolidBrush  & \Checkmark & \Checkmark\\
			RePU       & \Checkmark & \Checkmark & \Checkmark   & \Checkmark   \\ \bottomrule
		\end{tabular}%
	}
\end{table}

We consider the
$p$th order
Rectified Power units (RePU) activation function for a positive integer $p.$
The RePU activation function, denoted as $\sigma_p$,  is simply the power of ReLU,
\begin{align*}
	\sigma_p(x)=\left\{\begin{array}{lr}
		x^p,& x\ge0\\
		0, & x<0
	\end{array}\right..
\end{align*}
Note that when $p=0$, the activation function $\sigma_0$ is the Heaviside step function; when $p=1$, the activation function $\sigma_1$ is the familiar Rectified Linear unit (ReLU); when $p=2,3$, the activation functions $\sigma_{2}, \sigma_{3}$ are called rectified quadratic unit (ReQU) and rectified cubic unit (ReCU) respectively. In this work, we focus on the case with $ p \ge 2$, implying that the RePU activation function has a continuous $(p-1)$th continuous derivative.

With a RePU activation function, the network will be smooth and differentiable.
The architecture of a RePU activated multilayer perceptron can be expressed as a composition of a series of functions
\[
f(x)=\mathcal{L}_\mathcal{D}\circ\sigma_p\circ\mathcal{L}_{\mathcal{D}-1}
\circ\sigma_p\circ\cdots\circ\sigma_p\circ\mathcal{L}_{1}\circ\sigma_p\circ\mathcal{L}_0(x),\  x\in \mathbb{R}^{d_0},
\]
where $d_0$ is the dimension of the input data,  $\sigma_p(x)=\{\max(0, x)\}^p, p \ge 2,$ is a RePU activation function (defined for each component of $x$ if $x$ is a vector), and
$\mathcal{L}_{i}(x)=W_ix+b_i, i=0,1,\ldots,\mathcal{D}.$
Here $d_i$ is the width (the number of neurons or computational units) of the $i$-th layer, $W_i\in\mathbb{R}^{d_{i+1}\times d_i}$ is a weight matrix, and $b_i\in\mathbb{R}^{d_{i+1}}$ is the bias vector in the $i$-th linear transformation $\mathcal{L}_i$.
The input data $x$ is the first layer of the neural network and the output
is the last layer. Such a network $f$ has $\mathcal{D}$ hidden layers and $(\mathcal{D}+2)$ layers in total. We use a $(\mathcal{D}+2)$-vector $(d_0,d_1,\ldots,d_\mathcal{D},d_{\mathcal{D}+1})^\top$ to describe the width of each layer. 
The width $\mathcal{W}$ is defined as the maximum width of hidden layers, i.e.,
$\mathcal{W}=\max\{d_1,...,d_\mathcal{D}\}$; 	the size $\mathcal{S}$ is defined as the total number of parameters in the network $f$, i.e., $\mathcal{S}=\sum_{i=0}^\mathcal{D}\{d_{i+1}\times(d_i+1)\}$; 	
the number of neurons $\mathcal{U}$ is defined as the number of computational units in hidden layers, i.e., $\mathcal{U}=\sum_{i=1}^\mathcal{D} d_i$.  Note that the neurons in consecutive layers are connected to each other via weight matrices $W_i$,  $i=0,1,\ldots,\mathcal{D}$.

We use the notation $\mathcal{F}_{\mathcal{D},\mathcal{W}, \mathcal{U},\mathcal{S},\mathcal{B},\mathcal{B}^\prime}$ to denote a class of RePU activated multilayer perceptrons $f: \mathbb{R}^{d_0} \to \mathbb{R} $ with depth $\mathcal{D}$, width $\mathcal{W}$, number of  neurons $\mathcal{U}$, size $\mathcal{S}$ and $f$ satisfying $\Vert f \Vert_\infty\leq\mathcal{B}$ and $\max_{j=1,\ldots,d}\Vert \pj f \Vert_\infty\leq\mathcal{B}^\prime$ for some $0 <\mathcal{B}, \mathcal{B}^\prime< \infty$, where
$\Vert f \Vert_\infty$ is the sup-norm of a function $f$.

\subsection{Derivatives of RePU networks}
An advantage of RePU networks over piece-wise linear activated networks (e.g. ReLU networks) is that RePU networks are differentiable. Thus RePU networks are useful in many estimation problems involving derivative. To establish the learning theory for these problems, we need to study the properties of derivatives of RePU.

{\color{black}
Recall that a $\mathcal{D}$-hidden layer network activated by $p$th order RePU can be expressed by
$$f(x)=\mathcal{L}_\mathcal{D}\circ\sigma_p\circ\mathcal{L}_{\mathcal{D}-1}
\circ\sigma_p\circ\cdots\circ\sigma_p\circ\mathcal{L}_{1}\circ\sigma_p\circ\mathcal{L}_0(x),\  x\in \mathbb{R}^{d_0}.$$
Let $f_i:=\sigma_p\circ\mathcal{L}_i$ denote the $i$th linear transformation composited with RePU activation for $i=0,1,\ldots,\mathcal{D}-1$ and let $f_\mathcal{D}=\mathcal{L}_\mathcal{D}$ denotes the linear transformation in the last layer.
Then by the chain rule, the gradient of the network can be computed by
\begin{equation}\label{chainrule}
	\nabla f=\left(\prod_{k=0}^{\mathcal{D}-1}\big[\nabla f_{\mathcal{D}-k}\circ f_{\mathcal{D}-k-1}\circ\ldots\circ f_0\big]\right)\nabla f_0,
\end{equation}
where $\nabla$ denotes the gradient operator used in vector calculus.
With a differentiable RePU activation $\sigma_p$, the gradients $\nabla f_i$ in (\ref{chainrule}) can be exactly computed by $\sigma_{p-1}$ activated layers since $\nabla f_i(x)=\nabla[\sigma_p\circ\mathcal{L}_i(x)]=\nabla \sigma_p(W_ix+b_i)=pW_i^\top\sigma_{p-1}(W_ix+b_i)$. In addition, the $f_i, i=0,\ldots,\mathcal{D}$ are already RePU-activated layers.
Then, the network gradient $\nabla f$ can be represented by a network activated by $\sigma_p,\sigma_{p-1}$ (and possibly $\sigma_m$ for $1\le m\le p-2$) according to (\ref{chainrule}) with a proper architecture.
Below, we refer to the neural networks activated by the $\{\sigma_t:1\le t\le p\}$ as Mixed RePUs activated neural networks, i.e., the activation functions in Mixed RePUs network can be $\sigma_t$ for $1\le t\le p$, and for different neurons the activation function can be different.

The following theorem shows that the partial derivatives and the gradient $\nabla f_\phi$ of a RePU neural network indeed can be represented by a Mixed RePUs network with activation functions $\{\sigma_t:1\le t\le p\}$.
}

\begin{theorem}[Neural networks for partial derivatives]\label{thm2}
Let $\mathcal{F}:=\mathcal{F}_{\mathcal{D},\mathcal{W}, \mathcal{U},\mathcal{S},\mathcal{B},\mathcal{B}^\prime}$ be a class of RePU $\sigma_p$ activated neural networks $f:\mathcal{X}\to\mathbb{R}$ with depth (number of hidden layer) $\mathcal{D}$, width (maximum width of hidden layer) $\mathcal{W}$, number of neurons $\mathcal{U}$, number of parameters (weights and bias) $\mathcal{S}$ and $f$ satisfying $\Vert f \Vert_\infty\leq\mathcal{B}$ and $\max_{j=1,\ldots,d}\Vert \pj f \Vert_\infty\leq\mathcal{B}^\prime$. Then for any $f\in\mathcal{F}$ and any $j\in\{1,\ldots,d\}$, the partial derivative $\pj f$ can be implemented by a Mixed RePUs activated multilayer perceptron with depth $3\mathcal{D}+3$, width $6\mathcal{W}$, number of neurons $13\mathcal{U}$, number of parameters $23\mathcal{S}$ and bound $\mathcal{B}^\prime$.
\end{theorem}

Theorem \ref{thm2} shows that for each $j\in\{1,\ldots,d\}$, the partial derivative with respect to the $j$-th argument of the function $f\in\mathcal{F}$ can be exactly computed by a Mixed RePUs network. In addition, by paralleling the networks computing $\pj f,j=1,\ldots,d$, the whole vector of partial derivatives $\nabla f=(\frac{\partial}{\partial x_1}f,\ldots,\frac{\partial}{\partial x_d}f)$ can be computed by a Mixed RePUs network with depth $3\mathcal{D}+3$, width $6d\mathcal{W}$, number of neurons $13d\mathcal{U}$ and number of parameters $23d\mathcal{S}$.

 Let $\mathcal{F}_{j}^\prime:=\{{\partial}/{\partial x_j} f:f\in\mathcal{F}\}$
be the partial derivatives of the functions in $\mathcal{F}$ with respect to the $j$-th argument. And let $\mathcal{\widetilde{F}}$ denote the class of Mixed RePUs networks in Theorem \ref{thm2}. Then Theorem \ref{thm2} implies that the class of partial derivative functions is contained in a class of Mixed RePUs networks, i.e., $\mathcal{F}^\prime_j\subseteq \mathcal{\widetilde{F}}$ for $j=1,\ldots,d$. This further implies the complexity of $\mathcal{F}_{j}^\prime$ can be bounded by that of the class of Mixed RePUs networks $\mathcal{\widetilde{F}}$.  

The complexity of a function class is a key quantity in the analysis of generalization properties. Lower complexity in general implies a smaller generalization gap. The complexity of a function class can be measured in several ways, including Rademacher complexity, covering number, VC dimension, and Pseudo dimension. These measures depict the complexity of a function class differently but are closely related to each other.

In the following, we develop complexity upper bounds for the class of Mixed RePUs network functions. In particular,
these bounds lead to the upper bound for the Pseudo dimension of the function class $\widetilde{\mathcal{F}}$,
and that of $\mathcal{F}^\prime$.

\begin{lemma}[Pseudo dimension of Mixed RePUs multilayer perceptrons]\label{lemmapdim}
	Let $\mathcal{\widetilde{F}}$ be a function class implemented by Mixed RePUs activated multilayer perceptrons with depth $\tilde{\mathcal{D}}$, number of neurons (nodes) $\tilde{\mathcal{U}}$ and size or number of parameters (weights and bias) $\tilde{\mathcal{S}}$. Then the Pseudo dimension of $\mathcal{\widetilde{F}}$ satisfies
	\begin{align*}	\pdim(\mathcal{\widetilde{F}})\le
3p\tilde{\mathcal{D}}\tilde{\mathcal{S}}(\tilde{\mathcal{D}}+\log_2\tilde{\mathcal{U}}),
	\end{align*}
 where $\pdim(\mathcal{F})$ denotes the pseudo dimension of a function class $\mathcal{F}$
\end{lemma}

With Theorem \ref{thm2} and Lemma \ref{lemmapdim}, we can now obtain an upper bound for the
complexity of the class of derivatives of RePU neural networks. This facilitates establishing learning theories for statistical methods involving derivatives.

Due to the symmetry among the arguments of the input of networks in $\mathcal{F}$, the concerned complexities for $\mathcal{F}_{1}^\prime,\ldots,\mathcal{F}_{d}^\prime$ are generally the same. For notational simplicity, we use
$$\mathcal{F}^\prime:=\{\frac{\partial}{\partial x_1} f:f\in\mathcal{F}\}$$
in the main context to denote the quantities of complexities such as pseudo dimension, e.g., we use $\pdim(\mathcal{F}^\prime)$ instead of $\pdim(\mathcal{F}_{j}^\prime)$ for $j=1,\ldots,d$.

\section{Approximation power of RePU neural networks}
\label{sec_approx}

In this section, we establish error bounds for using RePU networks to simultaneously approximate smooth functions and their derivatives. 

We show that RePU neural networks, with an appropriate architecture, can represent multivariate polynomials with no error and thus can simultaneously approximate multivariate differentiable functions and their derivatives.  Moreover, we show that the RePU neural network can mitigate the ``curse of dimensionality" when the domain of the target function concentrates in a neighborhood of a low-dimensional manifold.

In the studies of ReLU network approximation properties \citep{yarotsky2017error,yarotsky2018optimal,shen2019deep, schmidt2020nonparametric},  the analyses rely on two key facts. First, the ReLU activation function can be used to construct continuous, piecewise linear bump functions with compact support, which forms a partition of unity of the domain. Second, deep ReLU networks can approximate the square function $x^2$ to any error tolerance, provided the network is large enough. Based on these facts, the ReLU network can compute Taylor's expansion to approximate smooth functions. However, due to the piecewise linear nature of ReLU, the approximation is restricted to the target function itself rather than its derivative. In other words, the error in approximation by ReLU networks is quantified using the $L_p$ norm, where $p\ge1$ or $p=\infty$. On the other hand, norms such as Sobolev, H\"older, or others can indicate the approximation of derivatives.  \cite{guhring2021approximation} extended the results by showing that the network activated by a general smooth function can approximate the partition of unity and polynomial functions, and obtain the approximation rate for smooth functions in the Sobolev norm which implies approximation of the target function and its derivatives.  RePU-activated networks have been shown to represent splines \citep{duan2021convergence,belomestny2022simultaneous}, thus they can approximate smooth functions and their derivatives based on the approximation power of splines.

RePU networks can also represent polynomials efficiently and accurately. 
This fact motivated us to derive our approximation results for RePU networks based on their representational power on polynomials. To construct RePU networks representing polynomials, our basic idea is to express basic operators as one-hidden-layer RePU networks and then compute polynomials by combining and composing these building blocks. For univariate input $x$, the identity map $x$, linear transformation $ax+b$, and square map $x^2$ can all be represented by one-hidden-layer RePU networks with only a few nodes. The multiplication operator $xy={(x+y)^2-(x-y)^2}/4$ can also be realized by a one-hidden-layer RePU network. Then univariate polynomials $\sum_{i=0}^{N} a_ix^i$ of degree $N\ge 0,$ can be computed by a RePU network with a proper composited construction based on Horner's method  (also known as Qin Jiushao’s algorithm) \citep{horner1819}. Further, a multivariate polynomial can be viewed as the product of univariate polynomials, then a RePU network with a suitable architecture can represent multivariate polynomials. Alternatively,  as mentioned in \cite{mhaskar1993approximation,chui1993realization}, any polynomial in $d$ variables with total degree not exceeding $N$ can be written as a linear combination of
$\binom{N+d}{d}$ quantities of the form $(w^\prime x+b)^N$ where $\binom{N+d}{d}=(N+d)!/(N!d!)$ denotes the combinatorial number and $(w^\prime x+b)$ denotes the linear combination of $d$ variables $x$. Given this fact, RePU networks can also be shown to represent polynomials based on proper construction.

\begin{theorem}[Representation of Polynomials by RePU networks]\label{represent}
	For any non-negative integer $N\in\mathbb{N}_0$ and positive integer $d\in\mathbb{N}^+$,  if 
 $f:\mathbb{R}^d\to\mathbb{R}$ is a polynomial of $d$ variables with total degree $N$, then $f$ can be exactly computed with no error by RePU activated neural network with 
\begin{itemize}
    \item [(1)] $2N-1$ hidden layers,
    $(6p+2)(2N^d-N^{d-1}-N)+2p(2N^d-N^{d-1}-N)/(N-1)=\mathcal{O}(pN^d)$ number of neurons, $(30p+2)(2N^d-N^{d-1}-N)+(2p+1)(2N^d-N^{d-1}-N)/(N-1)=\mathcal{O}(pN^d)$ number of parameters and width $12pN^{d-1}+6p(N^{d-1}-N)/(N-1)=\mathcal{O}(pN^{d-1})$; \vspace{0.2cm}

    \item [(2)] $\lceil\log_p(N)\rceil$ hidden layers, $2\lceil\log_p(N)\rceil(N+d)!/(N!d!)=\mathcal{O}(\log_p(N)N^d)$ number of neurons, $2(\lceil\log_p(N)\rceil+d+1)(N+d)!/(N!d!)=\mathcal{O}((\log_p(N)+d)N^d)$ number of parameters and width $2(N+d)!/(N!d!)=\mathcal{O}(N^d)$,
    \end{itemize}
 where $\lceil a\rceil$ denotes the smallest integer no less than $a\in\mathbb{R}$ and $a!$ denotes the factorial of integer $a$.
\end{theorem}

Theorem \ref{represent} establishes the capability of RePU networks to accurately represent multivariate polynomials of order $N$ through appropriate architectural designs. The theorem introduces two distinct network architectures, based on Horner's method \citep{horner1819} and Mhaskar's method \citep{mhaskar1993approximation} respectively, that can achieve this representation. The RePU network constructed using Horner's method exhibits a larger number of hidden layers but fewer neurons and parameters compared to the network constructed using Mhaskar's method. Neither construction is universally superior, and the choice of construction depends on the relationship between the dimension $d$ and the order $N$, allowing for potential efficiency gains in specific scenarios.

Importantly, RePU neural networks offer advantages over ReLU networks in approximating polynomials. For any positive integers $W$ and $L$, ReLU networks with a width of approximately $\mathcal{O}(WN^d)$ and a depth of $\mathcal{O}(LN^2)$ can only approximate, but not accurately represent, $d$-variate polynomials of degree $N$ with an accuracy of approximately $9N(W+1)^{-7NL}=\mathcal{O}(NW^{-LN})$ \citep{shen2019deep, hon2022simultaneous}. Furthermore, the approximation capabilities of ReLU networks for polynomials are generally limited to bounded regions, whereas RePU networks can precisely compute polynomials over the entire $\mathbb{R}^d$ space.

 Theorem \ref{represent} introduces novel findings that distinguish it from existing research \citep{li2019better,li2020powernet} in several aspects. First, it provides an explicit formulation of the RePU network's depth, width, number of neurons, and parameters in terms of the target polynomial's order $N$ and the input dimension $d$, thereby facilitating practical implementation. Second, the theorem presents Architecture (2), which outperforms previous studies in the sense that it  requires fewer hidden layers for polynomial representation. Prior works, such as \cite{li2019better,li2020powernet}, required RePU networks with $d(\lceil\log_p(N)\rceil+1)$ hidden layers, along with $\mathcal{O}(pN^d)$ neurons and parameters, to represent $d$-variate polynomials of degree $N$ on $\mathbb{R}^d$. However, Architecture (2) in Theorem \ref{represent} achieves a comparable number of neurons and parameters with only $\lceil\log_p(N)\rceil$ hidden layers. Importantly, the number of hidden layers $\lceil\log_p(N)\rceil$ depends only on the polynomial's degree $N$ and is independent of the input dimension $d$. This improvement bears particular significance in dealing with high-dimensional input spaces that is commonly encountered in machine-learning tasks. Lastly, Architecture (1) in Theorem \ref{represent} contributes an additional RePU network construction based on Horner's method \citep{horner1819}, complementing existing results based solely on Mhaskar's method \citep{mhaskar1993approximation} and providing an alternative choice for polynomial representation.

By leveraging the approximation power of multivariate polynomials, we can derive error bounds for approximating general multivariate smooth functions using RePU neural networks. Previously, approximation properties of RePU networks have been studied for target functions in different spaces, e.g. Sobolev space \citep{li2020powernet,li2019better,guhring2021approximation}, spectral Barron space  \citep{siegel2022high}, Besov space \citep{ali2021approximation} and H\"older space \citep{belomestny2022simultaneous}. Here we focus on the approximation of multivariate smooth functions and their derivatives in  $C^s$ space for $s\in\mathbb{N}^+$ defined in Definition \ref{defCs}.

\begin{definition}[Multivariate differentiable class $C^s$]
	\label{defCs}
	A function $f: \mathbb{B}\subset \mathbb{R}^{d}\to\mathbb{R}$  defined on a subset $\mathbb{B}$ of $\mathbb{R}^d$ is said to be in class $C^s(\mathbb{B})$ on $\mathbb{B}$ for a positive integer $s$, if all partial derivatives
	$$D^\alpha f:=\frac{\partial^\alpha }{\partial x_1^{\alpha_1}\partial x_2^{\alpha_2}\cdots\partial x_d^{\alpha_d}}f$$
	exist and are continuous on $\mathbb{B}$, for every $\alpha_1,\alpha_2,\ldots,\alpha_d$ non-negative integers, such that $\alpha_1+\alpha_2+\cdots+\alpha_d\le s$. In addition, we define the norm of $f$ over  $\mathbb{B}$ by
	$$\Vert f\Vert_{C^s} :=
	\sum_{\vert\alpha\vert_1\le s}\sup_{x \in \mathbb{B}}\vert D^\alpha f (x) \vert,$$
	where $\vert\alpha\vert_1:=\sum_{i=1}^d\alpha_i$  for any vector $\alpha=(\alpha_1,\alpha_2,\ldots,\alpha_d)\in\mathbb{R}^d$.
\end{definition}

{\color{black}

\begin{theorem}\label{approx2}
	Let $f$ be a real-valued function defined on a compact set $\mathcal{X}\subset\mathbb{R}^{d}$ belonging to  class $C^s$ for $0\le s<\infty$. For any $N\in\mathbb{N}^+$, there exists a RePU activated neural network $\phi_N$ with its depth $\mathcal{D}$, width $\mathcal{W}$, number of neurons $\mathcal{U}$ and size $\mathcal{S}$ specified as one of the following architectures:
\begin{itemize}
    \item [(1)]
    \begin{align*}
		\mathcal{D}&=2N-1,\quad \mathcal{W}=12pN^{d-1}+6p(N^{d-1}-N)/(N-1),\\
		\mathcal{U}&=(6p+2)(2N^d-N^{d-1}-N)+2p(2N^d-N^{d-1}-N)/(N-1),\\
		\mathcal{S}&=(30p+2)(2N^d-N^{d-1}-N)+(2p+1)(2N^d-N^{d-1}-N)/(N-1);
\end{align*}
\item [(2)]
       \begin{align*}
		\mathcal{D}&=\lceil\log_p(N)\rceil,\quad \mathcal{W}=2(N+d)!/(N!d!),\\
		\mathcal{U}&=2\lceil\log_p(N)\rceil(N+d)!/(N!d!),\\
		\mathcal{S}&=2(\lceil\log_p(N)\rceil+d+1)(N+d)!/(N!d!),
\end{align*}
\end{itemize}
such that for each multi-index $\alpha\in\mathbb{N}^d_0$ satisfying $\vert\alpha\vert_1\le\min\{s,N\}$, we have $$\sup_{\mathcal{X}}\vert D^\alpha (f-\phi_N)\vert\le C_{p,s,d,\mathcal{X}}\, \Vert f\Vert_{C^{\vert \alpha\vert_1}}\, N^{-(s-\vert\alpha\vert_1)},$$ where $C_{p,s,d,\mathcal{X}}$  is a positive constant depending only on $p,d,s$ and the diameter of $\mathcal{X}$.
\end{theorem}

Theorem \ref{approx2} gives a simultaneous approximation result for RePU network approximation since the error is measured in $\Vert\cdot\Vert_{C^{s}}$ norm. It improves over existing results focusing on $L_p$ norms, which  cannot guarantee the approximation of derivatives of the target function \citep{li2019better,li2020powernet}.
It is known that shallow neural networks with smooth activation can simultaneously approximate a smooth function and its derivatives \citep{xu2005simultaneous}. However, the simultaneous approximation of RePU neural networks with respect to norms involving derivatives is still an ongoing research area \citep{guhring2021approximation, duan2021convergence, belomestny2022simultaneous}.  For solving partial differential equations in a Sobolev space with smoothness order 2, \cite{duan2021convergence} showed that ReQU neural networks can simultaneously approximate the target function and its derivative in Sobolev norm $W^{1,2}$. To achieve an accuracy of $\epsilon$, the ReQU networks require $\mathcal{O}(\log_2 d)$ layers and $\mathcal{O}(4d\epsilon^{-d})$ neurons.
	Later \cite{belomestny2022simultaneous} proved that $\beta$-H\"older smooth functions ($\beta>2$) and their derivatives up to order $l$ can be simultaneously approximated with accuracy $\epsilon$ in H\"older norm by a ReQU network with width $\mathcal{O}(\epsilon^{-d/(\beta-l)})$, $\mathcal{O}(\log_2 d)$ layers, and $\mathcal{O}(\epsilon^{-d/(\beta-l)})$ nonzero parameters. \cite{guhring2021approximation} derived simultaneous approximation results for neural networks with general smooth activation functions. Based on \cite{guhring2021approximation}, a RePU neural network with constant layer and $\mathcal{O}(\epsilon^{-d/(\beta-l)})$ nonzero parameters can achieve an approximation accuracy $\epsilon$ measured in Sobolev norm up to $l$th order derivative for a $d$-dimensional Sobolev function with smoothness $\beta$.
	
To achieve the approximation accuracy $\epsilon$, our Theorem \ref{approx2} demonstrates that a RePU network requires a comparable number of neurons, namely $\mathcal{O}(\epsilon^{-d/(s-l)})$, to simultaneously approximate the target function up to its $l$-th order derivatives.  Our result differs from existing studies in several ways. First,
in contrast to \cite{li2019better, li2020powernet}, Theorem \ref{approx2} derives simultaneous approximation results for RePU networks. Second, Theorem \ref{approx2}  holds for general RePU networks ($p\ge2$), including the ReQU network ($p=2$) studied in \cite{duan2021convergence} and \cite{belomestny2022simultaneous}. Third, Theorem \ref{approx2} explicitly specifies the network architecture to facilitate the network design in practice, whereas existing studies determine network architectures solely in terms of orders \citep{li2019better, li2020powernet, guhring2021approximation}.
In addition, as  discussed in the next subsection, Theorem \ref{approx2} can be further improved and adapted to the low-dimensional structured data, which highlights the RePU networks' capability to mitigate the curse of dimensionality in estimation problems. We again refer to Table \ref{tab} for a summary comparison of our work with the existing results.

}

\begin{remark}
	Theorem \ref{represent} is based on the representation power of RePU networks on polynomials as in \cite{li2019better,li2020powernet} and \cite{ali2021approximation}. Other existing works derived approximation results based on the representation of the ReQU neural network on B-splines or tensor-product splines \citep{duan2021convergence,siegel2022high,belomestny2022simultaneous}. 
\end{remark}

	\subsection{Circumventing the curse of dimensionality}
	In Theorem \ref{approx2}, to achieve an approximate error $\epsilon$, the RePU neural network should have $O(\epsilon^{-d/s})$ many parameters. The number of parameters grows polynomially in the desired approximation accuracy $\epsilon$ with an exponent $-d/s$ depending on the dimension $d$. In statistical and machine learning tasks, such an approximation result can make the estimation suffer from the {\it curse of dimensionality}. In other words, when the dimension $d$ of the input data is large, the convergence rate becomes extremely slow. Fortunately, high-dimensional data often have approximate low-dimensional latent structures in many applications, such as computer vision and natural language processing \citep{belkin2003laplacian,hoffmann2009local,fefferman2016testing}. It has been shown that these low-dimensional structures can help mitigate the curse of dimensionality (improve the convergence rate) using ReLU networks \citep{schmidt2019deep,shen2019deep,jiao2023deep,chen2022nonparametric}.  We consider an assumption of approximate low-dimensional support of data distribution \citep{jiao2023deep}, and show that the RePU network can also mitigate the curse of dimensionality under this assumption.
	
\begin{assumption} \label{assump1}
  The predictor $X$ is supported on a $\rho$-neighborhood $\mathcal{M}_\rho$ of a compact $d_\mathcal{M}$-dimensional Riemannian submanifold $\mathcal{M}\subset\mathbb{R}^d$,
  where
		$$\mathcal{M}_\rho=\{x\in\mathbb{R}^d: \inf\{\Vert x-y\Vert_2: y\in\mathcal{M}\}\leq \rho\}, \ \rho \in (0, 1),$$
  and the Riemannian submanifold $\mathcal{M}$ has condition number $1/\tau$, volume $V$ and geodesic covering regularity $R$.	
\end{assumption}

We assume that the high-dimensional data $X$ concentrates in a $\rho$-neighborhood of a low-dimensional manifold. This assumption serves as a relaxation from the stringent requirements imposed by exact manifold assumptions \citep{chen2019efficient,schmidt2019deep}.

With a well-conditioned manifold $\mathcal{M}$, we show that RePU networks possess the capability to adaptively embed the data into a lower-dimensional space while approximately preserving distances. The dimensionality of the embedded representation, as well as the quality of the embedding in terms of its ability to preserve distances, are contingent upon the properties of the approximate manifold, including its radius $\rho$, condition number $1/\tau$, volume $V$, and geodesic covering regularity $R$. For in-depth definitions of these properties, we direct the interested reader to \cite{baraniuk2009random}.

{\color{black}

\begin{theorem}[Improved approximation results]\label{approx_lowdim}
		Suppose that Assumption \ref{assump1} holds. Let $f$ be a real-valued function defined on $\mathbb{R}^{d}$ belonging to class $C^s$ for $0\le s<\infty$. Let $d_\delta= c\cdot d_\mathcal{M}\log(d\cdot VR\tau^{-1}/\delta)/\delta^2$ be an integer satisfying $d_\delta\le d$ for some $\delta\in(0,1)$ and a universal constant $c>0$. Then for any $N\in\mathbb{N}^+$, there exists a RePU activated neural network $\phi_N$
		with its depth $\mathcal{D}$, width $\mathcal{W}$, number of neurons $\mathcal{U}$ and size $\mathcal{S}$ as one of the following architectures:
\begin{itemize}
    \item [(1)]
    \begin{align*}
		\mathcal{D}&=2N-1,\quad \mathcal{W}=12pN^{d_\delta-1}+6p(N^{d_\delta-1}-N)/(N-1),\\
		\mathcal{U}&=(6p+2)(2N^{d_\delta}-N^{d_\delta-1}-N)+2p(2N^{d_\delta}-N^{d_\delta-1}-N)/(N-1),\\
		\mathcal{S}&=(30p+2)(2N^{d_\delta}-N^{d_\delta-1}-N)+(2p+1)(2N^{d_\delta}-N^{d_\delta-1}-N)/(N-1);
\end{align*}
\item [(2)]
       \begin{align*}
		\mathcal{D}&=\lceil\log_p(N)\rceil,\quad \mathcal{W}=2(N+d_\delta)!/(N!d_\delta!),\\
 \mathcal{U}&=2\lceil\log_p(N)\rceil(N+d_\delta)!/(N!d_\delta!),\\
		\mathcal{S}&=2(\lceil\log_p(N)\rceil+d_\delta+1)(N+d_\delta)!/(N!d_\delta!),
\end{align*}
\end{itemize}
such that for each multi-index $\alpha\in\mathbb{N}^d_0$ satisfying $\vert\alpha\vert_1\le1$, $$\mathbb{E}_X\vert D^\alpha (f(X)-\phi_N(X))\vert\le C_{p,s,d_\delta,\mathcal{M}_\rho}\cdot (1-\delta)^{-2} \Vert f\Vert_{C^{\vert \alpha\vert_1}}\,  N^{-(s-\vert\alpha\vert_1)},$$
for $\rho\le C_1 N^{-(s-\vert \alpha\vert_1)}$ with  a universal constant $C_1>0$, where $C_{p,s,d_\delta,\mathcal{M}_\rho}$  is a positive constant depending only on $d_\delta,s,p$ and $\mathcal{M}_\rho$.
\end{theorem}	

When data has a low-dimensional structure, Theorem \ref{approx_lowdim} indicates that the RePU network can approximate $C^s$ smooth function up to $l$th order derivatives with an accuracy $\epsilon$ using $O(\epsilon^{d_\delta/(s-l)})$ neurons. Here the effective dimension $d_\delta$ scales linearly in the intrinsic manifold dimension $d_\mathcal{M}$ and logarithmically in the
ambient dimension $d$ and the features $1/\tau, V, R$ of the manifold.
Compared to Theorem \ref{approx2}, the effective dimensionality in Theorem \ref{approx_lowdim} is $d_\delta$ instead of $d$, which could be a significant improvement especially when the ambient dimension of data $d$ is large but the intrinsic dimension $d_\mathcal{M}$ is small.

}

Theorem \ref{approx_lowdim} shows that RePU neural networks are an effective tool for analyzing data that lies in a neighborhood of a low-dimensional manifold, indicating their potential to mitigate the curse of dimensionality. In particular, this property makes them well-suited to scenarios where the ambient dimension of the data is high, but its intrinsic dimension is low. To the best of our knowledge, our Theorem \ref{approx_lowdim} is the first result of the ability of RePU networks to mitigate the curse of dimensionality. A highlight of the comparison between our result and the existing recent results of \cite{li2019better}, \cite{li2020powernet}, \cite{duan2021convergence}, \cite{abdeljawad2022approximations} and \cite{belomestny2022simultaneous} is given in Table  \ref{tab}.


\section{Deep score estimation}\label{sec_dse}

Deep neural networks have revolutionized many areas of statistics and machine learning,
and one of the
important applications is score function estimation using the score matching method \citep{hyvarinen2005estimation}. Score-based generative models \citep{song2021scorebased}, which learn to generate samples by estimating the gradient of the log-density function, can benefit significantly from deep neural networks. Using a deep neural network allows for more expressive and flexible models, which can capture complex patterns and dependencies in the data. This is especially important for high-dimensional data, where traditional methods may struggle to capture all of the relevant features. By leveraging the power of deep neural networks, score-based generative models can achieve state-of-the-art results on a wide range of tasks, from image generation to natural language processing.  The use of deep neural networks in score function estimation represents a major advance in the field of generative modeling, with the potential to unlock new levels of creativity and innovation. We apply our developed theories of RePU networks to explore the statistical learning theories of deep score matching estimation (DSME).

Let $p_0(x)$ be a probability density function supported on $\mathbb{R}^d$ and $s_0(x)=\nabla_x \log p_0(x)$ be  its score function where $\nabla_x$ is the vector differential operator with respect to the input $x$. The goal of deep score estimation is to model and estimate $s_0$ by a function $s:\mathbb{R}^d\to\mathbb{R}^d$ based on samples $\{X_i\}_{i=1}^n$ from $p_0$ such that $s(x)\approx s_0(x)$.  Here $s$ belongs to a class of deep neural networks.

It worths noting that the neural network $s:\mathbb{R}^d\to\mathbb{R}^d$ used in deep score estimation is a vector-valued function. For a $d$-dimensional input $x=(x_1,\ldots,x_d)^\top\in\mathbb{R}^d$, the output $s(x)=(s_1(x),\ldots,s_d(x))^\top\in\mathbb{R}^d$ is also $d$-dimensional. We let $\nabla_x s$ denote the $n\times n$ Jacobian matrix of $s$ with its $(i,j)$ entry being ${\partial s_i}/{\partial x_j}$. With a slight abuse of notation, we denote $\mathcal{F}_n:=\mathcal{F}_{\mathcal{D},\mathcal{W}, \mathcal{U},\mathcal{S},\mathcal{B},\mathcal{B}^\prime}$ by a class of RePU activated multilayer perceptrons $s: \mathbb{R}^{d} \to \mathbb{R}^d $ with parameter $\theta$, depth $\mathcal{D}$, width $\mathcal{W}$, size $\mathcal{S}$, number of  neurons $\mathcal{U}$ and $s$ satisfying: (i) $\Vert s \Vert_\infty\leq\mathcal{B}$ for some $0 <\mathcal{B}< \infty$ where
$\Vert s \Vert_\infty:=\sup_{x\in\mathcal{X}}\Vert s(x)\Vert_\infty$ is the sup-norm of a vector-valued function $s$ over its domain $x\in\mathcal{X}$; (ii)  $\Vert (\nabla_x s)_{ii}\Vert_\infty \le \mathcal{B}^\prime$, $i=1,\ldots,d$, for some $0<\mathcal{B}^\prime<\infty$ where $(\nabla_x s)_{ii}$ is the $i$-th diagonal entry (in the $i$-th row and $i$-th column) of $\nabla_x s$. Here the parameters $\mathcal{D},\mathcal{W}, \mathcal{U}$ and $\mathcal{S}$ of $\mathcal{F}_n$ can depend on the sample size $n$, but we omit the dependence in their notations. In addition, we extend the definition of smooth multivariate function. We say a multivariate function $s=(s_1,\ldots,s_d)$ belongs to $C^m$ if $s_j$ belongs  to $C^m$  for each $j=1,\ldots,d$. Correspondingly, we define $\Vert s\Vert_{C^{m}}:=\max_{j=1,\ldots,d} \Vert s_j \Vert_{C^{m}}$.

\subsection{Non-asymptotic error bounds for DSME} \label{sec_dse_bound}

The development of theory for predicting the performance of score estimator using deep neural networks has been a crucial research area in recent times. Theoretical upper bounds for prediction errors have become increasingly important in understanding the limitations and potential of these models.

We are interested in establishing non-asymptoic error bounds for
DSME,
which is obtained by minimizing the expected squared distance $\mathbb{E}_{X}\Vert s(X)-s_0(X)\Vert^2_2$ over the class of functions $\mathcal{F}$. However, this objective is computationally infeasible because the explicit form of $s_0$ is unknown. Under proper conditions, the objective function has an equivalent formulation which is computationally feasible.
\begin{assumption}\label{assump2}
	The density $p_0$ of the data $X$ is differentiable.  The expectation $\mathbb{E}_X\Vert s_0(X)\Vert_2^2$ and $\mathbb{E}_X\Vert s(X)\Vert_2^2$ are finite for any $s\in\mathcal{F}$. And $s_0(x)s(x)\to0$ for any $s\in\mathcal{F}$ when $\Vert x\Vert\to\infty$.
\end{assumption}
Under Assumption \ref{assump2}, the population objective of score matching is equivalent to $J$ given in (\ref{obj0}).
With a finite sample $S_n=\{X_i\}_{i=1}^n$, the empirical version of $J$ is $$J_n(s)=\frac{1}{n}\sum_{i=1}^{n}\left\{{\rm tr}(\nabla_x s(X_i))+\frac{1}{2}\Vert s(X_i)\Vert_2^2\right\}.$$
Then, DSME is defined by
\begin{align}\label{erm_score}
	\hat{s}_n:=\arg\min_{s \in \mathcal{F}_n}J_n(s),
\end{align}
 which is the empirical risk minimizer over the class of RePU neural networks $\mathcal{F}_n$.

Our target is to give upper bounds of the excess risk of $\hat{s}_n$, which is defined as
\[
J(\hat{s}_n)-J(s_0) = \frac{1}{2}\mathbb{E}_X\Vert \hat{s}_n(X)-s_0(X)\Vert_2^2.
\]

To obtain an upper bound of $J(\hat{s}_n)-J(s_0)$, we decompose it into two parts of error, i.e. {\it stochastic error} and {\it approximation error}, and then derive upper bounds for them respectively. Let $s_n=\arg\min_{s \in \mathcal{F}_n} J(s)$, then
\begin{align*}
&J(\hat{s}_n)-J(s_0)\\
&=\{J(\hat{s}_n)-J_n(\hat{s}_n)\}+\{J_n(\hat{s}_n)-J_n(s_n)\}+\{J_n(s_n)-J(s_n)\}+\{J(s_n)-J(s_0)\}\\
	&\le \{J(\hat{s}_n)-J_n(\hat{s}_n)\}+\{J_n(s_n)-J(s_n)\}+\{J(s_n)-J(s_0)\}\\
	&\le 2\sup_{s\in\mathcal{F}_n} \vert J(s)-J_n(s)\vert+\inf_{s\in\mathcal{F}_n}\{J(s)-J(s_0)\},
\end{align*}
where we call $2\sup_{s\in\mathcal{F}_n} \vert J(s)-J_n(s)\vert$ the {\it stochastic error} and $\inf_{s\in\mathcal{F}_n}\{J(s)-J(s_0)\}$ the {\it approximation error}.

{\color{black}

It is important to highlight that the analysis of stochastic error and approximation error for DSME estimation are unconventional. On one hand, since $J(s)-J(s_0)=\mathbb{E}_X\Vert s(X)-s_0(X) \Vert_2^2$ holds for any $s$, the approximation error can be obtained by examining the squared distance approximation in the $L_2$ norm. Thus, Theorem \ref{approx2} provides a bound for the approximation error $\inf{s\in\mathcal{F}_n}[{J(s)-J(s_0)}]$. On the other hand, the empirical squared distance loss $\sum_{i=1}^n \Vert s(X_i)- s_0(X_i) \Vert^2_2/(2n)$ is not equivalent to the surrogate loss $J_n$. In other words, the minimizer $\hat{s}_n$ of $J_n$ may not be the same as the minimizer of the empirical squared distance $\sum_{i=1}^n \Vert s(X_i)- s_0(X_i) \Vert^2_2/(2n)$ over $s\in\mathcal{F}_n$. Consequently, the stochastic error can only be analyzed based on the formulation of $J$ rather than the squared loss. This implies that the stochastic error is dependent on the complexities of the RePU networks class $\mathcal{F}_n$, as well as their derivatives $\mathcal{F}_n^\prime$. Based on Theorem \ref{thm2}, Lemma \ref{lemmapdim} and the empirical process theory, it is expected that the stochastic error will be bounded by $\mathcal{O}((\pdim(\mathcal{F}_n)+\pdim(\mathcal{F}_n^\prime))^{1/2} n^{-1/2})$. Finally, by combining these two error bounds, we obtain the following bounds for the mean squared error of the empirical risk minimizer $\hat{s}_n$ defined in (\ref{erm_score}).

\begin{lemma}
\label{non-asymp-score}
	Suppose that Assumption \ref{assump2} hold and the target score function $s_0$ belongs to $C^m(\mathcal{X})$ for $m\in\mathbb{N}^+$. For any positive integer $N\in\mathbb{N}^+$, let $\mathcal{F}_n:=\mathcal{F}_{\mathcal{D},\mathcal{W},\mathcal{U},\mathcal{S},\mathcal{B},\mathcal{B}^\prime}$ be the class of RePU activated neural networks $f:\mathcal{X}\to\mathbb{R}^d$ with depth $\mathcal{D}=\lceil\log_p(N)\rceil$, width $\mathcal{W}=2(N+d)!/(N!d!)$, number of neurons $\mathcal{U}=2\lceil\log_p(N)\rceil(N+d)!/(N!d!)$ and size $\mathcal{S}=2(\lceil\log_p(N)\rceil+d+1)(N+d)!/(N!d!)$,
	and suppose that $\mathcal{B}\ge\Vert s_0\Vert_{C^0}$ and $\mathcal{B}^\prime\ge \max_{i=1,\ldots,d}\Vert (\nabla_xs_0)_{ii}\Vert_{C^1}$. Then the empirical risk minimizer $\hat{s}_n$ defined in (\ref{erm_score}) satisfies
\begin{align} 
\label{non-asymp-score2}
\mathbb{E} \Vert \hat{s}_n(X)-s_0(X)\Vert^2_2  &\le  \mathcal{E}_{sto}+\mathcal{E}_{app},
	\end{align}
with
\begin{align*}
	 \mathcal{E}_{sto}=&C_1 p d^3(\mathcal{B}^2+2\mathcal{B}^\prime)(\log n)^{1/2}{n}^{-1/2} (\log_p N)^2N^{d/2},\\
	\mathcal{E}_{app}=&{C}_2 N^{-2m} \Vert s_{0}\Vert_{C^0}^2,
\end{align*}
	where the expectation $\mathbb{E}$ is taken with respect to $X$ and $\hat{s}_n$, $C_1>0$ is a universal constant,  and $C_2>0$  is a constant depending only on $p,d,m$ and the diameter of $\mathcal{X}$.
\end{lemma}

\begin{remark}
Lemma \ref{non-asymp-score} established a bound on the mean squared error of the empirical risk minimizer. Specifically, this error is shown to be bounded by the sum of the stochastic error, denoted as $\mathcal{E}_{sto}$, and the approximation error, denoted as $\mathcal{E}_{app}$. On one hand, the stochastic error $\mathcal{E}_{sto}$ exhibits a decreasing trend with respect to the sample size $n$, but an increasing trend with respect to the network size as determined by $N$. On the other hand, the approximation error $\mathcal{E}_{app}$ decreases in the network size as determined by $N$. To attain a fast convergence rate with respect to the sample size $n$, it is necessary to carefully balance these two errors by selecting an appropriate $N$ based on a given sample size $n$.
\end{remark}

\begin{remark}
	In Lemma \ref{non-asymp-score}, the error bounds are stated in terms of the integer $N$. These error bounds can also be expressed in terms of the number of neurons $\mathcal{U}$ and size $\mathcal{S}$, given that we have specified the relationships between these parameters. Specifically, $\mathcal{U}=2\lceil\log_p(N)\rceil(N+d)!/(N!d!)$ and size $\mathcal{S}=2(\lceil\log_p(N)\rceil+d+1)(N+d)!/(N!d!)$, which relate the number of neurons and size of the network to  $N$ and the dimensionality of $X$.
\end{remark}

Lemma \ref{non-asymp-score} leads to the following error bound for the score-matching estimator.

\begin{theorem}[Non-asymptotic excess risk bounds]
\label{dseb1}
	Under the conditions of Theorem \ref{non-asymp-score},  
 we set $N=\lfloor n^{1/(d+4m)}\rfloor$.  Then by (\ref{non-asymp-score2}),  the empirical risk minimizer $\hat{s}_n$ defined in (\ref{erm_score}) satisfies
	\begin{align*}
		\mathbb{E} \Vert \hat{s}_n(X)-s_0(X)\Vert^2&\le C (\log n) n^{-\frac{2m}{d+4m}},
	\end{align*}
	where $C $ is a constant only depending on $p,\mathcal{B},\mathcal{B}^\prime,m,d,\mathcal{X}$ and $ \Vert s_0\Vert_{C^1}$.
\end{theorem}

In Theorem \ref{dseb1}, the convergence rate in the error bound is $n^{-\frac{2m}{d+4m}}$ up to a logarithmic factor. While this rate is slightly slower than the optimal minimax rate $n^{-\frac{2m}{d+2m}}$ for nonparametric regression \citep{stone1982optimal}, it remains reasonable considering the nature of score matching estimation. Score matching estimation involves derivatives and the target score function value is not directly observable, which deviates from the traditional nonparametric regression in \cite{stone1982optimal} where both predictors and responses are observed and no derivatives are involved. However, the rate $n^{-\frac{2m}{d+4m}}$ can be extremely slow for large $d$, suffering from the curse of dimensionality. To address this issue, we
derive error bounds under an approximate lower-dimensional support assumption as stated in Assumption \ref{assump1}, to mitigate the curse of dimensionality.

}

\begin{lemma}
\label{non-asymp-score-low}
	Suppose that Assumptions \ref{assump1}, \ref{assump2} hold and the target score function $s_0$ belongs to $C^m(\mathcal{X})$ for some $m\in\mathbb{N}^+$.
	Let $d_\delta=c\cdot d_\mathcal{M}\log(d\cdot VR\tau^{-1}/\delta)/\delta^2)$ be an integer with $d_\delta\le d$ for some $\delta\in(0,1)$ and universal constant $c>0$.
	For any positive integer $N\in\mathbb{N}^+$, let $\mathcal{F}_n:=\mathcal{F}_{\mathcal{D},\mathcal{W},\mathcal{U},\mathcal{S},\mathcal{B},\mathcal{B}^\prime}$ be the class of RePU activated neural networks $f:\mathcal{X}\to\mathbb{R}^d$ with depth $\mathcal{D}=\lceil\log_p(N)\rceil$, width $\mathcal{W}=2(N+d_\delta)!/(N!d_\delta!)$, number of neurons $\mathcal{U}=2\lceil\log_p(N)\rceil(N+d_\delta)!/(N!d_\delta!)$ and size $\mathcal{S}=2(\lceil\log_p(N)\rceil+d_\delta+1)(N+d_\delta)!/(N!d_\delta!)$. Suppose that $\mathcal{B}\ge\Vert s_0\Vert_{C^0}$ and $\mathcal{B}^\prime\ge \max_{i=1,\ldots,d}\Vert (\nabla_xs_0)_{ii}\Vert_{C^1}$. Then the empirical risk minimizer $\hat{s}_n$ defined in (\ref{erm_score}) satisfies
	{\color{black}
	\begin{align} 
	\label{non-asymp-score-low_}
		\mathbb{E} \Vert \hat{s}_n(X)-s_0(X)\Vert^2_2  &\le  \mathcal{E}_{sto}+\tilde{\mathcal{E}}_{app},
	\end{align}
with
	\begin{align*}
		\mathcal{E}_{sto}&=C_1 pd^2d_\delta(\mathcal{B}^2+2\mathcal{B}^\prime)(\log n)^{1/2}{n}^{-1/2} N^{d_\delta/2}, \\
		\tilde{\mathcal{E}}_{app}&={C}_2 (1-\delta)^{-2}  \Vert s_{0}\Vert_{C^0}^2 N^{-2m},
	\end{align*}
}
for $\rho\le C_\rho N^{-2m}$, where  $C_\rho,C_1>0$ are universal constants and $C_2>0$  is a constant depending only on $p,d,d_\delta,m$ and $\mathcal{M}_\rho$.
\end{lemma}

With an approximate low-dimensional support assumption, Lemma \ref{non-asymp-score-low} implies that a faster convergence rate for deep score estimator can be achieved.

{\color{black}
\begin{theorem}[Improved non-asymptotic excess risk bounds]
\label{non-asymp-score-low2}
In (\ref{non-asymp-score-low_}), we can set $N=\lfloor n^{d_\delta/(d_\delta+4m)}\rfloor$, then the empirical risk minimizer $\hat{s}_n$ defined in (\ref{erm_score}) satisfies
\begin{align*}
	\mathbb{E} \Vert \hat{s}_n(X)-s_0(X)\Vert^2&\le C n^{-\frac{2m}{d_\delta+4m}},
\end{align*}
where $C$ is a constant only depending on $p,\mathcal{B},\mathcal{B}^\prime,m,d,d_\delta,\mathcal{X}$ and $ \Vert s_0\Vert_{C^1}$.
\end{theorem}
}


\section{Deep isotonic regression}\label{sec_dir}
As another application of our results on RePU-activated networks, we propose
PDIR, a penalized deep isotonic regression approach using
RePU networks and a penalty function based on the derivatives of the networks to enforce monotonicity. We also establish the error bounds for PDIR.

 Suppose we have a random sample  $S:=\{(X_i,Y_i)\}_{i=1}^n$ from model (\ref{reg0}).
 Recall $R^\lambda$ is the proposed population objective function for isotonic regression defined in (\ref{iro1}).
We consider the empirical counterpart of the objective function $R^\lambda$:
\begin{equation}
	\label{perisk}
	\mathcal{R}^\lambda_n(f)=\frac{1}{n}\sum_{i=1}^{n}\Big\{\vert Y_i-f(X_i)\vert^2+ \frac{1}{d}\sum_{j=1}^{d} \lambda_j \rho(\dot{f}_j(X_i))\Big\}.
\end{equation}
A simple choice of the penalty function $\rho$ is
 $\rho(x)=\max\{-x,0\}$. In general, we can take  $\rho(x)=h(\max\{-x,0\})$ for a function $h$ with $h(0)=0$.
We focus on Lipschitz penalty functions as defined below.

\begin{assumption}[Lipschitz penalty function]\label{assump3}
	The penalty function $\rho(\cdot):\mathbb{R}\to[0,\infty)$ satisfies $\rho(x)=0$ if $x\ge0$. Besides, $\rho$ is $\kappa$-Lipschitz, i.e.,
	$\vert \rho(x_1)-\rho(x_2)\vert\le\kappa \vert x_1-x_2\vert,$
	for any $x_1,x_2\in\mathbb{R}$.
\end{assumption}

 Let the empirical risk minimizer of deep isotonic regression denoted by
\begin{equation}\label{erm}
	\hat{f}_n^\lambda:\in\arg\min_{f\in\mathcal{F}_n} \mathcal{R}^\lambda_n(f),
\end{equation}
where $\mathcal{F}_n$ is a class of functions computed by deep neural networks which may depend on and can be set to depend on the sample size $n$.
We refer to $\hat f^\lambda_n$ as a
penalized deep isotonic regression (PDIR) estimator.

\begin{figure}[H]
	\centering
	\includegraphics*[width=4 in, height=3.0 in]{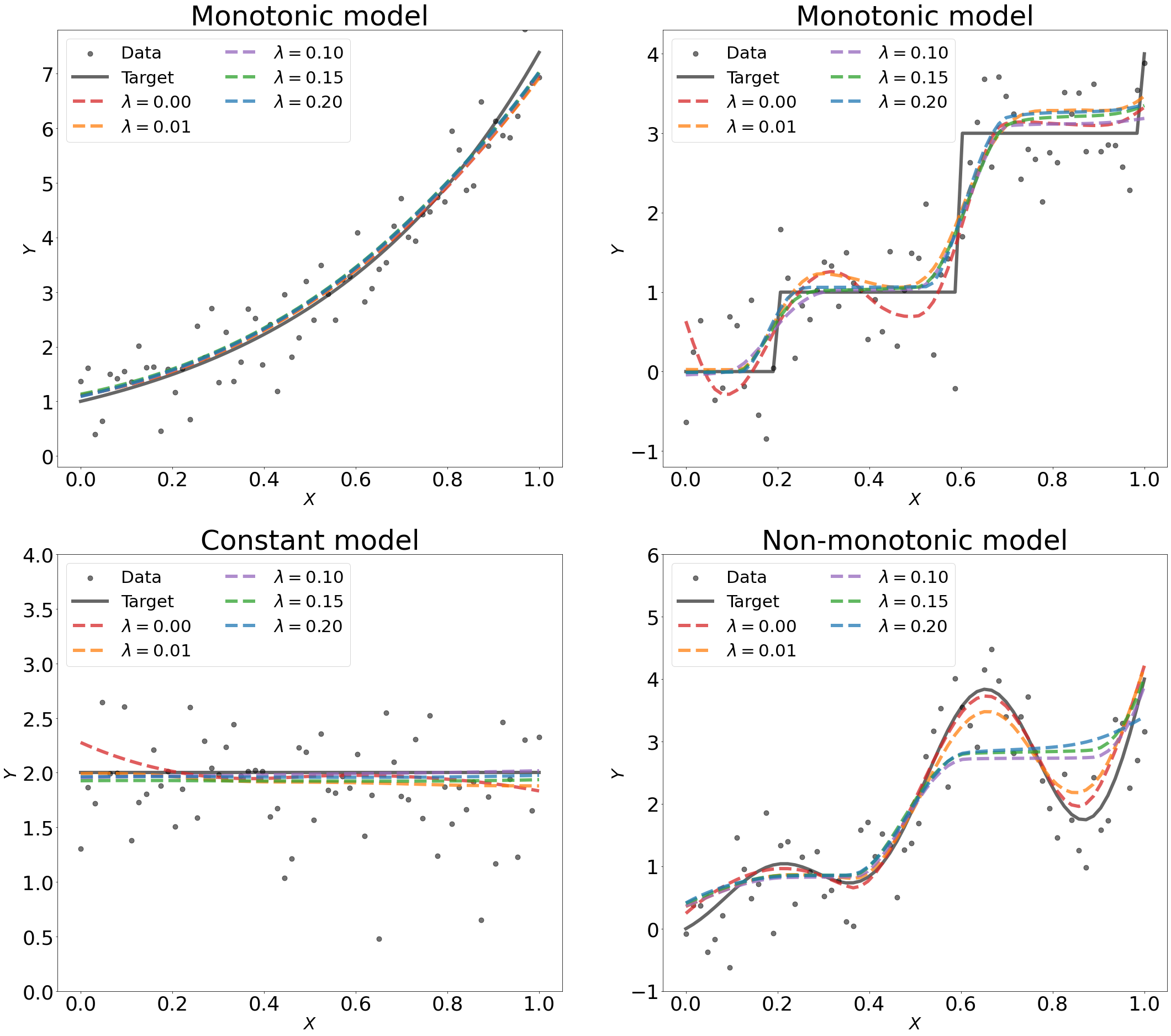}
	\caption{Examples of PDIR estimates. In all figures, the data points are depicted as grey dots, the underlying regression functions are plotted as solid black curves, and PDIR estimates with different levels of penalty parameter $\lambda$ are plotted as colored curves. In the top two figures, data are generated from models with monotonic regression functions. In the bottom left figure, the target function is a constant. In the bottom right figure, the model is misspecified, in which the underlying regression function is not monotonic. Small values of  $\lambda$ can lead to non-monotonic estimated functions.}
	\label{fig:dem}
\end{figure}

An illustration of PDIR is presented in Figure \ref{fig:dem}. In all subfigures, the data are depicted as grey dots, the underlying regression functions are plotted as solid black curves and PDIR estimates with different levels of penalty parameter $\lambda$ are plotted as colored curves. In the top two figures, data are generated from models with monotonic regression functions. In the bottom left figure, the target function is a constant. In the bottom right figure, the model is misspecified, in which the underlying regression function is not monotonic. Small values of  $\lambda$ can lead to non-monotonic and reasonable estimates,
suggesting that PDIR is robust against model misspecification.
We have conducted more numerical experiments to evaluate the performance of PDIR,  which indicates that PDIR tends to perform better than the existing isotonic regression methods considered in the comparison.
The results are given in the Supplementary Material.

\subsection{Non-asymptotic error bounds for PDIR} 
\label{sec4.1}

In this section, we state our main results on the bounds for the excess risk
of the PDIR estimator defined in (\ref{erm}).
Recall the definition of $R^\lambda$  in (\ref{iro1}).
For notational simplicity, we write
\begin{equation}\label{risk}
	\mathcal{R}(f)=\mathcal{R}^0(f) = \mathbb{E}\vert Y-f(X)\vert^2.
\end{equation}
The target function $f_0$ is the minimizer of risk $\mathcal{R}(f)$ over measurable functions, i.e., $f_0\in\arg\min_{f} \mathcal{R}(f).$
In isotonic regression, we assume that $f_0 \in \mathcal{F}_0.$
In addition, for any function $f$,  under the regression model (\ref{reg0}), we have
$$\mathcal{R}(f)-\mathcal{R}(f_0)=\mathbb{E}\vert f(X)-f_0(X)\vert^2.$$

We first state the conditions needed for establishing the excess risk bounds.

\begin{assumption}\label{assump4}
	(i) The target regression function $f_0: \mathcal X  \to \mathbb R$ defined in (\ref{reg0}) is coordinate-wisely nondecreasing on $\mathcal{X}$, i.e., $f_0(x)\le f_0(y)$ if $x\preceq y$
	for $x, y \in\mathcal{X}\subseteq\mathbb{R}^d$.
	(ii) The errors $\epsilon_i, i=1,\ldots,n,$
	are independent and identically distributed noise variables with $\mathbb{E}(\epsilon_i)=0$ and ${\rm Var}(\epsilon_i)\le\sigma^2$, and $\epsilon_i$'s are independent of $\{X_i\}_{i=1}^n$.
\end{assumption}
Assumption \ref{assump4} includes basic model assumptions on the errors and the monotonic target function $f_0$. In addition, we assume that the target function $f_0$ belongs to
 the class $C^s$.

Next, we state the following basic lemma for bounding the excess risk.

\begin{lemma}[Excess risk decomposition]\label{decom}
	For the empirical risk minimizer $\hat{f}^\lambda_n$ defined in (\ref{erm}), its excess risk can be upper bounded by
	\begin{align*}
		\mathbb{E}\vert\hat{f}^\lambda_n(X)-f_0(X)\vert^2=&\mathbb{E}\Big\{\mathcal{R}(\hat{f}_n^\lambda)-\mathcal{R}(f_0)\Big\}\le	\mathbb{E}\Big\{\mathcal{R}^\lambda(\hat{f}_n^\lambda)-\mathcal{R}^\lambda(f_0)\Big\}\\
		\le&\mathbb{E}\Big\{\mathcal{R}^\lambda(\hat{f}^\lambda_n)
		-2\mathcal{R}^\lambda_n(\hat{f}^\lambda_n)+\mathcal{R}^\lambda(f_0)\Big\}
		+2\inf_{f\in\mathcal{F}_n}\Big[\mathcal{R}^\lambda(f)-\mathcal{R}^\lambda(f_0)\Big].
	\end{align*}
\end{lemma}

The upper bound for the excess risk can be decomposed into two components: the stochastic error, given by the expected value of $\mathcal{R}^\lambda(\hat{f}^\lambda_n)-2\mathcal{R}^\lambda_n(\hat{f}^\lambda_n)+\mathcal{R}^\lambda(f_0)$, and the approximation error, defined as $\inf_{f\in\mathcal{F}_n}[\mathcal{R}^\lambda(f)-\mathcal{R}^\lambda(f_0)]$. To establish a bound for the stochastic error, it is necessary to consider the complexities of both RePU networks and their derivatives, which have been investigated in our Theorem \ref{thm2} and Lemma \ref{lemmapdim}. To establish a bound for the approximation error $\inf_{f\in\mathcal{F}_n}{\mathcal{R}^\lambda(f)-\mathcal{R}^\lambda(f_0)}$, we rely on the simultaneous approximation results in Theorem \ref{approx2}.


\begin{remark}
	The error decomposition in Lemma \ref{decom} differs from the canonical decomposition for score estimation  in section \ref{sec_dse_bound}, particularly pertaining to the stochastic error component. However, utilizing the decomposition in Lemma \ref{decom} enables us to derive a superior stochastic error bound by leveraging the properties of the PDIR loss function.
A similar decomposition for least squares loss without penalization can be found  in \cite{jiao2023deep}.
\end{remark}

{\color{black}

\begin{lemma}
	\label{non-asymp}
		Suppose that Assumptions \ref{assump3}, \ref{assump4} hold and the target function $f_0$ defined in (\ref{reg0}) belongs to $C^s$ for some $s\in\mathbb{N}^+$.
        For any positive integer $N\in\mathbb{N}^+$, let $\mathcal{F}_n:=\mathcal{F}_{\mathcal{D},\mathcal{W},\mathcal{U},\mathcal{S},\mathcal{B},\mathcal{B}^\prime}$ be the class of RePU activated neural networks $f:\mathcal{X}\to\mathbb{R}$ with depth $\mathcal{D}=\lceil\log_p(N)\rceil$, width $\mathcal{W}=2(N+d)!/(N!d!)$, number of neurons $\mathcal{U}=2\lceil\log_p(N)\rceil(N+d)!/(N!d!)$ and size $\mathcal{S}=2(\lceil\log_p(N)\rceil+d+1)(N+d)!/(N!d!)$.
		Suppose that $\mathcal{B}\ge\Vert f_0\Vert_{C^0}$ and $\mathcal{B}^\prime\ge\Vert f_0\Vert_{C^1}$. Then for $n\ge\max\{\pdim(\mathcal{F}_n),\pdim(\mathcal{F}^\prime_n)\}$, the excess risk of PDIR $\hat{f}^\lambda_n$ defined in (\ref{erm}) satisfies
	
		\begin{align}\label{non-asympb2}
			\mathbb{E} \vert \hat{f}^\lambda_n(X)-f_0(X)\vert^2 &\le \mathcal{E}_{sto}+ \mathcal{E}_{app},\\ \label{non-asympb3}
			\mathbb{E} \big[\frac{1}{d}\sum_{j=1}^{d}\lambda_j \rho (\pj\hat{f}^\lambda_{n}(X))\big] &\le \mathcal{E}_{sto}+ \mathcal{E}_{app},
		\end{align}
	with
	\begin{align*} \mathcal{E}_{sto}&=C_1 d^3\big\{p^3\mathcal{B}^3+(\kappa\bar{\lambda}\mathcal{B}^\prime)^2\big\}(\log n)^2n^{-1}(\log_p N)^3N^{d},\\
			\mathcal{E}_{app}&=C_2\cdot\Vert f_0\Vert^2_{C^1}(N^{-2s}+\kappa\bar{\lambda}N^{-(s-1)}) ,
\end{align*}
where the expectation $\mathbb{E}$ is taken with respect to $X$ and $\hat{f}^\lambda_n$, $\bar{\lambda}=\sum_{j=1}^{d}\lambda_j/d$ is the mean of the tuning parameters, $C_1>0$ is a universal constant and $C_2>0$  is a positive constant depending only on $d,s$ and the diameter of the support $\mathcal{X}$.
\end{lemma}

Lemma \ref{non-asymp} establishes two
error bounds for the PDIR estimator $\hat{f}^\lambda_n$: (\ref{non-asympb2}) for the mean squared error between $\hat{f}^\lambda_n$ and the target $f_0$, and (\ref{non-asympb3}) for controlling the non-monotonicity of $\hat{f}^\lambda_n$ via its partial derivatives $\pj\hat{f}_{n},j=1,\ldots,d,$ with respect to a measure defined in terms of $\rho$.  Both bounds (\ref{non-asympb2}) and (\ref{non-asympb3}) are encompasses both stochastic and approximation errors.  Specifically, the stochastic error is of order $\mathcal{O}(N^{d}/n)$, which represents an improvement over the canonical error bound of $\mathcal{O}([N^{d}/n]^{1/2})$, up to logarithmic factors in $n$. This advancement is owing to the decomposition in Lemma \ref{decom} and the properties of PDIR loss function, which is different from traditional decomposition techniques.

\begin{remark}
In (\ref{non-asympb3}), the estimator $\hat{f}^{\lambda}_n$ is encouraged to exhibit monotonicity, as the expected monotonicity penalty on the estimator $\mathbb{E}[\frac{1}{d}\sum_{j=1}^{d}\lambda_j \rho (\pj\hat{f}^\lambda_{n}(X))]$ is bounded.
Notably, when $\mathbb{E}[\rho (\pj\hat{f}^\lambda_{n}(X))]=0$, the estimator $\hat{f}^\lambda_{n}$ is almost surely monotonic in its $j$th argument with respect to the probability measure of $X$. Based on (\ref{non-asympb3}),
guarantees of the estimator's monotonicity with respect to a single argument can be obtained. Specifically, for those $j$ where $\lambda_j\not= 0$, we have $\mathbb{E}[\rho(\pj \hat{f}^\lambda_n(X))]\leq d(\mathcal{E}_{sto}+\mathcal{E}_{app})/\lambda_j$, which provides a guarantee of the estimator's monotonicity with respect to its $j$th argument. Moreover,  larger values of $\lambda_j$ lead to smaller bounds, which is consistent with the intuition that larger values of $\lambda_j$ better promote
monotonicity of $\hat{f}^\lambda_n$ with respect to its $j$th argument.
\end{remark}

\begin{theorem} [Non-asymptotic excess risk bounds]
\label{non-asymp2}
Under the conditions of Lemma \ref{non-asymp}
to achieve the smallest error bound in (\ref{non-asympb2}), we can set $N=\lfloor n^{1/(d+2s)}\rfloor$ and $\lambda_j= n^{-(s+1)/(d+2s)}$ for $j=1,\ldots,d$.  Then we have
	\begin{align*}
		\mathbb{E} \vert \hat{f}^\lambda_n(X)-f_0(X)\vert^2&\le C (\log n)^{5} n^{-\frac{2s}{d+2s}},
	\end{align*}
and
\begin{align*}
	\mathbb{E}[\rho(\pj \hat{f}^\lambda_n(X))]\leq C (\log n)^5 n^{-\frac{s-1}{d+2s}},
\end{align*}
for $j=1,\ldots,d$, where $C>0$ is a constant depending only on $\mathcal{B},\mathcal{B}^\prime,s,d,\mathcal{X}, \Vert f_0\Vert_{C^s}$ and $\kappa$.
\end{theorem}

By Theorem \ref{non-asymp2}, our proposed PDIR estimator obtained with proper network architecture and tuning parameter achieves the minimax optimal rate $\mathcal{O}(n^{-\frac{2s}{d+2s}})$ up to logarithms for the nonparametric regression \citep{stone1982optimal}. Meanwhile, the PDIR estimator $\hat{f}^\lambda_n$ is guaranteed to be monotonic as measured by $\mathbb{E}[\rho(\pj \hat{f}^\lambda_n(X))]$ at a rate of $\mathcal{O(}n^{-(s-1)/(d+2s)})$ up to a logarithmic factor.

}

\begin{remark}
In Theorem \ref{non-asymp2}, we choose $\lambda_j=n^{-(s+1)/(d+2s)},\ j=1,\ldots,d,$ to attain the optimal rate of the expected mean squared error of $\hat{f}^\lambda_n$ up to a logarithmic factor. Additionally, we guarantee that the estimator $\hat{f}^\lambda_n$ to be monotonic at a rate of $n^{-(s-1)/(d+2s)}$ up to a logarithmic factor as measured by $\mathbb{E}[\rho(\pj \hat{f}^\lambda_n(X))]$.
The choice of $\lambda_j$ is not unique for ensuring the consistency of $\hat{f}^\lambda_n$. In fact, any choice of $\bar{\lambda}=o((\log n)^{-2}n^{(s-1)/(d+2s)})$ will result in a consistent $\hat{f}^\lambda_n$. However, larger values of $\bar{\lambda}$ lead to a slower convergence rate of the expected mean squared error, but   better guarantee for the monotonicity of $\hat{f}^\lambda_n$.
\end{remark}

The smoothness $s$ of the target function $f_0$ is unknown in practice and how to determine the smoothness of an unknown function is an important but nontrivial problem.
Note that the convergence rate $(\log n)^{5} n^{-2s/(d+2s)}$
suffers from the curse of dimensionality since it can be extremely slow if $d$ is large.

High-dimensional data  have low-dimensional latent structures in many applications. Below we show that PDIR can mitigate the curse of dimensionality if the data distribution is supported on an approximate low-dimensional manifold.

\begin{lemma}\label{non-asymp-low}
	Suppose that Assumptions \ref{assump1}, \ref{assump3}, \ref{assump4} hold and the target function $f_0$ defined in (\ref{reg0}) belongs to $C^s$ for some $s\in\mathbb{N}^+$.
    Let $d_\delta=c\cdot d_\mathcal{M}\log(d\cdot VR\tau^{-1}/\delta)/\delta^2)$ be an integer with $d_\delta\le d$ for some $\delta\in(0,1)$ and universal constant $c>0$.
    For any positive integer $N\in\mathbb{N}^+$, let $\mathcal{F}_n:=\mathcal{F}_{\mathcal{D},\mathcal{W},\mathcal{U},\mathcal{S},\mathcal{B},\mathcal{B}^\prime}$ be the class of RePU activated neural networks $f:\mathcal{X}\to\mathbb{R}$ with depth $\mathcal{D}=\lceil\log_p(N)\rceil$, width $\mathcal{W}=2(N+d_\delta)!/(N!d_\delta!)$, number of neurons $\mathcal{U}=2\lceil\log_p(N)\rceil(N+d_\delta)!/(N!d_\delta!)$ and size $\mathcal{S}=2(\lceil\log_p(N)\rceil+d_\delta+1)(N+d_\delta)!/(N!d_\delta!)$.
		Suppose that $\mathcal{B}\ge\Vert f_0\Vert_{C^0}$ and $\mathcal{B}^\prime\ge\Vert f_0\Vert_{C^1}$. Then for $n\ge\max\{\pdim(\mathcal{F}_n),\pdim(\mathcal{F}^\prime_n)\}$, the excess risk of the PDIR estimator $\hat{f}^\lambda_n$ defined in (\ref{erm}) satisfies
	
		\begin{align}\label{non-asymp-low2}
				\mathbb{E} \vert \hat{f}^\lambda_n(X)-f_0(X)\vert^2 &\le \mathcal{E}_{sto}+ \tilde{\mathcal{E}}_{app},\\ \label{non-asymp-low3}
				\mathbb{E} \big[\frac{1}{d}\sum_{j=1}^{d}\lambda_j \rho (\pj\hat{f}^\lambda_{n}(X))\big] &\le \mathcal{E}_{sto}+ \tilde{\mathcal{E}}_{app},
		\end{align}
			with
		\begin{align*}		\mathcal{E}_{sto}&=C_1d^2d_\delta\big\{p^3\mathcal{B}^3+(\kappa\bar{\lambda}\mathcal{B}^\prime)^2\big\}(\log n)^2n^{-1}N^{d_\delta},\\
		\tilde{\mathcal{E}}_{app}&=C_2(1-\delta)^2\Vert f_0\Vert^2_{C^s} (N^{-2s}+\kappa\bar{\lambda} N^{-(s-1)}),
		\end{align*}
for $\rho\le C_\rho N^{-(s-\vert \alpha\vert_1)},$ where $C_\rho, C_1>0$ are universal constants and $C_2>0$	is a constant depending only on $d_\delta,s$ and the diameter of the support $\mathcal{M}_\rho$.
\end{lemma}

Based on  Lemma \ref{non-asymp-low}, we obtain the following result.

\begin{theorem}[Improved non-asymptotic excess risk bounds]
\label{reg-lowD}
Under the conditions of Lemma \ref{non-asymp-low}, to achieve the smallest error bound in (\ref{non-asymp-low2}), we can set $N=\lfloor n^{1/(d_\delta+2s)}\rfloor$ and { $\lambda_j=n^{-(s+1)/(d_\delta+2s)}$ } for $j=1,\ldots,d$. Then we have
\begin{align*}
	\mathbb{E} \vert \hat{f}^\lambda_n(X)-f_0(X)\vert^2&\le C (\log n)^{5} n^{-\frac{2s}{d_\delta+2s}},
\end{align*}
and for $j=1,\ldots,d$,
\begin{align*}
	\mathbb{E}[\rho(\pj \hat{f}^\lambda_n(X))]\leq C (\log n)^5 n^{-\frac{2s}{d_\delta+2s}},
\end{align*}
where $C>0$ is a constant depending only on $\mathcal{B},\mathcal{B}^\prime,s,d,d_\delta,\mathcal{M}_\rho, \Vert f_0\Vert_{C^s}$ and $\kappa$.
\end{theorem}

In Theorem \ref{reg-lowD}, the effective dimension is $d_\delta$ rather than large $d$. Therefore, the rate of convergence
is an improvement over the result in Theorem \ref{non-asymp2}
when the intrinsic dimension $d_\delta$ is smaller than the ambient dimension $d$.

\subsection{PDIR under model misspecification}
\label{misspec}

In this subsection, we investigate PDIR under model misspecification when Assumption \ref{assump4} (i) is not satisfied, meaning that the underlying regression function $f_0$ may not be monotonic.

Let $S:={(X_i,Y_i)}_{i=1}^n$ be a random sample from model (\ref{reg0}). Recall that the penalized risk of the deep isotonic regression is given by
\begin{equation*}
	\mathcal{R}^\lambda(f)=\mathbb{E}\vert Y-f(X)\vert^2+\frac{1}{d}\sum_{j=1}^d \lambda_j \mathbb{E}{\rho(\dot{f}_j(X))}.
\end{equation*}
If  $f_0$ is not monotonic, the penalty $\sum_{j=1}^d \lambda_j \mathbb{E}[\rho(\dot{f}_j(X))/d]$ is non-zero, and consequently, $f_0$ is not a minimizer of the risk $\mathcal{R}^\lambda$ when $\lambda_j\not=0, \forall j$. Intuitively, the deep isotonic regression estimator will exhibit a bias towards the target $f_0$ due to the additional penalty terms in the risk. However, it is reasonable to expect that the estimator $\hat{f}^\lambda_n$ will have a smaller bias if $\lambda_j,j=1,\ldots,d$ are small. In the following lemma, we establish a non-asymptotic upper bound for our proposed deep isotonic regression estimator while adapting to model misspecification.

\begin{lemma}\label{non-asymp-mis}
	Suppose that Assumptions \ref{assump3} and \ref{assump4} (ii) hold and the target function $f_0$ defined in (\ref{reg0}) belongs to $C^s$ for some $s\in\mathbb{N}^+$.
 For any positive integer  $N\in\mathbb{N}^+$ let $\mathcal{F}_n:=\mathcal{F}_{\mathcal{D},\mathcal{W},\mathcal{U},\mathcal{S},\mathcal{B},\mathcal{B}^\prime}$ be the class of RePU activated neural networks $f:\mathcal{X}\to\mathbb{R}^d$ with depth $\mathcal{D}=\lceil\log_p(N)\rceil$, width $\mathcal{W}=2(N+d)!/(N!d!)$, number of neurons $\mathcal{U}=2\lceil\log_p(N)\rceil(N+d)!/(N!d!)$ and size $\mathcal{S}=2(\lceil\log_p(N)\rceil+d+1)(N+d)!/(N!d!)$.
Suppose that $\mathcal{B}\ge\Vert f_0\Vert_{C^0}$ and $\mathcal{B}^\prime\ge\Vert f_0\Vert_{C^1}$. Then for $n\ge\max\{\pdim(\mathcal{F}_n),\pdim(\mathcal{F}^\prime_n)\}$, the excess risk of the PDIR estimator $\hat{f}^\lambda_n$ defined in (\ref{erm}) satisfies
	\begin{align}\label{non-asymp-mis2}
		\mathbb{E} \vert \hat{f}^\lambda_n(X)-f_0(X)\vert^2 \le &\mathcal{E}_{sto}+ \mathcal{E}_{app} +\mathcal{E}_{mis},\\ \label{non-asymp-mis3}
		\mathbb{E} \big[\frac{1}{d}\sum_{j=1}^{d}\lambda_j \rho (\pj\hat{f}^\lambda_{n}(X))\big] &\le \mathcal{E}_{sto}+ \mathcal{E}_{app}+\mathcal{E}_{mis},
	\end{align}
	with
	\begin{align*}
\mathcal{E}_{sto}&=C_1p^2d^3(\mathcal{B}^2+\kappa\bar{\lambda}\mathcal{B}^\prime)(\log n)^{1/2} n^{-1/2} (\log_p N)^{3/2}N^{d/2},\\
	\mathcal{E}_{app}&=C_2 \Vert f_0\Vert^2_{C^1}(N^{-2s}+\kappa\bar{\lambda}N^{-(s-1)}),\\
		\mathcal{E}_{mis}&=\frac{1}{d}\sum_{j=1}^d\lambda_j\mathbb{E}[\rho(\pj f_0(X))],
	\end{align*}
where the expectation $\mathbb{E}$ is taken with respect to $X$ and $\hat{f}^\lambda_n$, $\bar{\lambda}=\sum_{j=1}^{d}\lambda_j/d$ is the mean of the tuning parameters, $C_1>0$ is a universal constant and $C_2>0$  is a positive constant depending only on $d,s$ and the diameter of the support $\mathcal{X}$.
\end{lemma}

Lemma \ref{non-asymp-mis} is a generalized version of Lemma \ref{non-asymp} for PDIR, as it holds regardless of whether the target function is isotonic or not. In Lemma \ref{non-asymp-mis}, the expected mean squared error of the PDIR estimator $\hat{f}^\lambda_n$ can be bounded by three errors: stochastic error $\mathcal{E}_{sto}$, approximation error $\mathcal{E}_{app}$, and misspecification error $\mathcal{E}_{mis}$, without the monotonicity assumption. Compared with Lemma \ref{non-asymp} with the monotonicity assumption, the approximation error is identical, the stochastic error is worse in terms of order, and the misspecification error appears as an extra term in the inequality. With an appropriate setup of $N$ for the neural network architecture with respect to the sample size $n$, the stochastic error and approximation error can converge to zero, albeit at a slower rate than that in Theorem \ref{non-asymp2}. However, the misspecification error remains constant for fixed tuning parameters $\lambda_j$. Thus, we can let the tuning parameters $\lambda_j$ converge to zero to achieve consistency.

\begin{remark}
It is worth noting that if the target function is isotonic, then the misspecification error vanishes, leading the scenario to that of isotonic regression. However, the convergence rate based on Lemma \ref{non-asymp-mis} is slower than that in Lemma \ref{non-asymp}. The reason is that Lemma \ref{non-asymp-mis} is general and holds without prior knowledge of the monotonicity of the target function. If knowledge is available about the non-isotonicity of the $j$th argument of the target function $f_0$, setting the corresponding $\lambda_j=0$ decreases the misspecification error and helps improve the upper bound. 
\end{remark}

{\color{black}
	
\begin{theorem} [Non-asymptotic excess risk bounds]
	\label{non-asymp-mis4}
    Under the conditions of Lemma \ref{non-asymp-mis}, to achieve the fastest convergence rate in (\ref{non-asymp-mis2}), we can set $N=\lfloor n^{1/(d+4s)}\rfloor$ and $\lambda_j= n^{-2s/(d+4s)}$ for $j=1,\ldots,d$.  Then we have
	\begin{align*}
		\mathbb{E} \vert \hat{f}^\lambda_n(X)-f_0(X)\vert^2&\le C (\log n) n^{-\frac{2s}{d+4s}},
	\end{align*}
 where $C>0$ is a constant  depending only on $\mathcal{B},\mathcal{B}^\prime,s,d,\mathcal{X}, \Vert f_0\Vert_{C^s}$ and $\kappa$.
\end{theorem}

According to Lemma \ref{non-asymp-mis}, under the misspecification model, the prediction error of PDIR attains its minimum when $\lambda_j=0$ for $j=1,\ldots,d$, and the misspecification error $\mathcal{E}_{mis}$ vanishes. Consequently, the optimal convergence rate with respect to $n$ can be achieved by setting $N=\mathcal{O}(\lfloor n^{1/(d+4s)}\rfloor)$ and $\lambda_j=0$ for $j=1,\ldots,d$. It is worth noting that the prediction error of PDIR can achieve this rate as long as $\bar{\lambda}= \mathcal{O}n^{-2s/(d+4s)}$.

\begin{remark}
According to Theorem \ref{non-asymp-mis4},  there is no unique choice of $\lambda_j$ that ensures the consistency of PDIR. Consistency is guaranteed even under a misspecified model when the $\lambda_j$ for $j=1,\ldots,d$ tend to zero as $n\to\infty$.
Additionally, selecting a smaller value of $\bar{\lambda}$ provides a better upper bound for (\ref{non-asymp-mis2}), and an optimal rate up to logarithms of $n$ can be achieved with a sufficiently small $\bar{\lambda}=O(n^{-2s/(d+4s)})$. An example demonstrating the effects of tuning parameters is visualized in the last subfigure of Figure \ref{fig:dem}.
\end{remark}
}

\section{Related works}
\label{related}
In this section, we briefly review the papers in the existing literature that are most related to the present work.

\subsection{ReLU and RePU networks}
Deep learning has achieved impressive success in a wide range of applications. A fundamental reason for these successes is the ability of deep neural networks to approximate high-dimensional functions and extract effective data representations. There has been much effort devoted to studying the approximation properties of deep neural networks in recent years.
Many interesting results have been obtained concerning the approximation power of deep neural networks for multivariate functions. Examples include \cite{
chen2019efficient}, \cite{
schmidt2020nonparametric},
\cite{jiao2023deep}.
These works focused on the power of ReLU-activated neural networks for approximating various types of smooth functions.

For the approximation of the square function by ReLU networks, \citet{yarotsky2017error} first used ``sawtooth" functions, which achieves an error rate of $\mathcal{O}(2^{-L})$ with width 6 and depth $\mathcal{O}(L)$ for any positive integer $L\in\mathbb{N}^+$. General construction of ReLU networks for approximating a square function can achieve an error $N^{-L}$ with width $3N$ and depth $L$ for any positive integers $N, L\in\mathbb{N}^+$ \citep{lu2021deep}. Based on this basic fact, the ReLU networks approximating multiplication and polynomials can be constructed correspondingly. However, the network complexity in terms of network size (depth and width) for a ReLU network to achieve precise approximation can be large compared to that of a RePU network since a RePU network can exactly compute polynomials with fewer layers and neurons.

The approximation results of the RePU network are generally obtained by converting splines or polynomials into RePU networks and making use of the approximation results of splines and polynomials. The universality of sigmoidal deep neural networks has been studied in the pioneering works \citep{mhaskar1993approximation,chui1994neural}.  In addition, the approximation properties of shallow Rectified Power Unit (RePU) activated network were studied in \cite{klusowski2018approximation,siegel2022high}. The approximation rates of deep RePU neural networks on target functions in different spaces have also been explored, including Besov spaces \citep{ali2021approximation}, Sobolev spaces \citep{li2019better,li2020powernet,duan2021convergence,abdeljawad2022approximations} , and H\"older space \citep{belomestny2022simultaneous}.  Most of the existing results on the expressiveness of neural networks measure the quality of approximation with respect to $L_p$ where $ p\ge1$ norm. However, fewer papers have studied the approximation of derivatives of smooth functions \citep{duan2021convergence,guhring2021approximation,belomestny2022simultaneous}.

\subsection{Related works on score estimation}

Learning a probability distribution from data is a fundamental task in statistics and machine learning 
for efficient generation of new samples from the learned distribution. Likelihood-based models approach this problem by directly learning the probability density function, but they have several limitations, such as an intractable normalizing constant and approximate maximum likelihood training.

One alternative approach to circumvent these limitations is to model the score function \citep{liu2016kernelized}, which is the gradient of the logarithm of the probability density function. Score-based models can be learned using a variety of methods, including parametric score matching methods \citep{hyvarinen2005estimation,sasaki2014clustering}, autoencoders as its denoising variants \citep{vincent2011connection}, sliced score matching \citep{song2020sliced}, nonparametric score matching \citep{sriperumbudur2017density,sutherland2018efficient}, and kernel estimators based on Stein's methods \citep{li2017gradient,shi2018spectral}. These score estimators have been applied in many research problems, such as gradient flow and optimal transport methods \citep{gao2019deep,gao2022deep}, gradient-free adaptive MCMC \citep{strathmann2015gradient}, learning implicit models \citep{warde2016improving},  inverse problems \citep{jalal2021robust}. Score-based generative learning models, especially those using deep neural networks, have achieved state-of-the-art performance in many downstream tasks and applications, including image generation   \citep{song2019generative,song2020improved,song2021scorebased,ho2020denoising,dhariwal2021diffusion,ho2022cascaded}, music generation \citep{mittal2021symbolic}, and audio synthesis \citep{chen2020wavegrad,kong2020diffwave,popov2021grad}.

However, there is a lack of  theoretical understanding of nonparametric score estimation using deep neural networks. The existing studies mainly considered kernel based methods.
 \cite{zhou2020nonparametric}
 studied regularized nonparametric score estimators using vector-valued reproducing kernel Hilbert space, which connects the kernel exponential family estimator \citep{sriperumbudur2017density} with the score estimator based on Stein's method \citep{li2017gradient,shi2018spectral}. Consistency and convergence rates of these kernel-based score estimator are also established under the correctly-specified model assumption in \cite{zhou2020nonparametric}. For denoising autoencoders, \cite{block2020generative} obtained generalization bounds for general nonparametric estimators also under the correctly-specified model assumption.

For sore-based learning using deep neural networks, the main difficulty for establishing the theoretical foundation is the lack of knowledge of differentiable neural networks since the derivatives of neural networks are involved in the estimation of score function. Previously, the non-differentiable Rectified Linear Unit (ReLU) activated deep neural network has received much attention due to its attractive properties in computation and optimization, and has been extensively studied in terms of its complexity \citep{bartlett1998almost,anthony1999,bartlett2019nearly} and approximation power \citep{yarotsky2017error,petersen2018optimal,shen2019deep,lu2020deep,jiao2023deep}, based on which statistical learning theories for deep non-parametric estimations were established \citep{bauer2019deep,schmidt2020nonparametric,jiao2023deep}.
For deep neural networks with differentiable activation functions, such as ReQU and RepU,
the simultaneous approximation power on a smooth function and its derivatives were studied recently  \citep{ali2021approximation,belomestny2022simultaneous,siegel2022high,hon2022simultaneous}, but the statistical properties of differentiable networks are still largely unknown.
To the best of our knowledge,
the statistical learning theory has only been investigated for ReQU networks in \cite{shen2022estimation}, where they have
developed network representation of the derivatives of ReQU networks and studied their complexity.

\subsection{Related works on isotonic regression}

There is a rich and extensive literature on univariate isotonic regression, which is too vast to be adequately summarized here. So we refer to the books
\citet{bbbb1972} and \cite{rwd1988} for a systematic treatment of this topic and review of earlier works.
For more recent developments on the error analysis of nonparametric isotonic regression,
we refer to  \citet{durot2002sharp,zhang2002risk,durot2007mathbb,
durot2008monotone,groeneboom2014nonparametric,chatterjee2015risk}, and \citet{yang2019contraction}, among others.

The least squares isotonic regression estimators under fixed design were extensively studied. With a fixed design at fixed points $x_1, \ldots, x_n$, the $L_p$ risk of the least squares estimator is defined by
	$\mathcal{R}_{n,p}(\hat{f}_0)=\mathbb{E}(n^{-1}\sum_{i=1}^n\vert \hat{f}_0(x_i)-f_0(x_i)\vert^p)^{1/p},$
	where the least squares estimator $\hat{f}_0$ is defined by
	\begin{equation}\label{lse0}
		\hat{f}_0=\arg\min_{f\in\mathcal{F}_0}\frac{1}{n} \sum_{i=1}^n\{Y_i-f(x_i)\}^2.
	\end{equation}
The problem can be restated in terms of isotonic vector estimation on directed acyclic graphs.
Specifically, the design points $\{x_1,\ldots,x_n\}$ induce a directed acyclic graph $G_x(V(G_x),E(G_x))$ with vertices $V(G_x)=\{1,\ldots,n\}$ and edges $E(G_x)=\{(i,j):x_i\preceq x_j\}$. The class of isotonic vectors on $G_x$ is defined by
	$$\mathcal{M}(G):=\{\theta\in\mathbb{R}^{V(G_x)}:\theta_x\leq \theta_y{\rm\ for } x\preceq y\}.$$
Then the least squares estimation in (\ref{lse0})
becomes that of searching for a target vector $\theta_0=\{(\theta_0)_i\}_{i=1}^n:=\{f_0(x_i)\}_{i=1}^n\in\mathcal{M}(G_x)$.
The least squares estimator $\hat{\theta}_0=\{(\hat{\theta}_0)_i\}_{i=1}^n:=\{\hat{f}_0(x_i)\}_{i=1}^n$ is actually the projection of $\{Y_i\}_{i=1}^n$ onto the polyhedral convex cone $\mathcal{M}(G_x)$ \citep{han2019isotonic}. 

For univariate isotonic least squares regression with a bounded total variation target function $f_0$, \citet{zhang2002risk} obtained sharp upper bounds for $\mathcal{R}_{n,p}$ risk of the least squares estimator $\hat{\theta}_0$ for $1\le p<3$.
Shape-constrained estimators were also considered in different settings where automatic rate-adaptation phenomenon happens \citep{chatterjee2015risk,gao2017minimax, bellec2018sharp}.
We also refer to \citet{kim2018adaptation,chatterjee2019adaptive} for other examples of adaptation in univariate shape-constrained problems.
	
Error analysis for the least squares estimator  in multivariate isotonic regression is more difficult.
For two-dimensional isotonic regression, where $X\in \mathbb{R}^d$ with $d=2$  and Gaussian noise, \citet{chatterjee2018matrix} considered the fixed lattice design case and obtained sharp error bounds.
\citet{han2019isotonic} extended the results of \citet{chatterjee2018matrix}
 to the case with $d\ge3$, both from a worst-case perspective and an adaptation point of view. 	 They also proved parallel results for random designs
assuming  the density of the covariate $X$ is bounded away from  zero and infinity on the support.

\citet{deng2020isotonic} considered a class of block estimators for multivariate isotonic regression in
$\mathbb{R}^d$ involving rectangular upper and lower sets under, which is defined as any estimator in-between the following max-min and min-max estimator. Under a $q$-th moment condition on the noise, they developed $L_q$ risk bounds for such estimators for isotonic regression on graphs.
{\color{black}
Furthermore, the block estimator possesses an oracle property in variable selection: when $f_0$ depends on only an unknown set of $s$ variables, the $L_2$ risk of the block estimator automatically achieves the minimax
rate  up to a logarithmic factor based on the knowledge of the set of the $s$ variables.
}

Our proposed method and theoretical results are different from those in the aforementioned papers in several aspects.
First, the resulting estimates from our method are smooth instead of piecewise constant as those based on the existing methods. Second, our method can mitigate the curse of dimensionality under an
approximate low-dimensional manifold support assumption, which is weaker than the exact low-dimensional space assumption in the existing work. Finally,  our method possesses a robustness property against model specification in the sense that it still yields consistent estimators if the monotonicity assumption is not strictly satisfied. However, the properties of the existing isotonic regression methods under model misspecification are unclear.

\section{Conclusions}\label{conclusion}

{\color{black}

In this work, motivated by the problems of score estimation and isotonic regression, we have studied the properties of RePU-activated neural networks, including a novel generalization result for the derivatives of RePU networks and improved approximation error bounds for RePU networks with approximate low-dimensional structures. We have established non-asymptotic excess risk bounds for DSME, a deep score matching estimator;  and PDIR, our proposed penalized deep isotonic regression method.

Our findings highlight the potential of RePU-activated neural networks in addressing challenging problems in machine learning and statistics. The ability to accurately represent the partial derivatives of RePU networks with RePUs mixed-activated networks is a valuable tool in many applications that require the use of neural network derivatives. Moreover, the improved approximation error bounds for RePU networks with low-dimensional structures demonstrate their potential to mitigate the curse of dimensionality in high-dimensional settings.

Future work can investigate further the properties of RePU networks, such as their stability, robustness, and interpretability. It would also be interesting to explore the use of RePU-activated neural networks in other applications, such as nonparametric variable selection  and more general shape-constrained estimation problems.
Additionally, our work can be extended to other smooth activation functions beyond RePUs, such as Gaussian error linear unit  and scaled exponential linear unit, and study their derivatives and approximation properties.
}

\appendix
\numberwithin{equation}{section}
\makeatletter
\newcommand{\section@cntformat}{Appendix \thesection:\ }

\renewcommand{\thetable}{S\arabic{table}}
\renewcommand{\thefigure}{S\arabic{figure}}

\makeatother


\section*{Appendix}

This appendix contains results from simulation studies to evaluate the performance of PDIR and proofs and supporting lemmas for the theoretical results stated in the paper.

\section{Numerical studies}\label{appendix_a}
In this section, we conduct simulation studies to evaluate the performance of PDIR and compare it with the existing isotonic regression methods. The methods included in the simulation include
\begin{itemize}
	\item The isotonic least squares estimator, denoted by {\it Isotonic LSE}, is defined as the minimizer of the mean square error on the training data subject to the monotone constraint. As the squared loss only involves the values at $n$ design points, then the isotonic LSE (with no more than $\binom{n}{2}$ linear constraints) can be computed with quadratic programming or using convex optimization algorithms \citep{dykstra1983algorithm,kyng2015fast,stout2015isotonic}. Algorithmically, this turns out to be mappable to a network flow problem \citep{picard1976maximal,spouge2003least}. In our implementation, we compute {\it Isotonic LSE} via the Python package {\it multiisotonic}\footnote{https://github.com/alexfields/multiisotonic}.
	
	\item The block estimator \citep{deng2020isotonic}, denoted by {\it Block estimator}, is defined as any estimator between the block min-max and max-min estimators \citep{fokianos2020integrated}. In the simulation, we take the {\it Block estimator}  as the mean of max-min and min-max estimators as suggested in \citep{deng2020isotonic}.   The {\it Isotonic LSE} is shown to has an explicit mini-max representation on the design points for isotonic regression on graphs in general \citep{rwd1988}.
	As in \citet{deng2020isotonic}, we use brute force which exhaustively calculates means over all blocks and finds the max-min value for each point $x$. The computation cost via brute force is of order $n^3$.
	
	\item Deep isotonic regression estimator as described in Section \ref{sec_dir}, denoted by \textit{PDIR}. Here we focus on using RePU $\sigma_{p}$ activated network with $p=2$. We implement it in Python via \textit{Pytorch} and use \textit{Adam} \citep{kingma2014adam} as the optimization algorithm with default learning rate 0.01 and default $\beta=(0.9,0.99)$ (coefficients used for computing running averages of gradients and their squares). The tuning parameters are chosen in the way that $\lambda_j=\log(n)$ for $j=1,\ldots,d$.
	
	\item Deep nonparametric regression estimator, denoted by \textit{DNR}, which is actually the {\it PDIR} without penalty. The implementation is the same as that of {\it PDIR}, but the tuning parameters $\lambda_j=0$ for $j=1,\ldots,d$.
\end{itemize}

\subsection{Estimation and evaluation}

For the proposed {\it PDIR} estimator, we set the tuning parameter $\lambda_j=\log(n)$ for $j=1,\ldots,d$ across the simulations.
For each target function $f_0$, according to model (\ref{reg0}) we generate the training data $S_{\rm train}=(X_i^{\rm train},Y_i^{\rm train})_{i=1}^n$ with sample size $n$ and train the {\it Isotonic LSE}, {\it Block estimator}, {\it PDIR} and {\it DNR} estimators on $S_{\rm train}$. We would mention that the {\it Block estimator} has no definition when the input $x$ is ``outside" the domain of training data $S_{\rm train}=(X_i^{\rm train},Y_i^{\rm train})_{i=1}^n$, i.e., there exist no $i,j\in\{1,\ldots,n\}$ such that $X^{\rm train}_i\preceq x\preceq X^{\rm train}_j$. In view of this, in our simulation we focus on using the training data with lattice design of the covariates $(X^{\rm train}_i)_{i=1}^n$ for ease of presentation on the {\it Block estimator}. For {\it PDIR} and {\it DNR} estimators, such fixed lattice design of the covariates are not necessary and the obtained estimators can be smoothly extended to large domains which covers the domain of the training samples.

For each $f_0$, we also generate the testing data $S_{\rm test}(X_t^{\rm test},Y_t^{\rm test})_{t=1}^T$ with sample size $T$ from the same distribution of the training data. For the proposed method and for each obtained estimator $\hat{f}_n$, we calculate the mean squared error (MSE) on the testing data $S_{\rm test}=(X_t^{\rm test},Y_t^{\rm test})_{t=1}^T$. We calculate the $L_1$ distance between the estimator $\hat{f}_n$ and the corresponding target function $f_0$ on the testing data by
\begin{align*}
	\Vert\hat{f}_n-f_0\Vert_{L^1(\nu)}=\frac{1}{T}\sum_{t=1}^T \Big\vert\hat{f}_n(X_t^{\rm test})-f_0(X_t^{\rm test})\Big\vert,
\end{align*}
and we also calculate the $L_2$ distance between the estimator $\hat{f}_n$ and the target function $f_0$, i.e.
\begin{align*}
	\Vert\hat{f}_n-f_0\Vert_{L^2(\nu)}=\sqrt{\frac{1}{T}\sum_{t=1}^T \Big\vert\hat{f}_n(X_t^{\rm test})-f_0(X_t^{\rm test})\Big\vert^2}.
\end{align*}

In the simulation studies, for each data generation model we generate $T=100^d$ testing data by the lattice points (100 even lattice points for each dimension of the input) where $d$ is the dimension of the input. We report the mean squared error, $L_1$  and $L_2$ distances to the target function defined above and their standard deviations over $R = 100$ replications under different scenarios.  The specific forms of $f_0$ are given in the  data generation models below.

\subsection{Univariate models}\label{sec:univariate}

We consider three basic  univariate models, including ``Linear'', ``Exp'', ``Step", ``Constant'' and ``Wave'', which corresponds to different specifications of the target function $f_0$. The formulae are given below.

\begin{itemize}
	\setlength\itemsep{0.05 cm}
	\item [(a)] Linear :
	$$Y=f_0(x)+\epsilon=2x+\epsilon,$$
	\item [(b)] Exp: $$Y=f_0(X)+\epsilon=\exp(2X)+\epsilon,$$
	\item [(c)] Step: $$Y=f_0(X)+\epsilon=\sum h_i I(X\ge t_i)+\epsilon,$$
	\item [(d)] Constant: $$Y=f_0(X)+\epsilon=\exp(2X)+\epsilon,$$
	\item [(e)] Wave: $$Y=f_0(X)+\epsilon=4X+2X\sin(4\pi X)+\epsilon, $$
\end{itemize}
where $(h_i)=(1,2,2)$, $(t_i)=(0.2,0.6,1)$ and $\epsilon\sim N(0,\frac{1}{4})$ follows normal distribution.
We use the linear model as a baseline model in our simulations and expect all the methods
perform well under the linear model. The ``Step" model is monotonic but not smooth even continuous. The ``Constant'' is a monotonic but not strictly monotonic model. And the ``Wave'' is a
nonlinear, smooth but non monotonic model. These models are chosen so that we can evaluate the
performance of \textit{Isotonic LSE},  \textit{Block estimator} {\it PDIR} and \textit{DNR} under different types of models, including the conventional and misspecified cases.

\begin{figure}[H]
	\includegraphics[width=0.95\textwidth]{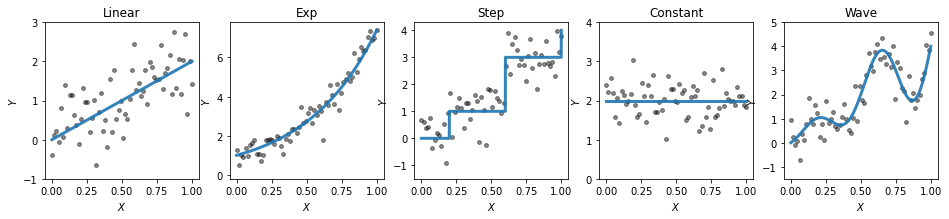}
	\caption{Univariate data generation models. The target functions are depicted by solid curves in blue and instance samples with size $n=64$ are depicted as black dots.}
	\label{fig:target1D}
\end{figure}

For these models, we use the lattice design for the ease of presentation of {\it Block estimator}, where $(X_i^{\rm train})_{i=1}^n$ are the lattice points evenly distributed on interval $[0,1]$.
Figure \ref{fig:target1D} shows all these univariate data generation models.

\begin{figure}[H]
	\includegraphics[width=0.95\textwidth]{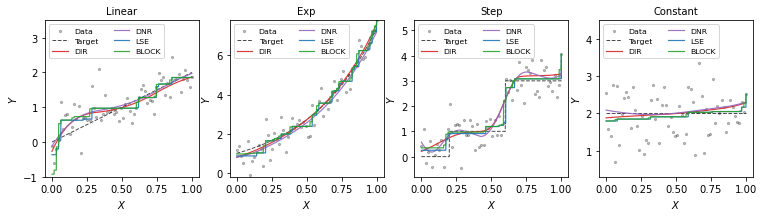}
	\caption{An instance of the estimated curves for the ``Linear'', ``Exp", ``Step" and ``Constant" models when sample size $n=64$. The training data is depicted as grey dots. The target functions are depicted as dashed curves in black, and the estimated functions are represented by solid curves with different colors.}
	\label{fig:univariate}
\end{figure}

Figures \ref{fig:univariate} shows an instance of the estimated curves for the ``Linear'', ``Exp", ``Step" and ``Constant" models when sample size $n=64$.
In these plots, the training data is depicted as grey dots. The target functions are depicted as dashed curves in black, and the estimated functions are represented by solid curves with different colors. The summary statistics are presented in Table \ref{tab:1}. Compared with the piece-wise constant estimates of {\it Isotonic LSE} and {\it Block estimator}, the PDIR estimator is smooth and it works reasonably well under univariate models, especially for models with smooth target functions.

\begin{table}[H]
	\renewcommand\arraystretch{1.3}
	\centering
	\caption{Summary statistics for the simulation results under different univariate models ($d=1$). The averaged mean squared error of the estimates on testing data and $L_1, L_2$ distance to the target function are calculated over 100 replications. The standard deviations are reported in parenthesis.}
	\label{tab:1}
	\resizebox{\textwidth}{!}{%
		\begin{tabular}{@{}cc|ccc|ccc@{}}
			\toprule
			\multirow{2}{*}{Model}    & \multirow{2}{*}{Method} & \multicolumn{3}{c|}{$n=64$}                    & \multicolumn{3}{c}{$n=256$}                   \\
			&                         & MSE           & $L_1$         & $L_2$         & MSE           & $L_1$         & $L_2$         \\ \midrule
			\multirow{4}{*}{Linear}
			& DNR                     & 0.266 (0.011) & 0.101 (0.035) & 0.122 (0.040) &  \bf 0.253 (0.011) &\bf 0.055 (0.020) &\bf 0.068 (0.023) \\
			& PDIR                     & \bf 0.265 (0.012) & \bf 0.098 (0.037) &\bf 0.118 (0.041) & 0.254 (0.012) & 0.058 (0.024) & 0.070 (0.027) \\
			& Isotonic LSE            & 0.282 (0.013) & 0.140 (0.027) & 0.177 (0.035) & 0.262 (0.012) & 0.088 (0.012) & 0.113 (0.017) \\
			& Block                   & 0.330 (0.137) & 0.165 (0.060) & 0.243 (0.155) & 0.277 (0.033) & 0.106 (0.021) & 0.153 (0.060) \\ \hline
			\multirow{4}{*}{Exp}      & DNR                     & \bf 0.268 (0.014) & 0.103 (0.043) & 0.124 (0.049) & 0.256 (0.012) & 0.055 (0.024) & 0.068 (0.027) \\
			& PDIR                     & 0.268 (0.017) &  \bf 0.102 (0.049) &\bf 0.124 (0.056) &\bf 0.255 (0.012) & \bf 0.055 (0.022) &\bf 0.068 (0.026) \\
			& Isotonic LSE            & 0.312 (0.018) & 0.195 (0.028) & 0.246 (0.034) & 0.274 (0.014) & 0.120 (0.014) & 0.153 (0.018) \\
			& Block                   & 0.302 (0.021) & 0.177 (0.028) & 0.223 (0.034) & 0.272 (0.012) & 0.115 (0.015) & 0.146 (0.017) \\\hline
			\multirow{4}{*}{Step}     & DNR                     & 0.375 (0.045) & 0.259 (0.059) & 0.347 (0.061) & 0.315 (0.017) & 0.169 (0.022) & 0.253 (0.018) \\
			& PDIR                     & 0.366 (0.042) & 0.245 (0.058) & 0.335 (0.057) & 0.311 (0.018) & 0.153 (0.025) & 0.245 (0.018) \\
			& Isotonic LSE            &\bf 0.304 (0.020) &\bf 0.151 (0.041) &\bf 0.228 (0.039) &\bf 0.275 (0.014) & \bf 0.081 (0.022) &\bf 0.155 (0.020) \\
			& Block                   & 0.382 (0.217) & 0.208 (0.082) & 0.327 (0.160) & 0.295 (0.046) & 0.108 (0.035) & 0.197 (0.086) \\ \hline
			\multirow{4}{*}{Constant} & DNR                     & 0.266 (0.012) & 0.102 (0.038) & 0.122 (0.042) & 0.258 (0.013) & 0.057 (0.021) & 0.069 (0.023) \\
			& PDIR                     &\bf 0.260 (0.011) &\bf 0.080 (0.045) &\bf 0.092 (0.049) &\bf 0.257 (0.012) & 0.051 (0.025) & 0.060 (0.028) \\
			& Isotonic LSE            & 0.265 (0.013) & 0.087 (0.044) & 0.114 (0.052) & 0.258 (0.012) &\bf 0.044 (0.020) & 0.068 (0.025) \\
			& Block                   & 0.264 (0.012) & 0.085 (0.044) & 0.108 (0.049) & 0.258 (0.012) & 0.044 (0.020) &\bf 0.066 (0.025) \\ \hline
			\multirow{4}{*}{Wave}     & DNR                     &\bf 0.289 (0.023) &\bf 0.156 (0.039) &\bf 0.192 (0.044) &\bf 0.262 (0.014) &\bf 0.089 (0.025) &\bf 0.110 (0.029) \\
			& PDIR                     & 0.530 (0.030) & 0.398 (0.026) & 0.528 (0.018) & 0.511 (0.022) & 0.368 (0.014) & 0.510 (0.009) \\
			& Isotonic LSE            & 0.525 (0.027) & 0.399 (0.022) & 0.524 (0.015) & 0.495 (0.020) & 0.353 (0.009) & 0.494 (0.004) \\
			& Block                   & 0.516 (0.024) & 0.391 (0.022) & 0.519 (0.017) & 0.497 (0.023) & 0.358 (0.012) & 0.500 (0.013) \\ \bottomrule
		\end{tabular}%
	}
\end{table}

\clearpage

\subsection{Bivariate models}\label{sec:bivariate}

We consider several basic multivariate models, including polynomial model (``Polynomial''), concave model (``Concave''), step model (``Step"),  partial model (``Partial"), constant model (``Constant'') and wave model (``Wave''), which correspond to different specifications of the target function $f_0$. The formulae are given below.

\begin{figure}[H]
	\includegraphics[width=0.95\textwidth]{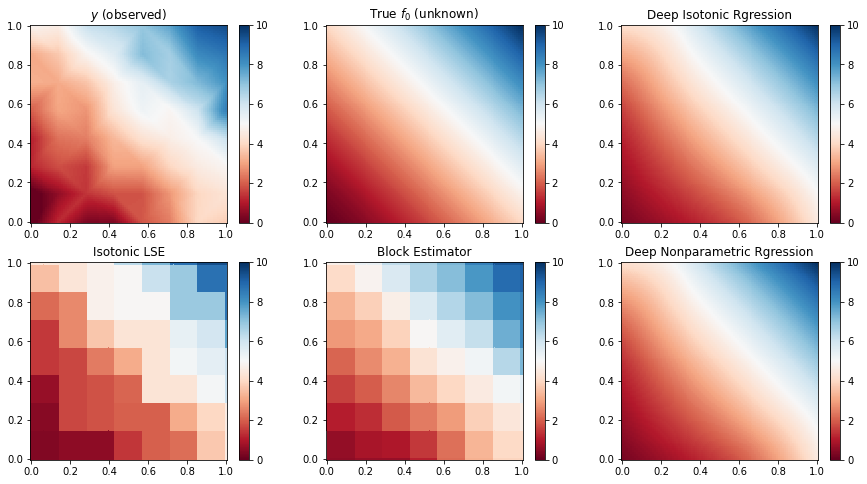}
	\caption{Heatmaps for the target function $f_0$, the observed training data, and its deep isotonic regression and isotonic least squares estimate (isotonic LSE) under model (a) when $d=2$ and $n=64$.}
	\label{fig:a}
\end{figure}

\begin{figure}[H]
	\includegraphics[width=\textwidth]{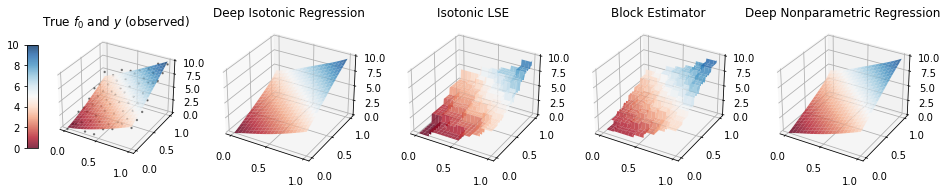}
	\caption{3D surface plots for the target function $f_0$, the observed training data, and its deep isotonic regression and isotonic least squares estimate (isotonic LSE) under model (a) when $d=2$ and $n=64$.}
	\label{fig:a1}
\end{figure}

\begin{figure}[H]
	\includegraphics[width=0.95\textwidth]{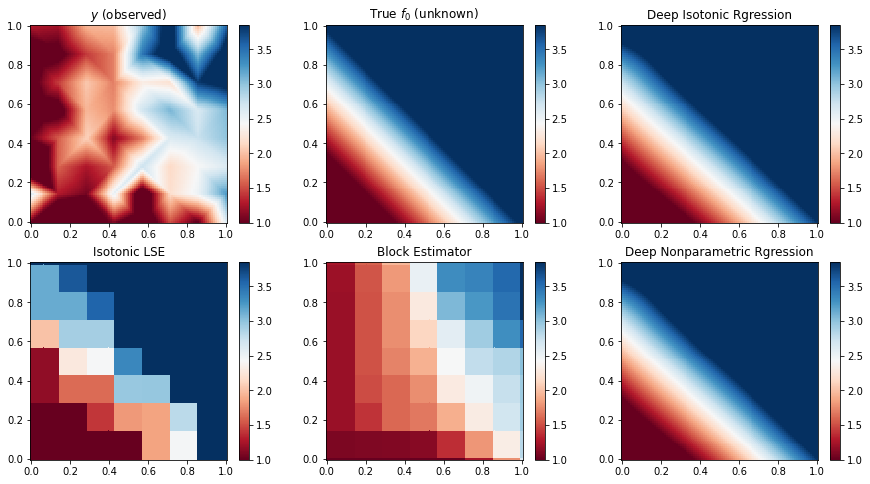}
	\caption{Heatmaps for the target function $f_0$, the observed training data, and its deep isotonic regression and isotonic least squares estimate (isotonic LSE) under model (b) when $d=2$ and $n=64$.}
	\label{fig:b}
\end{figure}

\begin{figure}[H]
	\includegraphics[width=\textwidth]{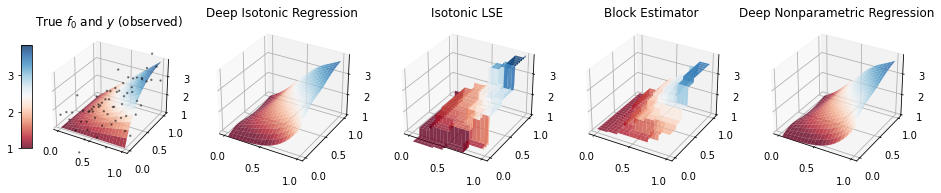}
	\caption{3D surface plots for the target function $f_0$, the observed training data, and its deep isotonic regression and isotonic least squares estimate (isotonic LSE) under model (b) when $d=2$ and $n=64$.}
	\label{fig:b1}
\end{figure}

\begin{figure}[H]
	\includegraphics[width=0.95\textwidth]{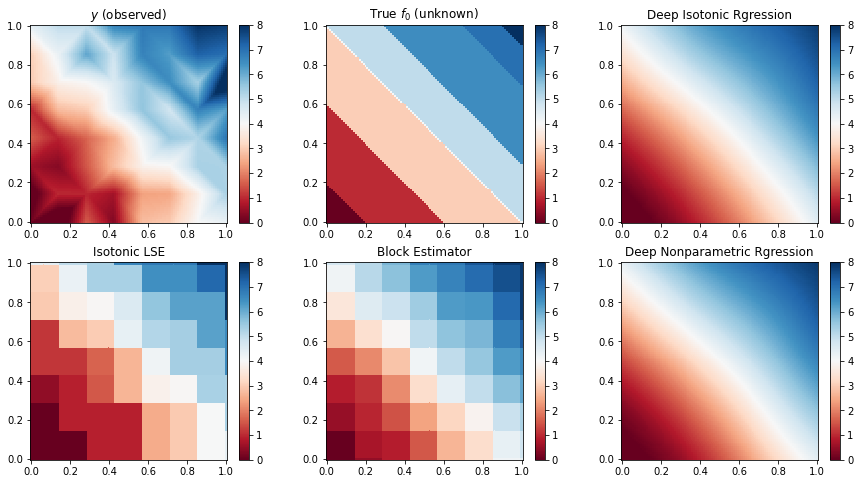}
	\caption{Heatmaps for the target function $f_0$, the observed training data, and its deep isotonic regression and isotonic least squares estimate (isotonic LSE) under model (c) when $d=2$ and $n=64$.}
	\label{fig:c}
\end{figure}

\begin{figure}[H]
	\includegraphics[width=\textwidth]{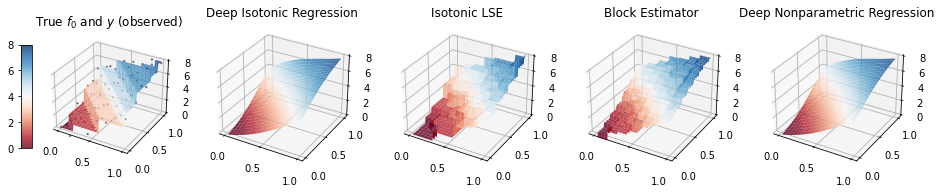}
	\caption{3D surface plots for the target function $f_0$, the observed training data, and its deep isotonic regression and isotonic least squares estimate (isotonic LSE) under model (c) when $d=2$ and $n=64$.}
	\label{fig:c1}
\end{figure}

\begin{figure}[H]
	\includegraphics[width=0.95\textwidth]{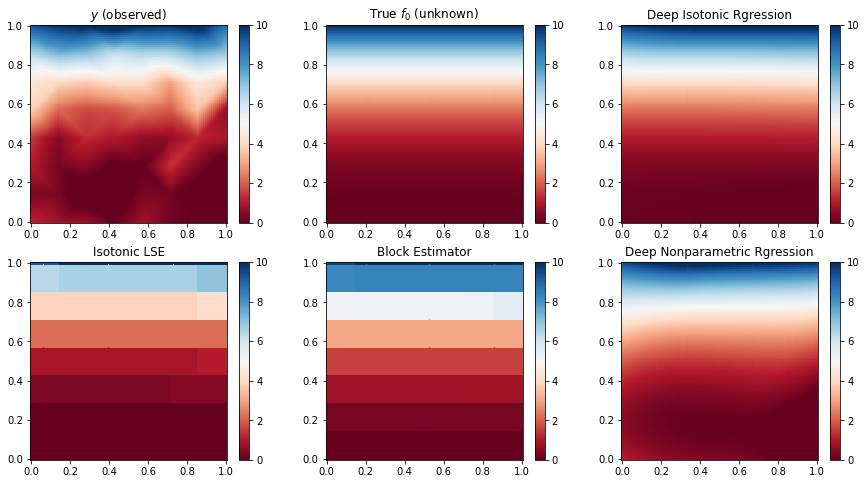}
	\caption{Heatmaps for the target function $f_0$, the observed training data, and its deep isotonic regression and isotonic least squares estimate (isotonic LSE) under model (d) when $d=2$ and $n=64$.}
	\label{fig:d}
\end{figure}

\begin{figure}[H]
	\includegraphics[width=\textwidth]{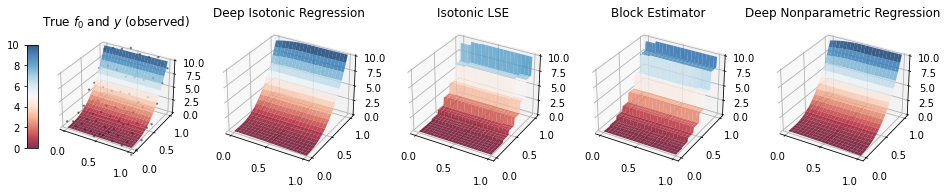}
	\caption{3D surface plots for the target function $f_0$, the observed training data, and its deep isotonic regression and isotonic least squares estimate (isotonic LSE) under model (d) when $d=2$ and $n=64$.}
	\label{fig:d1}
\end{figure}

\begin{figure}[H]
	\includegraphics[width=0.95\textwidth]{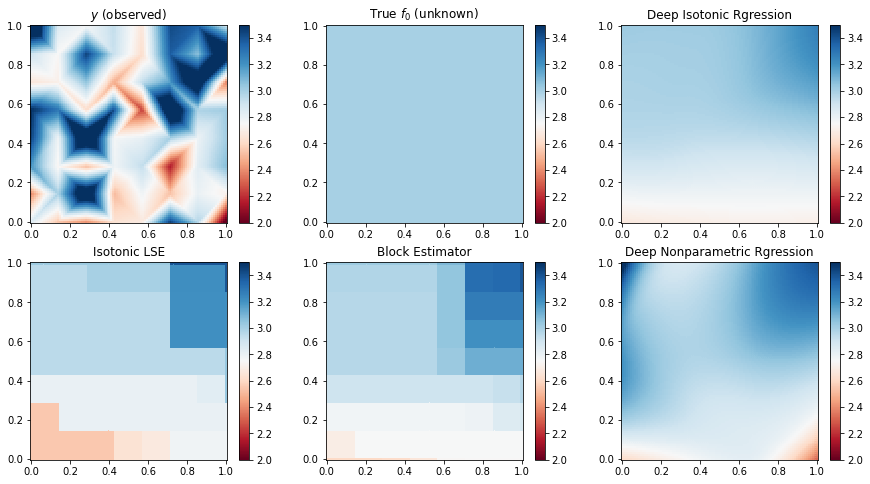}
	\caption{Heatmaps for the target function $f_0$, the observed training data, and its deep isotonic regression and isotonic least squares estimate (isotonic LSE) under model (e) when $d=2$ and $n=64$.}
	\label{fig:e}
\end{figure}

\begin{figure}[H]
	\includegraphics[width=\textwidth]{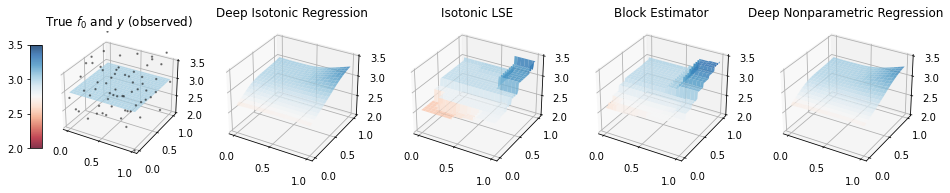}
	\caption{3D surface plots for the target function $f_0$, the observed training data, and its deep isotonic regression and isotonic least squares estimate (isotonic LSE) under model (e) when $d=2$ and $n=64$.}
	\label{fig:e1}
\end{figure}

\begin{figure}[H]
	\includegraphics[width=0.95\textwidth]{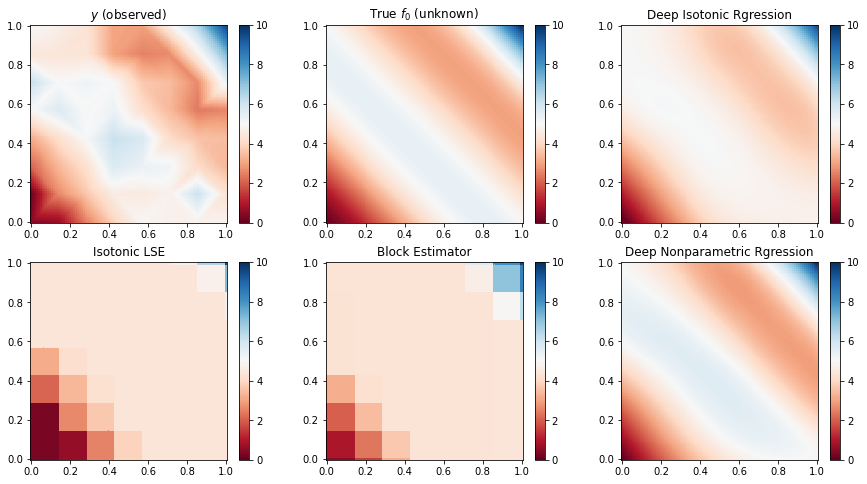}
	\caption{Heatmaps for the target function $f_0$, the observed training data, and its deep isotonic regression and isotonic least squares estimate (isotonic LSE) under model (f) when $d=2$ and $n=64$.}
	\label{fig:f}
\end{figure}

\begin{figure}[H]
	\includegraphics[width=\textwidth]{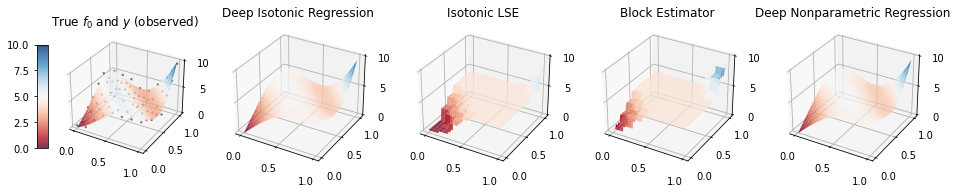}
	\caption{3D surface plots for the target function $f_0$, the observed training data, and its deep isotonic regression and isotonic least squares estimate (isotonic LSE) under model (f) when $d=2$ and $n=64$.}
	\label{fig:f1}
\end{figure}

\begin{itemize}
	\setlength\itemsep{0.05 cm}
	\item [(a)] Polynomial:$$Y=f_0(X)+\epsilon=\frac{10}{2^{3/4}}(x_1+x_2)^{3/4}+\epsilon,$$
	\item [(b)] Concave: $$Y=f_0(X)+\epsilon=1+3x_1(1-\exp(-3x_2))+\epsilon,$$
	\item [(c)] Step: $$Y=f_0(X)+\epsilon=\sum h_i I(x_1+x_2\ge t_i)+\epsilon,$$
	\item [(d)] Partial: $$Y=f_0(X)+\epsilon=10 x_2^{8/3}+\epsilon,$$
	\item [(e)] Constant: $$Y=f_0(X)+\epsilon=3+\epsilon,$$
	\item [(f)] Wave: $$Y=f_0(X)+\epsilon=5(x_1+x_2)+3(x_1+x_2)\sin(\pi (x_1+x_2))+\epsilon, $$
\end{itemize}
where $X=(x_1,x_2)$, $(h_i)=(1,2,2,1.5,0.5,1)$, $(t_i)=(0.2,0.6,1.0,1.3,1.7,1.9)$ and $\epsilon\sim N(0,\frac{1}{4})$ follows normal distribution. The ``Polynomial" and ``Concave" models are monotonic models. The ``Step" model is monotonic but not smooth even continuous. In ``Partial " model, the response is related to only one covariate. The ``Constant'' is a monotonic but not strictly monotonic model and the ``Wave'' is a nonlinear, smooth but non monotonic model. We use the lattice design for the ease of presentation of {\it Block estimator}, where $(X_i^{\rm train})_{i=1}^n$ are the lattice points evenly distributed on interval $[0,1]^2$. Simulation results over 100 replications are summarized in Table \ref{tab:2}. And for each model, we take an instance from the replications to present the heatmaps and the 3D surface of the predictions of these estimates; see Figure \ref{fig:a}-\ref{fig:f1}. In heatmaps, we show the observed data (linearly interpolated), the true target function $f_0$ and the estimates of different methods.  We can see that compared with the piece-wise constant estimates of {\it Isotonic LSE} and {\it Block estimator}, the DIR estimator is smooth and works reasonably well under bivariate models, especially for models with smooth target functions.

\begin{table}[H]
	\renewcommand\arraystretch{1.3}
	\centering
	\caption{Summary statistics for the simulation results under different bivariate models ($d=2$). The averaged mean squared error of the estimates on testing data and $L_1, L_2$ distance to the target function are calculated over 100 replications. The standard deviations are reported in parenthesis.}
	\label{tab:2}
	\resizebox{\textwidth}{!}{%
		\begin{tabular}{@{}cc|ccc|ccc@{}}
			\toprule
			\multirow{2}{*}{Model}      & \multirow{2}{*}{Method} & \multicolumn{3}{c|}{$n=64$}                                               & \multicolumn{3}{c}{$n=256$}                                              \\
			&                         & MSE                    & $L_1$                  & $L_2$                  & MSE                    & $L_1$                  & $L_2$                  \\ \midrule
			\multirow{4}{*}{Polynomial} & DNR                     & 4.735 (0.344)          & \textbf{0.138 (0.041)} & 0.172 (0.046)          & \textbf{4.655 (0.178)} & 0.078 (0.022)          & 0.098 (0.025)          \\
			& PDIR                     & \textbf{4.724 (0.366)} & 0.140 (0.045)          & \textbf{0.171 (0.049)} & 4.688 (0.181)          & \textbf{0.077 (0.023)} & \textbf{0.096 (0.026)} \\
			& Isotonic LSE            & 8.309 (0.405)          & 0.755 (0.061)          & 0.884 (0.061)          & 6.052 (0.153)          & 0.364 (0.019)          & 0.444 (0.021)          \\
			& Block                   & 4.780 (0.284)          & 0.319 (0.020)          & 0.397 (0.026)          & 4.747 (0.129)          & 0.210 (0.011)          & 0.264 (0.017)          \\\hline
			\multirow{4}{*}{Concave}    & DNR                     & 0.282 (0.016)          & 0.142 (0.038)          & 0.176 (0.042)          & 0.261 (0.007)          & 0.083 (0.022)          & 0.103 (0.024)          \\
			& PDIR                     & \textbf{0.276 (0.015)} & \textbf{0.129 (0.038)} & \textbf{0.158 (0.043)} & \textbf{0.260 (0.007)} & \textbf{0.077 (0.024)} & \textbf{0.096 (0.028)} \\
			& Isotonic LSE            & 0.393 (0.042)          & 0.308 (0.051)          & 0.375 (0.055)          & 0.294 (0.010)          & 0.163 (0.020)          & 0.207 (0.022)          \\
			& Block                   & 0.303 (0.015)          & 0.183 (0.025)          & 0.229 (0.030)          & 0.275 (0.006)          & 0.125 (0.012)          & 0.157 (0.014)          \\\hline
			\multirow{4}{*}{Step}       & DNR                     & 0.561 (0.030)          & 0.462 (0.020)          & 0.557 (0.025)          & 0.519 (0.011)          & 0.432 (0.010)          & 0.519 (0.010)          \\
			& PDIR                     & \textbf{0.561 (0.030)} & \textbf{0.461 (0.019)} & \textbf{0.557 (0.024)} & 0.519 (0.011)          & 0.431 (0.010)          & 0.520 (0.010)          \\
			& Isotonic LSE            & 1.462 (0.104)          & 0.852 (0.046)          & 1.100 (0.047)          & 0.700 (0.028)          & 0.430 (0.019)          & 0.672 (0.021)          \\
			& Block                   & 0.657 (0.031)          & 0.503 (0.022)          & 0.638 (0.023)          & \textbf{0.461 (0.016)} & \textbf{0.321 (0.014)} & \textbf{0.457 (0.014)} \\\hline
			\multirow{4}{*}{Partial}    & DNR                     & 12.67 (0.530)          & 0.129 (0.033)          & 0.161 (0.039)          & 12.77 (0.338)          & 0.079 (0.023)          & 0.099 (0.027)          \\
			& PDIR                     & 12.70 (0.548)          & \textbf{0.112 (0.037)} & \textbf{0.136 (0.042)} & 12.72 (0.285)          & \textbf{0.063 (0.021)} & \textbf{0.080 (0.026)} \\
			& Isotonic LSE            & 17.78 (0.532)          & 0.739 (0.052)          & 1.062 (0.055)          & 15.00 (0.262)          & 0.378 (0.024)          & 0.528 (0.028)          \\
			& Block                   & \textbf{12.31 (0.571)} & 0.435 (0.041)          & 0.578 (0.043)          & \textbf{12.50 (0.313)} & 0.237 (0.028)          & 0.313 (0.049)          \\\hline
			\multirow{4}{*}{Constant}   & DNR                     & 0.278 (0.016)          & 0.131 (0.038)          & 0.160 (0.042)          & 0.260 (0.005)          & 0.079 (0.020)          & 0.097 (0.022)          \\
			& PDIR                     & 0.266 (0.013)          & 0.094 (0.047)          & 0.111 (0.050)          & \textbf{0.255 (0.005)} & \textbf{0.052 (0.022)} & \textbf{0.063 (0.025)} \\
			& Isotonic LSE            & 0.280 (0.021)          & 0.121 (0.047)          & 0.161 (0.056)          & 0.262 (0.006)          & 0.076 (0.025)          & 0.108 (0.026)          \\
			& Block                   & \textbf{0.265 (0.012)} & \textbf{0.089 (0.040)} & \textbf{0.110 (0.046)} & 0.256 (0.005)          & 0.059 (0.022)          & 0.075 (0.024)          \\\hline
			\multirow{4}{*}{Wave}       & DNR                     & \textbf{0.306 (0.020)} & \textbf{0.189 (0.036)} & \textbf{0.233 (0.042)} & \textbf{0.269 (0.009)} & \textbf{0.108 (0.025)} & \textbf{0.135 (0.029)} \\
			& PDIR                     & 0.459 (0.058)          & 0.390 (0.056)          & 0.454 (0.063)          & 0.581 (0.039)          & 0.493 (0.028)          & 0.574 (0.033)          \\
			& Isotonic LSE            & 1.380 (0.085)          & 0.918 (0.032)          & 1.063 (0.040)          & 0.989 (0.024)          & 0.760 (0.014)          & 0.860 (0.012)          \\
			& Block                   & 0.978 (0.022)          & 0.750 (0.014)          & 0.854 (0.012)          & 0.892 (0.021)          & 0.693 (0.009)          & 0.802 (0.008)          \\ \bottomrule
		\end{tabular}%
	}
\end{table}

\subsection{Tuning parameter}
In this subsection, we investigate the numerical performance regrading to different choice of tuning parameters under different models.

\begin{figure}[H]
	\centering
	\includegraphics[width=0.48\textwidth]{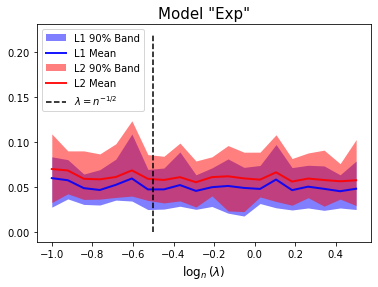}
	\includegraphics[width=0.465\textwidth]{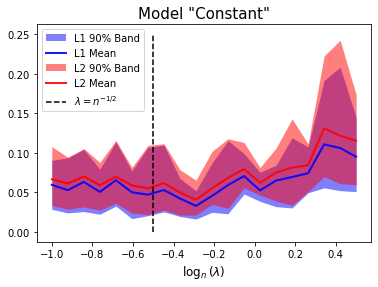}
	\includegraphics[width=0.465\textwidth]{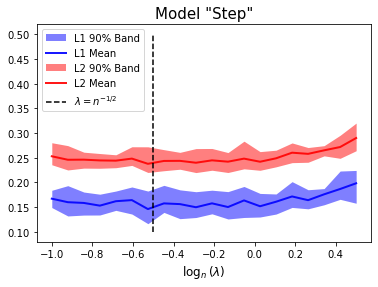}
	\includegraphics[width=0.465\textwidth]{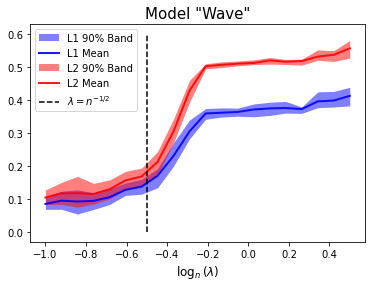}
	\caption{$L_1$ and $L_2$ distances between estimates and the targets with different tuning parameters under univariate models with size $n=256$. For each value of tuning parameter $\lambda$, the mean $L_1$ and $L_2$ distances (solid {\color{blue}blue} and {\color{red}red} curves) and their 90\% empirical band ({\color{blue}blue} and {\color{red}red} ranges) are calculated over 20 replications. A vertical dashed line is presented at $\lambda=\log(n)$.}
	\label{fig:tuning_1D}
\end{figure}

For univariate models, we calculate the testing statistics $L_1$ and $L_2$ for tuning parameter $\lambda$ on the 20 lattice points in the interval $[0,3\log(n)]$. For each $\lambda$, we simulate 20 times replications and reports the average $L_1$, $L_2$ statistics and their 90\% empirical band. For each replication, we train the PDIR using $n=256$ training samples and $T=1,000$ testing samples. In our simulation, four univariate models in section \ref{sec:univariate} are considered including ``Exp", ``Constant", ``Step" and misspecified model ``Wave" and the results are reported in Figure \ref{fig:tuning_1D}. We can see that for isotonic models ``Exp" and ``Step", the estimate is not sensitive to the choice of the tuning parameter $\lambda$ in $[0,3\log(n)]$, which all leads to reasonable estimates. For ``Constant" model, a not strictly isotonic model, errors slightly increase  as the tuning parameter $\lambda$ increases in $[0,3\log(n)]$. Overall, the choice $\lambda=\log(n)$ can lead to reasonably well estimates for correctly specified models. For misspecified model ``Wave", the estimates deteriorates quickly as the tuning parameter $\lambda$ increases around 0, and after that the additional negative effect of the increasing $\lambda$ becomes slight.

\begin{figure}[H]
	\centering
	\includegraphics[width=0.465\textwidth]{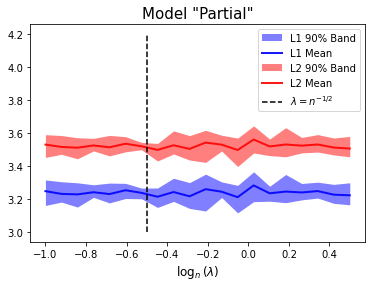}
	\includegraphics[width=0.465\textwidth]{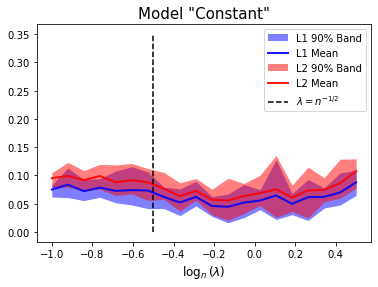}
	\includegraphics[width=0.465\textwidth]{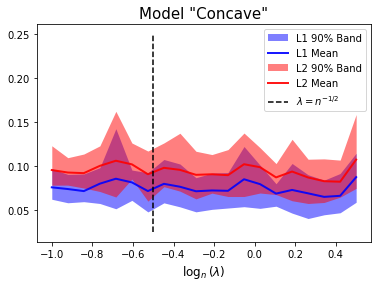}
	\includegraphics[width=0.465\textwidth]{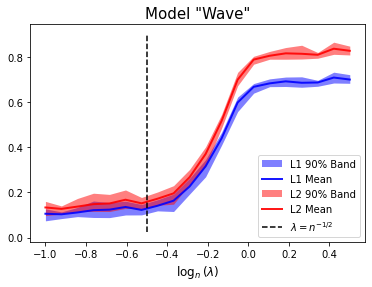}
	\caption{$L_1$ and $L_2$ distances between estimates and the targets with different tuning parameters under bivariate models with size $n=256$. For each value of tuning parameter $(\lambda_1,\lambda_2)$ with $\lambda_1=\lambda_2$, the mean $L_1$ and $L_2$ distances (solid {\color{blue}blue} and {\color{red}red} curves) and their 90\% empirical band ({\color{blue}blue} and {\color{red}red} ranges) are calculated over 20 replications. A vertical dashed line is presented at $\lambda_1=\lambda_2=\log(n)$.}
	\label{fig:tuning_2D}
\end{figure}

For bivariate models, our first simulation studies focus on the case where tuning parameters have the same value, i.e., $\lambda_1=\lambda_2$. We calculate the testing statistics $L_1$ and $L_2$ for tuning parameters $\lambda_1=\lambda_2$ on the 20 lattice points in the interval $[0,3\log(n)]$.  For each $\lambda$, we simulate 20 times replications and reports the average $L_1$, $L_2$ statistics and their 90\% empirical band. For each replication, we train the PDIR using $n=256$ training samples and $T=10,000$ testing samples. In our simulation, four bivariate models in section \ref{sec:bivariate} are considered including ``Partial", ``Constant", ``Concave" and misspecified model ``Wave" and the results are reported in Figure \ref{fig:tuning_2D}. The observation are similar to those of univariate models, that is, the estimates are not sensitive to the choices of tuning parameters over $[0,3\log(n)]$ for correctly specified models, i.e., isotonic models.  But for misspecified model ``Wave", the estimates deteriorates quickly as the tuning parameter $\lambda$ increases around 0, and after that increasing $\lambda$ slightly spoils the estimates.

\begin{figure}[H]
	\centering
	\includegraphics[width=0.95\textwidth]{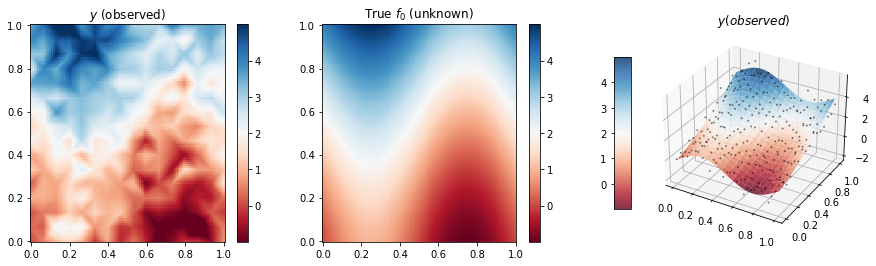}
	\caption{Heatmaps for the observed training data, the target function $f_0$, and 3D surface plots for the target function $f_0$ under model (g) when $d=2$ and $n=256$.}
	\label{fig:g}
\end{figure}

In the second simulation study of bivariate models, we can choose different values for different components of the turning parameter $\lambda=(\lambda_1,\lambda_2)$, i.e.,  $\lambda_1$ can be different from  $\lambda_2$. We investigate this by considering the follow bivariate model, where the target function $f_0$ is monotonic in its second argument and non-monotonic in its first one.
\begin{itemize}
	\item [Model (g)]:
	$$Y=f_0(X)+\epsilon=2\sin(2\pi x_1)+4(x_2)^{4/3}+\epsilon,$$
\end{itemize}
where $X=(x_1,x_2)$ and $\epsilon\sim N(0,\frac{1}{4})$ follows normal distribution. Heatmaps for the observed training data, the target function $f_0$, and 3D surface plots for the target function $f_0$ under model (g) when $d=2$ and $n=256$ are presented in Figure \ref{fig:g}.

\begin{figure}[H]
	\centering
	\includegraphics[width=0.8\textwidth]{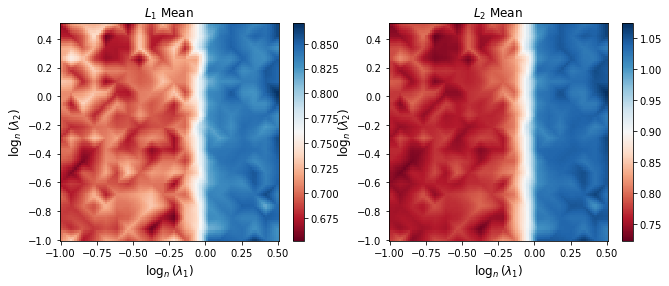}
	\caption{Heatmaps for the observed training data, the target function $f_0$, and 3D surface plots for the target function $f_0$ under model (g) when $d=2$ and $n=256$.}
	\label{fig:g1}
\end{figure}

For model (g), we calculate the mean testing statistics $L_1$ and $L_2$ for tuning parameter $\lambda=(\lambda_1,\lambda_2)$ on the 400 grid points on the region $[0,3\log(n)]\times[0,3\log(n)]$. For each $\lambda=(\lambda_1,\lambda_2)$, we simulate 5 times replications and reports the average $L_1$, $L_2$ statistics. For each replication, we train the PDIR using $n=256$ training samples and $T=10,000$ testing samples. The mean $L_1$ and $L_2$ distance from the target function and the estimates on the testing data under different $\lambda$ are depicted in Figure \ref{fig:g1}. We see that the target function $f_0$ is increasing in its second argument, and the estimates is insensitive to the tuning parameter $\lambda_2$, while the target function $f_0$ is non-monotonic in its first argument, and the estimates deteriorate when $\lambda_1$ gets larger. The simulation results suggest that we should only penalize the gradient with respect to the monotonic arguments but not non-monotonic ones. The estimates is sensitive to the turning parameter for misspecified model, especially when the turning parameter increases around $0$.

\section{Proofs}
\label{appendix_b}

\subsection*{Proof of Theorem \ref{thm2}}

For integer $p\ge2$, let $\sigma_{p-1}(x)=\max\{0,x\}^{p-1}$ and $\sigma_p(x)=\max\{0,x\}^p$ denote the RePUs activation functions respectively. Let $(d_0,d_1,\ldots,d_{\mathcal{D}+1})$ be vector of the width (number of neurons) of each layer in the original RePU network where $d_0=d$ and $d_{\mathcal{D}+1}=1$ in our problem. We let $f^{(i)}_u$ be the function (subnetwork of the RePU network) from $\mathcal{X}\subset\mathbb{R}^{d}$ to $\mathbb{R}$ which takes $X=(x_1,\ldots,x_d)$ as input and outputs the $u$-th neuron of the $i$-th layer for  $u=1,\ldots,d_i$ and $i=1,\ldots,\mathcal{D}+1$.

We construct Mixed RePUs activated subnetworks to compute $(\pj f^{(i)}_1,\ldots,\pj f^{(i)}_{d_i})$ iteratively for $i=1,\ldots,\mathcal{D}+1$, i.e., we construct the partial derivatives of the original RePU subnetworks(up to $i$-th layer) step by step. Without loss of generality, we compute $(\pj f^{(i)}_1,\ldots,\pj f^{(i)}_{d_i})$ for $j\in\{1,\ldots,d\}$, and the construction is the same for all other $j\in\{1,\ldots,d\}$. We illustrate the details of the construction of the Mixed RePUs subnetworks for the first two layers ($i=1,2$) and the last layer $(i=\mathcal{D}+1)$ and apply induction for layers $i=3,\ldots,\mathcal{D}$.
Note that the derivative of RePU activation function $\sigma_p$ is $\sigma^\prime_p(x)=p\sigma_{p-1}(x)$, then when $i=1$ for any $u=1,\ldots,d_1$,
\begin{align}\label{pderlayer1}
	\pj f^{(1)}_u=\pj \sigma_p\Big(\sum_{i=1}^{d_0}w^{(1)}_{ui}x_i+b_u^{(1)}\Big)=p\sigma_{p-1}\Big(\sum_{i=1}^{d_0}w^{(1)}_{ui}x_i+b_u^{(1)}\Big)\cdot w_{u,d_0}^{(1)},
\end{align}
where we denote $w^{(1)}_{ui}$ and $b_u^{(1)}$ by the corresponding weights and bias in $1$-th layer of the original RePU network. Now we intend to construct a 4 layer (2 hidden layers) Mixed RePUs network with width $(d_0,3d_1,10d_1,2d_1)$ which takes $X=(x_1,\ldots,x_{d_0})$ as input and outputs
$$(f^{(1)}_1,\ldots,f^{(1)}_{d_1},\pj f^{(1)}_1,\ldots,\pj f^{(1)}_{d_1})\in\mathbb{R}^{2d_1}.$$
Note that the output of such network contains all the quantities needed to calculated  $(\pj f^{(2)}_1,\ldots,\pj f^{(2)}_{d_2})$, and the process of construction  can be continued iteratively and the induction proceeds. In the first hidden layer, we can obtain $3d_1$ neurons $$(f^{(1)}_1,\ldots,f^{(1)}_{d_1},p\vert w^{(1)}_{1,d_0}\vert ,\ldots,p\vert w^{(1)}_{d_1,d_0}\vert ,\sigma_{p-1}(\sum_{i=1}^{d_0}w^{(1)}_{1i}x_i+b^{(1)}_1),\ldots,\sigma_{p-1}(\sum_{i=1}^{d_0}w^{(1)}_{d_1i}x_i+b^{(1)}_{d_1})),$$
with weight matrix $A^{(1)}_1$ having $2d_0d_1$ parameters, bias vector $B^{(1)}_1$ and activation function vector $\Sigma_1$ being
\begin{align*}A^{(1)}_1=\left[
	\begin{array}{ccccc}
		w^{(1)}_{1,1} &w^{(1)}_{1,2} & \cdots& \cdots&w^{(1)}_{1,d_0} \\
		w^{(1)}_{2,1} &w^{(1)}_{2,2} &\cdots& \cdots&w^{(1)}_{2,d_0} \\
		\ldots& \ldots&\ldots& \ldots&\ldots\\
		w^{(1)}_{d_1,1} &w^{(1)}_{d_1,2} &\cdots&\cdots&w^{(1)}_{d_1,d_0} \\
		0 & 0& 0&0 & 0\\
		\ldots& \ldots&\ldots& \ldots&\ldots\\
		0 & 0& 0&0 & 0\\
		w^{(1)}_{1,1} &w^{(1)}_{1,2} & \cdots& \cdots&w^{(1)}_{1,d_0} \\
		w^{(1)}_{2,1} &w^{(1)}_{2,2} &\cdots& \cdots&w^{(1)}_{2,d_0} \\
		\ldots& \ldots&\ldots& \ldots&\ldots\\
		w^{(1)}_{d_1,1} &w^{(1)}_{d_1,2} &\cdots&\cdots&w^{(1)}_{d_1,d_0} \\
	\end{array}\right]\in\mathbb{R}^{3d_1\times d_0},
\end{align*}
\begin{align*}
B^{(1)}_1=\left[\begin{array}{c}
		b^{(1)}_1\\
		b^{(1)}_2\\
		\ldots\\
		b^{(1)}_{d_1}\\
		p\vert w^{(1)}_{1,d_0}\vert \\
		p\vert w^{(1)}_{2,d_0}\vert \\
		\ldots\\
		p\vert w^{(1)}_{d_1,d_0}\vert \\
		b^{(1)}_1\\
		b^{(1)}_2\\
		\ldots\\
		b^{(1)}_{d_1}\\
	\end{array}\right]\in\mathbb{R}^{3d_1},\quad\Sigma^{(1)}_1=\left[\begin{array}{c}
		\sigma_p\\
		\ldots\\
		\sigma_p\\
		\sigma_{1}\\
		\ldots\\
		\sigma_{1}\\
		\sigma_{p-1}\\
		\ldots\\
		\sigma_{p-1}\\
	\end{array}\right],
\end{align*}
where the first $d_1$ activation functions of $\Sigma_1$ are chosen to be $\sigma_{p}$, the last $d_1$ activation functions are chosen to be $\sigma_{p-1}$ and the rest $\sigma_{1}$.
In the second hidden layer, we can obtain $6d_1$ neurons. The first $2d_1$ neurons of the second hidden layer (or the third layer) are
$$(\sigma_{1}(f^{(1)}_1),\sigma_{1}(-f^{(1)}_1)),\ldots,\sigma_{1}(f^{(1)}_{d_1}),\sigma_{1}(-f^{(1)}_{d_1})),$$
which intends to implement identity map such that $(f^{(1)}_1,\ldots,f^{(1)}_{d_1})$ can be kept and outputted in the next layer since identity map can be realized by $x=\sigma_1(x)-\sigma_1(-x)$. The rest $4d_1$ neurons of the second hidden layer (or the third layer) are
\begin{align*}
	\left[\begin{array}{c}
		\sigma_2(p\cdot w^{(1)}_{1,d_0}+\sigma_{p-1}(\sum_{i=1}^{d_0}w^{(1)}_{1i}x_i+b^{(1)}_{1}))\\
		\sigma_2(p\cdot w^{(1)}_{1,d_0}-\sigma_{p-1}(\sum_{i=1}^{d_0}w^{(1)}_{1i}x_i+b^{(1)}_{1}))\\
		\sigma_2(-p\cdot w^{(1)}_{1,d_0}+\sigma_{p-1}(\sum_{i=1}^{d_0}w^{(1)}_{1i}x_i+b^{(1)}_{1})\\
		\sigma_2(-p\cdot w^{(1)}_{1,d_0}-\sigma_{p-1}(\sum_{i=1}^{d_0}w^{(1)}_{1i}x_i+b^{(1)}_{1}))\\
		\ldots\\
		\sigma_2(p\cdot w^{(1)}_{d_1,d_0}+\sigma_{p-1}(\sum_{i=1}^{d_0}w^{(1)}_{d_1i}x_i+b^{(1)}_{d_1}))\\
		\sigma_2(p\cdot w^{(1)}_{d_1,d_0}-\sigma_{p-1}(\sum_{i=1}^{d_0}w^{(1)}_{d_1i}x_i+b^{(1)}_{d_1}))\\
		\sigma_2(-p\cdot w^{(1)}_{d_1,d_0}+\sigma_{p-1}(\sum_{i=1}^{d_0}w^{(1)}_{d_1i}x_i+b^{(1)}_{d_1})\\
		\sigma_2(-p\cdot w^{(1)}_{d_1,d_0}-\sigma_{p-1}(\sum_{i=1}^{d_0}w^{(1)}_{d_1i}x_i+b^{(1)}_{d_1}))\\
	\end{array}\right]\in\mathbb{R}^{4d_1},
\end{align*}
which is ready for implementing the multiplications in (\ref{pderlayer1}) to obtain $(\pj f^{(1)}_1,\ldots,\pj f^{(1)}_{d_1})\in\mathbb{R}^{d_1}$ since
\begin{align*}
	x\cdot y=\frac{1}{4}\{(x+y)^2-(x-y)^2\}=\frac{1}{4}\{\sigma_2(x+y)+\sigma_2(-x-y)-\sigma_2(x-y)-\sigma_2(-x+y)\}.
\end{align*}
In the second hidden layer (the third layer), the bias vector is zero $B^{(1)}_2=(0,\ldots,0)\in\mathbb{R}^{6d_1}$, activation functions vector $$\Sigma^{(1)}_2=(\underbrace{\sigma_1,\ldots,\sigma_1}_{2d_1\ {\rm times}},\underbrace{\sigma_2,\ldots,\sigma_2}_{4d_1\ {\rm times}}),$$
and the corresponding weight matrix $A^{(1)}_2$  can be formulated correspondingly without difficulty which contains $2d_1+8d_1=10d_1$ non-zero parameters. Then in the last layer, by the identity maps and multiplication operations with weight matrix $A^{(1)}_3$ having $2d_1+4d_1=6d_1$ parameters, bias vector $B^{(1)}_3$ being zeros, we obtain
$$(f^{(1)}_1,\ldots,f^{(1)}_{d_1},\pj f^{(1)}_1,\ldots,\pj f^{(1)}_{d_1})\in\mathbb{R}^{2d_1}.$$
Such Mixed RePUs neural network has 2 hidden layers (4 layers), $11d_1$ hidden neurons,
$2d_0d_1+3d_1+10d_1+6d_1=2d_0d_1+19d_1$ parameters and its width is $(d_0,3d_1,6d_1,2d_1)$. It worth noting that the RePU activation functions do not apply to the last layer since the construction here is for a single network.  When we are combining two consecutive subnetworks into one deep neural network, the RePU activation functions should apply to the last layer of the first subnetwork. Hence, in the construction of the whole big network, the last layer of the subnetwork here should output $4d_1$ neurons
\begin{align*}
	&(\sigma_1(f^{(1)}_1),\sigma_1(-f^{(1)}_1)\ldots,\sigma_1(f^{(1)}_{d_1}),\sigma_1(-f^{(1)}_{d_1}),\\
	&\qquad\sigma_1(\pj f^{(1)}_1),\sigma_1(-\pj f^{(1)}_1)\ldots,\sigma_1(\pj f^{(1)}_{d_1}),\sigma_1(-\pj f^{(1)}_{d_1}))\in\mathbb{R}^{4d_1},
\end{align*}
to keep  $(f^{(1)}_1,\ldots,f^{(1)}_{d_1},\pj f^{(1)}_1,\ldots,\pj f^{(1)}_{d_1})$ in use in the next subnetwork. Then for this Mixed RePUs neural network, the weight matrix $A^{(1)}_3$ has $2d_1+8d_1=10d_1$ parameters, the bias vector $B^{(1)}_3$ is zeros and the activation functions vector $\Sigma^{(1)}_3$ has all $\sigma_1$ as elements. And such Mixed RePUs neural network has 2 hidden layers (4 layers), $13d_1$ hidden neurons,
$2d_0d_1+3d_1+10d_1+10d_1=2d_0d_1+23d_1$ parameters and its width is $(d_0,3d_1,6d_1,4d_1)$.

Now we consider the second step, for any $u=1,\ldots,d_2$,
\begin{align}\label{pderlayer2}
	\pj f^{(2)}_u=\pj \sigma_p\Big(\sum_{i=1}^{d_1}w^{(2)}_{ui}f^{(1)}_{i}+b_u^{(2)}\Big)=p\sigma_{p-1}\Big(\sum_{i=1}^{d_1}w^{(2)}_{ui}f^{(1)}_i+b_u^{(2)}\Big)\cdot \sum_{i=1}^{d_1}w_{u,i}^{(2)}\pj f^{(1)}_i,
\end{align}
where $w^{(2)}_{ui}$ and $b_u^{(2)}$ are defined correspondingly as the weights and bias in $2$-th layer of the original RePU network.
By the previous constructed subnetwork, we can start with its outputs
\begin{align*}
	&(\sigma_1(f^{(1)}_1),\sigma_1(-f^{(1)}_1)\ldots,\sigma_1(f^{(1)}_{d_1}),\sigma_1(-f^{(1)}_{d_1}),\\
	&\qquad\sigma_1(\pj f^{(1)}_1),\sigma_1(-\pj f^{(1)}_1)\ldots,\sigma_1(\pj f^{(1)}_{d_1}),\sigma_1(-\pj f^{(1)}_{d_1}))\in\mathbb{R}^{4d_1},
\end{align*}
as the inputs of the second subnetwork we are going to build.  In the first hidden layer of the second subnetwork, we can obtain $3d_2$ neurons
\begin{align*}
	&\Big(f^{(2)}_1,\ldots,f^{(2)}_{d_2},\vert \sum_{i=1}^{d_1}w_{1,i}^{(2)}\pj f^{(1)}_i\vert ,\ldots,\vert \sum_{i=1}^{d_1}w_{d_2,i}^{(2)}\pj f^{(1)}_i\vert,\\
	&\qquad\qquad\sigma_{p-1}(\sum_{i=1}^{d_1}w^{(2)}_{1i}f^{(1)}_i+b^{(1)}_1),\ldots,\sigma_{p-1}(\sum_{i=1}^{d_1}w^{(2)}_{d_2i}f^{(1)}_i+b^{(2)}_{d_2})\Big),
\end{align*}
with weight matrix $A^{(2)}_1\in\mathbb{R}^{3d_2\times4d_1}$ having $6d_1d_2$ non-zero parameters, bias vector $B^{(2)}_1\in\mathbb{R}^{3d_2}$ and activation functions vector $\Sigma^{(2)}_1=\Sigma^{(1)}_1$.
Similarly, the second hidden layer can be constructed to have $6d_2$ neurons with  weight matrix $A^{(2)}_2\in\mathbb{R}^{3d_2\times6d_2}$ having $2d_2+8d_2=10d_2$ non-zero parameters, zero bias vector $B^{(2)}_1\in\mathbb{R}^{6d_2}$ and activation functions vector $\Sigma^{(2)}_2=\Sigma^{(1)}_2$. The second hidden layer here serves exactly the same as that in the first subnetwork, which intends to implement the identity map for
$$(f^{(2)}_1,\ldots,f^{(2)}_{d_2}),$$
and implement the multiplication in (\ref{pderlayer2}). Similarly, the last layer can also be constructed as that in the first subnetwork, which outputs
\begin{align*}
	&(\sigma_1(f^{(2)}_1),\sigma_1(-f^{(2)}_1)\ldots,\sigma_1(f^{(2)}_{d_2}),\sigma_1(-f^{(2)}_{d_2}),\\
	&\qquad\sigma_1(\pj f^{(2)}_1),\sigma_1(-\pj f^{(2)}_1)\ldots,\sigma_1(\pj f^{(2)}_{d_2}),\sigma_1(-\pj f^{(2)}_{d_2}))\in\mathbb{R}^{4d_2},
\end{align*}
with the weight matrix $A^{(2)}_3$ having $2d_2+8d_2=10d_2$ parameters, the bias vector $B^{(2)}_3$ being zeros and the activation functions vector $\Sigma^{(1)}_3$ with elements being  $\sigma_1$. Then the second Mixed RePUs  subnetwork has 2 hidden layers (4 layers), $17d_2$ hidden neurons,
$6d_1d_2+3d_2+10d_2+10d_2=6d_1d_2+23d_2$ parameters and its width is $(4d_1,3d_2,6d_2,4d_2)$.

Then we can continuing this process of construction. For integers $k=3,\ldots,\mathcal{D}$ and for any $u=1,\ldots,d_{k}$,
\begin{align*}
	\pj f^{(k)}_u&=\pj \sigma_p\Big(\sum_{i=1}^{d_{k-1}}w^{(k)}_{ui}f^{(k-1)}_{i}+b_u^{(k)}\Big)\\
	&=p\sigma_{p-1}\Big(\sum_{i=1}^{d_{k-1}}w^{(k)}_{ui}f^{(k-1)}_i+b_u^{(k)}\Big)\cdot \sum_{i=1}^{d_{k-1}}w_{u,i}^{(k)}\pj f^{(k-1)}_i,
\end{align*}
where $w^{(k)}_{ui}$ and $b_u^{(k)}$ are defined correspondingly as the weights and bias in $k$-th layer of the original RePU network. We can construct a Mixed RePUs network taking
\begin{align*}
	&(\sigma_1(f^{(k-1)}_1),\sigma_1(-f^{(k-1)}_1)\ldots,\sigma_1(f^{(k-1)}_{d_{k-1}}),\sigma_1(-f^{(k-1)}_{d_{k-1}}),\\
	&\qquad\sigma_1(\pj f^{(k-1)}_1),\sigma_1(-\pj f^{(k-1)}_1)\ldots,\sigma_1(\pj f^{(k-1)}_{d_{k-1}}),\sigma_1(-\pj f^{(k-1)}_{d_{k-1}}))\in\mathbb{R}^{4d_{k-1}},
\end{align*}
as input, and it outputs
\begin{align*}
	&(\sigma_1(f^{(k)}_1),\sigma_1(-f^{(k)}_1)\ldots,\sigma_1(f^{(k)}_{d_{k}}),\sigma_1(-f^{(k)}_{d_{k}}),\\
	&\qquad\sigma_1(\pj f^{(k)}_1),\sigma_1(-\pj f^{(k)}_1)\ldots,\sigma_1(\pj f^{(k)}_{d_{k}}),\sigma_1(-\pj f^{(k)}_{d_{k}}))\in\mathbb{R}^{4d_{k}},
\end{align*}
with 2 hidden layers, $13d_k$ hidden neurons, $6d_{k-1}d_k+23d_k$ parameters and its width is $(4d_{k-1},3d_{k},6d_{k},4d_k)$.

Iterate this process until the $k=\mathcal{D}+1$ step, where the last layer of the original RePU network has only $1$ neurons. For the RePU activated neural network $f\in\mathcal{F}_n=\mathcal{F}_{\mathcal{D},\mathcal{W},\mathcal{U},\mathcal{S},\mathcal{B}}$, the output of the network $f:\mathcal{X}\to\mathbb{R}$ is a scalar and the partial derivative with respect to $x_j$ is
\begin{align*}
	\pj f=\pj \sum_{i=1}^{d_{\mathcal{D}}}w^{(\mathcal{D})}_{i}f^{(\mathcal{D})}_i+b^{(\mathcal{D})}=\sum_{i=1}^{d_{\mathcal{D}}}w^{(\mathcal{D})}_{i} \pj f^{(\mathcal{D})}_i,
\end{align*}
where $w^{(\mathcal{D})}_{i}$ and $b^{(\mathcal{D})}$ are the weights and bias parameter in the last layer of the original RePU network.
The the constructed $(\mathcal{D}+1)$-th subnetwork taking
\begin{align*}
	&(\sigma_1(f^{(\mathcal{D})}_1),\sigma_1(-f^{(\mathcal{D})}_1)\ldots,\sigma_1(f^{(\mathcal{D})}_{d_{\mathcal{D}}}),\sigma_1(-f^{(\mathcal{D})}_{d_{\mathcal{D}}}),\\
	&\qquad\sigma_1(\pj f^{(\mathcal{D})}_1),\sigma_1(-\pj f^{(\mathcal{D})}_1)\ldots,\sigma_1(\pj f^{(\mathcal{D})}_{d_{\mathcal{D}}}),\sigma_1(-\pj f^{(\mathcal{D})}_{d_{\mathcal{D}}}))\in\mathbb{R}^{4d_{\mathcal{D}}},
\end{align*}
as input and  it outputs $\pj f^{(\mathcal{D}+1)}=\pj f$ which is the partial derivative of the whole RePU network with respect to its $j$-th argument $x_j$. The subnetwork should have 2 hidden layers width $(4d_{\mathcal{D}},2,8,1)$ with $11$ hidden neurons, $4d_{\mathcal{D}}+2+16=4d_{\mathcal{D}}+18$ non-zero parameters.

Lastly, we combing all the $\mathcal{D}+1$ subnetworks in order to form a big Mixed RePUs network which takes $X=(x_1,\ldots,x_{d})\in\mathbb{R}^{d}$ as input and outputs $\pj f$ for $f\in\mathcal{F}_n=\mathcal{F}_{\mathcal{D},\mathcal{W}, \mathcal{U},\mathcal{S},\mathcal{B},\mathcal{B}^\prime}$. Recall that here $\mathcal{D},\mathcal{W}, \mathcal{U},\mathcal{S}$ are the depth, width, number of neurons and number of parameters of the original RePU network respectively, and we have $\mathcal{U}=\sum_{i=0}^{\mathcal{D}+1}d_i$ and $\mathcal{S}=\sum_{i=0}^{\mathcal{D}}d_id_{i+1}+d_{i+1}.$ Then the big Mixed RePUs network has $3\mathcal{D}+3$ hidden layers (totally 3$\mathcal{D}+5$ layers), $d_0+\sum_{i=1}^{\mathcal{D}}13d_{i}+11\le 13\mathcal{U}$ neurons, $2d_0d_1+23d_1+\sum_{i=1}^\mathcal{D}(6d_{i}d_{i+1}+23d_{i+1})+4d_\mathcal{D}+18\le23\mathcal{S}$ parameters and its width is $6\max\{d_1,\ldots,d_\mathcal{D}\}=6\mathcal{W}$. This completes the proof. $\hfill \Box$

\subsection*{Proof of Lemma \ref{lemmapdim}}
We follow the idea of the proof of Theorem 6 in \citet{bartlett2019nearly} to prove a somewhat stronger result, where we give the upper bound of the Pseudo dimension of class of Mixed RePUs networks $\mathcal{F}$ in terms of the depth, size and number of neurons of the network. Instead of the VC dimension of ${\rm sign}(\mathcal{F})$ given in \citet{bartlett2019nearly}, our Pseudo dimension  bound is stronger since ${\rm VCdim}({\rm sign}(\mathcal{F}))\le\pdim(\mathcal{F})$.

Let $\mathcal{Z}$ denote the domain of the functions $f\in\mathcal{F}$ and let $t\in\mathbb{R}$, we consider a new class of functions
$$\tilde{\mathcal{F}}:=\{\tilde{f}(z,t)=\s(f(z)-t):f\in\mathcal{F}\}.$$
Then it is clear that $\pdim(\mathcal{F})\le\vdim(\tilde{\mathcal{F}})$ and we next bound the VC dimension of $\tilde{\mathcal{F}}$.
Recall that the the total number of parameters (weights and biases) in the neural network implementing functions in $\mathcal{F}$ is $\mathcal{S}$, we let $\theta\in\mathbb{R}^{\mathcal{S}}$ denote the parameters vector of the network $f(\cdot,\theta):\mathcal{Z}\to\mathbb{R}$ implemented in $\mathcal{F}$. And here we intend to derive a bound for
$$K(m):=\Big\vert\{(\s(f(z_1,\theta)-t_1),\ldots,\s(f(z_m,\theta)-t_m)):\theta\in\mathbb{R}^{\mathcal{S}}\}\Big\vert$$
which uniformly holds for all choice of $\{z_i\}_{i=1}^m$ and $\{t_i\}_{i=1}^m$. Note that the maximum of $K(m)$ over all all choice of $\{z_i\}_{i=1}^m$ and $\{t_i\}_{i=1}^m$ is just the growth function of $\tilde{\mathcal{F}}$. To give a uniform bound of $K(m)$, we use the Theorem 8.3 in \citet{anthony1999} as a main tool to deal with the analysis.
\begin{lemma}[Theorem 8.3 in \citet{anthony1999}]\label{polynum}
	Let $p_1,\ldots,p_m$ be polynomials in $n$ variables of degree at most $d$. If $n\le m$, define
	$$K:=\vert\{(\s(p_1(x),\ldots,\s(p_m(x))):x\in\mathbb{R}^n)\}\vert,$$
	i.e. $K$ is the number of possible sign vectors given by the polynomials. Then $K\le2(2emd/n)^n$.
\end{lemma}

Now if we can find a partition $\mathcal{P}=\{P_1,\ldots,P_N\}$ of the parameter domain $\mathbb{R}^{\mathcal{S}}$ such that within each region $P_i$, the functions $f(z_j,\cdot)$ are all fixed polynomials of bounded degree, then $K(m)$ can be bounded via the following sum
\begin{align}\label{Km}
	K(m)\le\sum_{i=1}^N\Big\vert\{(\s(f(z_1,\theta)-t_1),\ldots,\s(f(z_m,\theta)-t_m)):\theta\in P_i\}\Big\vert,
\end{align}
and each term in this sum can be bounded via Lemma \ref{polynum}.
Next, we construct the partition follows the same way as in \citet{bartlett2019nearly} iteratively layer by layer. We define the a sequence of successive refinements $\mathcal{P}_1,\ldots,\mathcal{P}_{\mathcal{D}}$ satisfying the following properties:
\begin{itemize}
	\item [1.] The cardinality $\vert\mathcal{P}_1\vert=1$ and for each $n\in\{1,\ldots,\mathcal{D}\}$,
	$$\frac{\vert\mathcal{P}_{n+1}\vert}{\vert\mathcal{P}_{n}\vert}\le2\Big(\frac{2emk_n(1+(n-1)p^{n-1})}{\mathcal{S}_n}\Big)^{\mathcal{S}_n},$$
	where $k_n$ denotes the number of neurons in the $n$-th layer and $\mathcal{S}_n$ denotes the total number of parameters (weights and biases) at the inputs to units in all the layers up to layer $n$.
	\item [2.] For each $n\in\{1,\ldots,\mathcal{D}\}$, each element of $P$ of $\mathcal{P}_n$, each $j\in\{1,\ldots,m\}$, and each unit $u$ in the $n$-th layer, when $\theta$ varies in $P$, the net input to $u$ is a fixed polynomial function in $\mathcal{S}_n$ variables of $\theta$, of total degree no more than $1+(n-1)p^{n-1}$ where for each layer the activation functions are  $\sigma_{1},\ldots,\sigma_{p}$ for some integer $p\ge2$ (this polynomial may depend on $P,j$ and $u$).
\end{itemize}
One can define $\mathcal{P}_1=\mathbb{R}^\mathcal{S}$, and it can be verified that $\mathcal{P}_1$ satisfies property 2 above. Note that in our case, for fixed $z_j$ and $t_j$ and any subset $P\subset\mathbb{R}^{\mathcal{S}}$,  $f(z_j,\theta)-t_j$ is a polynomial with respect to $\theta$ with degree the same as that of $f(z_j,\theta)$, which is no more than $1+(\mathcal{D}-1)p^{\mathcal{D}-1}$. Then the construction of $\mathcal{P}_1,\ldots,\mathcal{P}_{\mathcal{D}}$ and its verification for properties 1 and 2 can follow the same way in \citet{bartlett2019nearly}. Finally we obtain a partition $\mathcal{P}_{\mathcal{D}}$ of $\mathbb{R}^\mathcal{S}$ such that for $P\in\mathcal{P}_\mathcal{D}$, the network output in response to any $z_j$ is a fixed polynomial of $\theta\in P$ of degree no more than $1+(\mathcal{D}-1)p^{\mathcal{D}-1}$ (since the last node just outputs its input). Then by Lemma \ref{polynum}
$$\Big\vert\{(\s(f(z_1,\theta)-t_1),\ldots,\s(f(z_m,\theta)-t_m)):\theta\in P\}\Big\vert\le2\Big(\frac{2em(1+(\mathcal{D}-1)p^{\mathcal{D}-1})}{\mathcal{S}_\mathcal{D}}\Big)^{\mathcal{S}_\mathcal{D}}.$$
Besides, by property 1 we have
\begin{align*}
	\vert\mathcal{P}_\mathcal{D}\vert&\le\Pi_{i=1}^\mathcal{D-1}2\Big(\frac{2emk_i(1+(i-1)p^{i-1})}{\mathcal{S}_i}\Big)^{\mathcal{S}_i}.
\end{align*}
Then using (\ref{Km}), and since the sample $z_1,\ldots,Z_m$ are arbitrarily chosen, we have
\begin{align*}
	K(m)&\le\Pi_{i=1}^\mathcal{D}2\Big(\frac{2emk_i(1+(i-1)p^{i-1})}{\mathcal{S}_i}\Big)^{\mathcal{S}_i}\\
	&\le2^\mathcal{D}\Big(\frac{2em\sum k_i(1+(i-1)p^{i-1})}{\sum\mathcal{S}_i}\Big)^{\sum\mathcal{S}_i}\\
	&\le\Big(\frac{4em(1+(\mathcal{D}-1)p^{\mathcal{D}-1})\sum k_i}{\sum\mathcal{S}_i}\Big)^{\sum\mathcal{S}_i}\\
	&\le\Big(4em(1+(\mathcal{D}-1)p^{\mathcal{D}-1})\Big)^{\sum\mathcal{S}_i},
\end{align*}
where the second inequality follows from weighted arithmetic and geometric means inequality, the third holds since $\mathcal{D}\le\sum\mathcal{S}_i$ and the last holds since $\sum k_i\le\sum \mathcal{S}_i$. Since $K(m)$ is the growth function of $\tilde{\mathcal{F}}$, we have
\begin{align*}
	2^{\pdim(\mathcal{F})}\le2^{\vdim(\tilde{\mathcal{F}})}\le K(\vdim(\tilde{\mathcal{F}}))\le2^\mathcal{D}\Big(\frac{2eR\cdot\vdim(\tilde{\mathcal{F}})}{\sum\mathcal{S}_i}\Big)^{\sum\mathcal{S}_i}
\end{align*}
where $R:=\sum_{i=1}^\mathcal{D} k_i(1+(i-1)p^{i-1})\le \mathcal{U}+\mathcal{U}(\mathcal{D}-1)p^{\mathcal{D}-1}.$
Since $\mathcal{U}>0$ and $2eR\ge16$, then by Lemma 16 in \citet{bartlett2019nearly} we have
$$\pdim(\mathcal{F})\le\mathcal{D}+(\sum_{i=1}^n\mathcal{S}_i)\log_2(4eR\log_2(2eR)).$$
Note that $\sum_{i=1}^\mathcal{D}\mathcal{S}_i\le\mathcal{D}\mathcal{S}$ and $\log_2(R)\le\log_2(\mathcal{U}\{1+(\mathcal{D}-1)p^{\mathcal{D}-1}\})\le\log_2(\mathcal{U})+p\mathcal{D}$, then we have
$$\pdim(\mathcal{F})\le \mathcal{D}+\mathcal{D}\mathcal{S}(2p\mathcal{D}+2\log_2\mathcal{U}+6)\le3p\mathcal{D}\mathcal{S}(\mathcal{D}+\log_2\mathcal{U})$$
for some universal constant $c>0$.
$\hfill \Box$

\subsection*{Proof of Theorem \ref{represent}}
We begin our proof with consider the simple case, which is to construct a proper RePU network to represent a univariate polynomial with no error. We can leverage Horner’s method or Qin Jiushao's algorithm in China to construct such networks. Suppose $f(x)=a_0+a_1x+\cdots+a_Nx^N$ is a univariate polynomial of degree $N$, then it can be written as
$$f(x)=a_0+x(a_1+x(a_2+x(a_3+\cdots+x(a_{N-1}+xa_N)))).$$
We can iteratively calculate a sequence of  intermediate variables $b_1,\ldots,b_N$ by
\begin{align*}
	b_k=\Big\{\begin{array}{lr}
		a_{N-1}+xa_N, \qquad k=1,\\
		a_{N-k}+xb_{N-1}, \ \ \  k=2,\ldots,N.\\
	\end{array}\Big.
\end{align*}
Then we can obtain $b_N=f(x)$. By (iii) in Lemma \ref{repu_lemma},  we know that a RePU network with one hidden layer and no more than $2p$ nodes can represent any polynomial of the input with order no more than $p$. Obviously, for input $x$, the identity map $x$, linear transformation $ax+b$ and square map $x^2$ are all polynomials of $x$ with order no more than $p$. In addition, it is not hard to see that the multiplication operator $xy=\{(x+y)^2-(x-y)^2\}/4$ can be represented by a RePU network with one hidden layer and $4p$ nodes. Then to calculate $b_1$ needs a RePU network with 1 hidden layer and $2p$ hidden neurons, and to calculate $b_2$ needs a RePU network with 3 hidden layer, $2\times2p+1\times4p+2$ hidden neurons. By induction, to calculate $b_N=f(x)$ for $N\ge1$ needs a RePU network with $2N-1$ hidden layer, $N\times2p+(N-1)\times4p+(N-1)\times2=(6p+2)(N-1)+2p$ hidden neurons, $(N-1)(30p+2)+2p+1$ parameters(weights and bias), and its width equals to $6p$.

Apart from the construction based on the Horner's method, another construction is shown in Theorem 2 of \citet{li2019powernet}, where the constructed RePU network has $\lceil\log_pN\rceil+2$ hidden layers, $O(N)$ neurons and $O(pN)$ parameters (weights and bias).

Now we consider converting RePU networks to multivariate polynomial $f$ with total degree $N$ on $\mathbb{R}^d$. For any $d\in\mathbb{N}^+$ and $N\in\mathbb{N}_0$, let
$$f^d_N(x_1,\ldots,x_d)=\sum_{i_1+\cdots+i_d=0}^N a_{i_1,i_2,\ldots,i_d}x_1^{i_1}x_2^{i_2}\cdots x_d^{i_d},$$
denote the polynomial with total degree $N$ of $d$ variables, where $i_1,i_2,\ldots,i_d$ are non-negative integers, $\{a_{i_1,i_2,\ldots,i_d}: i_1+\cdots+i_d\le N\}$ are coefficients in $\mathbb{R}$. Note that the multivariate polynomial $f^d_N$ can be written as
\begin{align*}
	f^d_N(x_1,\ldots,x_d)=\sum_{i_1=0}^N\Big(\sum_{i_2+\cdots+i_d=0}^{N-i_1} a_{i_1,i_2\ldots,i_d}x_2^{i_2}\cdots x_d^{i_d}\Big)x_1^{i_1},
\end{align*}
and we can view $f^d_N$ as a univariate polynomial of $x_1$ with degree $N$ if $x_2,\ldots,x_d$ are given and for each $i_1\in\{0,\ldots,N\}$ the $(d-1)$-variate polynomial $\sum_{i_2+\cdots+i_d=0}^{N-i_1} a_{i_1,i_2\ldots,i_d}x_2^{i_2}\cdots x_d^{i_d}$ with degree no more than $N$ can be computed by a proper RePU network.
This reminds us the construction of RePU network for $f^d_N$ can be implemented recursively via composition of $f^{1}_N,f^2_N,\ldots,f^{d}_N$ by induction.

By Horner's method we have constructed a RePU network with $2N-1$ hidden layers, $(6p+2)(N-1)+2p$ hidden neurons and $(N-1)(30p+2)+2p+1$ parameters to exactly compute $f^1_N$. Now we start to show $f^2_N$ can be computed by RePU networks. We can write $f^2_N$ as
$$f^2_N(x_1,x_2)=\sum_{i+j=0}^Na_{ij}x_1^ix_2^j=\sum_{i=0}^N\Big(\sum_{j=0}^{N-i}a_{ij}x_2^j\Big)x_1^i.$$
Note that for $i\in\{0,\ldots,N\}$, the the degree of polynomial $\sum_{j=0}^{N-i}a_{ij}x_2^j$ is $N-i$ which is less than $N$. But we can still view it as a polynomial with degree $N$ by padding (adding zero terms) such that $\sum_{j=0}^{N-i}a_{ij}x_2^j=\sum_{j=0}^{N}a^*_{ij}x_2^j$ where $a^*_{ij}=a_{ij}$ if $i+j\le N$ and $a^*_{ij}=0$ if $i+j> N$. In such a way, for each $i\in\{0,\ldots,N\}$ the polynomial $\sum_{j=0}^{N-i}a_{ij}x_2^j$ can be computed by a RePU network with $2N-1$ hidden layers, $(6p+2)(N-1)+2p$ hidden neurons, $(N-1)(30p+2)+2p+1$ parameters and its width equal to $6p$. Besides, for each $i\in\{0,\ldots,N\}$, the monomial $x^i$ can also be computed by a RePU network with $2N-1$ hidden layers, $(6p+2)(N-1)+2p$ hidden neurons, $(N-1)(30p+2)+2p+1$ parameters and its width equal to $6p$, in whose implementation the identity maps are used after the $(2i-1)$-th hidden layer. Now we parallel these two sub networks to get a RePU network which takes $x_1$ and $x_2$ as input and outputs $(\sum_{j=0}^{N-i}a_{ij}x_2^j)x^i$ with width $12p$, hidden layers $2N-1$, number of neurons $2\times[(6p+2)(N-1)+2p]$ and size $2\times[(N-1)(30p+2)+2p+1]$. Since for each $i\in\{0,\ldots,N\}$, such paralleled RePU network can be constructed, then with straightforward paralleling of $N$ such RePU networks, we obtain a RePU network exactly computes $f^2_N$ with width $12pN$, hidden layers $2N-1$, number of neurons $2\times[(6p+2)(N-1)+2p]\times N\le 14pN^2$ and number of parameters $2\times[(N-1)(30p+2)+2p+1]\times N\le 62pN^2$.

Similarly for polynomial $f^3_N$ of $3$ variables, we can write $f^3_N$ as
$$f^3_N(x_1,x_2,x_3)=\sum_{i+j+k=0}^Na_{ijk}x_1^ix_2^jx_3^k=\sum_{i=0}^N\Big(\sum_{j+k=0}^{N-i}a_{ijk}x_2^jx_3^k\Big)x_1^i.$$
By our previous argument, for each $i\in\{0,\ldots,N\}$, there exists a RePU network which takes $(x_1,x_2,x_3)$ as input and outputs $\Big(\sum_{j+k=0}^{N-i}a_{ijk}x_2^jx_3^k\Big)x_1^i$ with width $12pN+6p$, hidden layers $2N-1$, number of neurons $2N\times[(6p+2)(N-1)+2p]+[(6p+2)(N-1)+2p]$ and parameters $2N\times[(N-1)(30p+2)+2p+1]+[(N-1)(30p+2)+2p+1]$. And by paralleling $N$ such subnetworks, we obtain a RePU network that exactly computes $f^3_N$ with width $(12pN+6p)\times N=12pN^2+6pN$, hidden layers $2N-1$, number of neurons $2N^2\times[(6p+2)(N-1)+2p]+N\times[(6p+2)(N-1)+2p]$ and number of parameters $2N^2\times[(N-1)(30p+2)+2p+1]+N\times[(N-1)(30p+2)+2p+1]$.

Continuing this process, we can construct RePU networks exactly compute polynomials of any $d$ variables with total degree $N$. With a little bit abuse of notations, we let $\mathcal{W}_k$, $\mathcal{D}_k$, $\mathcal{U}_k$ and $\mathcal{S}_k$ denote the width, number of hidden layers, number of neurons and number of parameters (weights and bias) respectively of the RePU network computing $f^k_N$ for $k=1,2,3,\ldots$. We have known that
\begin{align*}
	\mathcal{D}_1=2N-1,\quad\mathcal{W}_1=6p,\quad\mathcal{U}_1=(6p+2)(N-1)+2p,\quad\mathcal{S}_1=(N-1)(30p+2)+2p+1.
\end{align*}
Besides, based on the iterate procedure of the network construction, by induction we can see that for $k=2,3,4,\ldots$ the following equations hold,
\begin{align*}
	\mathcal{D}_k=&2N-1,\\
	\mathcal{W}_k=&N\times(\mathcal{W}_{k-1}+\mathcal{W}_1),\\
	\mathcal{U}_k=&N\times(\mathcal{U}_{k-1}+\mathcal{U}_1),\\
	\mathcal{S}_k=&N\times(\mathcal{S}_{k-1}+\mathcal{S}_1).
\end{align*}
Then based on the values of $\mathcal{D}_1,\mathcal{W}_1,\mathcal{U}_1,\mathcal{S}_1$ and the recursion formula, we have for $k=2,3,4,\ldots$
\begin{align*}
	\mathcal{D}_k=2N-1,
\end{align*}
\begin{align*}
	\mathcal{W}_k=12pN^{k-1}+6p\frac{N^{k-1}-N}{N-1},
\end{align*}

\begin{align*}
	\mathcal{U}_k=&N\times(\mathcal{U}_{k-1}+\mathcal{U}_1)=2\mathcal{U}_1N^{k-1}+\mathcal{U}_1\frac{N^{k-1}-N}{N-1}\\
	=&
	(6p+2)(2N^k-N^{k-1}-N)+2p(\frac{2N^k-N^{k-1}-N}{N-1}),
\end{align*}

\begin{align*}
	\mathcal{S}_k=&N\times(\mathcal{S}_{k-1}+\mathcal{S}_1)=2\mathcal{S}_1N^{k-1}+\mathcal{S}_1\frac{N^{k-1}-N}{N-1}\\
	=&(30p+2)(2N^k-N^{k-1}-N)+(2p+1)(\frac{2N^k-N^{k-1}-N}{N-1}).
\end{align*}

This completes our proof. $\hfill \Box$

\subsection*{Proof of Theorem \ref{approx2}}
The proof is straightforward by and leveraging the approximation power of multivariate polynomials based on Theorem \ref{represent}. The theories for polynomial approximation have been extensively studies on various spaces of smooth functions. We refer to \citet{bagby2002multivariate} for the polynomial approximation on smooth functions in our proof.
\begin{lemma}[Theorem 2 in \citet{bagby2002multivariate}]\label{approx_poly}
	Let $f$ be a function of compact support on $\mathbb{R}^d$ of class $C^s$ where $s\in\mathbb{N}^+$ and let $K$ be a compact subset of $\mathbb{R}^d$ which contains the support of $f$. Then for each nonnegative integer $N$ there is a polynomial $p_N$ of degree at most $N$ on $\mathbb{R}^d$ with the following property: for each multi-index $\alpha$ with $\vert\alpha\vert_1\le\min\{s,N\}$ we have
	$$\sup_{K}\vert D^\alpha (f-p_N)\vert\le\frac{C}{N^{s-\vert\alpha\vert_1}}\sum_{\vert\alpha\vert_1\le s}\sup_K\vert D^\alpha f\vert,$$
	where $C$ is a positive constant depending only on $d,s$ and $K$.
\end{lemma}

The proof of Lemma \ref{approx_poly} can be found in \citet{bagby2002multivariate} based on the Whitney extension theorem (Theorem 2.3.6 in \citet{hormander2015analysis}) and by examining the proof of Theorem 1 in \citet{bagby2002multivariate}, the dependence of the constant $C$ in Lemma \ref{approx_poly} on the $d,s$ and $K$ can be detailed.

 To use Lemma \ref{approx_poly}, we need to find a RePU network to compute the $p_N$ for each $N\in\mathbb{N}^+$. By Theorem \ref{represent}, we know that any $p_N$ of $d$ variables can be exactly computed by a RePU network $\phi_N$ with $2N-1$ hidden layers,  $(6p+2)(2N^d-N^{d-1}-N)+2p(2N^d-N^{d-1}-N)/(N-1)$ number of neurons, $(30p+2)(2N^d-N^{d-1}-N)+(2p+1)(2N^d-N^{d-1}-N)/(N-1)$ number of parameters (weights and bias) and network width $12pN^{d-1}+6p(N^{d-1}-N)/(N-1)$.  Then we have
$$\sup_{K}\vert D^\alpha (f-\phi_N)\vert\le C_{s,d,K}N^{-(s-\vert\alpha\vert_1)}\Vert f\Vert_{C^s},$$
where $C_{s,d,K}$ is a positive constant depending only on $,d,s$ and $K$.  Note that the number neurons $\mathcal{U}=\mathcal{O}(18pN^d)$, which implies $(\mathcal{U}/18p)^{1/d}\le N$. Then we also have
$$\sup_{K}\vert D^\alpha (f-\phi_N)\vert\le C_{p,s,d,K}\mathcal{U}^{-(s-\vert\alpha\vert_1)/d}\Vert f\Vert_{C^s},$$
where $C_{p,s,d,K}$ is a positive constant depending only on $,d,s$ and $K$. This completes the proof. $\hfill \Box$

\subsection*{Proof of Theorem \ref{approx_lowdim}} The idea of our proof is based on projecting the data to a low-dimensional space and then use deep RePU neural network to approximate the low-dimensional function.

Given any integer $d_\delta=O(d_\mathcal{M}{\log(d/\delta)}/{\delta^2})$ satisfying $d_\delta\le d$, by Theorem 3.1 in \cite{baraniuk2009random} there exists a linear projector $A\in\mathbb{R}^{d_\delta\times d}$ that maps a low-dimensional manifold in a high-dimensional space to a low-dimensional space nearly preserving the distance. Specifically, there exists a matrix $A\in\mathbb{R}^{d_\delta\times d}$ such that
	$AA^T=(d/d_\delta)I_{d_\delta}$ where $I_{d_\delta}$ is an identity matrix of size $d_\delta\times d_\delta$, and
	$$(1-\delta)\Vert x_1-x_2\Vert_2\leq\Vert Ax_1-Ax_2\Vert_2\leq(1+\delta)\Vert x_1-x_2\Vert_2,$$
	for any $x_1,x_2\in\mathcal{M}.$
	
	Note that for any $z\in A(\mathcal{M})$, there exists a unique $x\in\mathcal{M}$ such that $Ax=z$.
	Then for any $z\in A(\mathcal{M})$, define $x_z=\mathcal{SL}(\{x\in\mathcal{M}: Ax=z\})$ where $\mathcal{SL}(\cdot)$ is a set function which returns a unique element of a set. If $Ax=z$ where $x\in\mathcal{M}$ and $z\in A(\mathcal{M})$, then $x=x_z$ by our argument since $\{x\in\mathcal{M}: Ax=z\}$ is a set with only one element when $z\in A(\mathcal{M})$. We can see that $\mathcal{SL}:A(\mathcal{M})\to\mathcal{M}$ is a differentiable function with the norm of its derivative locates in $[1/(1+\delta),1/(1-\delta)]$, since
	$$\frac{1}{1+\delta}\Vert z_1-z_2\Vert_2\leq\Vert x_{z_1}-x_{z_2}\Vert_2\leq\frac{1}{1-\delta}\Vert z_1-z_2\Vert_2,$$
	for any $z_1,z_2\in A(\mathcal{M})$. 
	For the high-dimensional function $f_0:\mathcal{X}\to\mathbb{R}^1$, we define its low-dimensional representation $\tilde{f}_0:\mathbb{R}^{d_\delta}\to\mathbb{R}^1$ by
	$$\tilde{f}_0(z)=f_0(x_z), \quad {\rm for\ any} \ z\in A(\mathcal{M})\subseteq\mathbb{R}^{d_\delta}.$$
	Recall that $f_0\in C^s(\mathcal{X})$, then $\tilde{f}_0\in C^s(A(\mathcal{M}))$. Note that $\mathcal{M}$ is compact manifold and $A$ is a linear mapping, then by the extended version of Whitney' extension theorem in \cite{fefferman2006whitney}, there exists a function $\tilde{F}_0\in C^s(A(\mathcal{M}_\rho))$ such that $\tilde{F}_0(z)=\tilde{f}_0(z)$ for any $z\in A(\mathcal{M}_\rho)$ and $\Vert \tilde{F}_0\Vert_{C^1}\le (1+\delta)\Vert f_0\Vert_{C^1}$. By Theorem \ref{approx2}, for any $N\in\mathbb{N}^+$, there exists a function $\tilde{f}_n: \mathbb{R}^{d_\delta}\to\mathbb{R}^1$ implemented by a RePU network with its depth $\mathcal{D}$, width $\mathcal{W}$, number of neurons $\mathcal{U}$ and size $\mathcal{S}$ specified as
	\begin{gather*}
		\mathcal{D}=2N-1,\qquad \mathcal{W}=12pN^{d_\delta-1}+6p(N^{d_\delta-1}-N)/(N-1)\\
		\mathcal{U}=(6p+2)(2N^{d_\delta}-N^{d_\delta-1}-N)+2p(2N^{d_\delta}-N^{d_\delta-1}-N)/(N-1),\\
		\mathcal{S}=(30p+2)(2N^{d_\delta}-N^{d_\delta-1}-N)+(2p+1)(2N^{d_\delta}-N^{d_\delta-1}-N)/(N-1),
	\end{gather*}
	such that for each multi-index $\alpha\in\mathbb{N}^d_0$ with $\vert \alpha\vert_1\le1$, we have
	$$
	\vert D^\alpha(\tilde{f}_n(z)-\tilde{F}_0(z))\vert\leq C_{s,d_\delta,A(\mathcal{M}_\rho)} N^{-(s-\vert\alpha\vert_1)}\Vert \tilde{F}_0\Vert_{C^1},$$
	for all $z\in A(\mathcal{M}_\rho)$ where  $C_{s,d_\delta,A(\mathcal{M}_\rho)} >0$ is a constant depending only on $s,d_\delta,A(\mathcal{M}_\rho)$.
	
	By Theorem \ref{represent}, the linear projection $A$ can be computed by a RePU network with 1 hidden layer and its width no more than 18p. If we define $f^*_n=\tilde{f}_n\circ A$ which is $f^*_n(x)=\tilde{f}_n(Ax)$ for any $x\in\mathcal{X}$, then $f^*_n\in\mathcal{F}_{\mathcal{D},\mathcal{W},\mathcal{U},\mathcal{S},\mathcal{B}}$ is also a RePU network with one more layer than $\tilde{f}_n$. For any $x\in\mathcal{M}_\rho$, there exists a $\tilde{x}\in\mathcal{M}$ such that $\Vert x-\tilde{x}\Vert_2\leq \rho$. Then, for each multi-index $\alpha\in\mathbb{N}^d_0$ with $\vert\alpha\vert_1\le 1$, we have
	\begin{align*}
		&\vert D^\alpha (f^*_n(x)-f_0(x))\vert=\vert D^\alpha( \tilde{f}_n(Ax)-\tilde{F}_0(Ax)+\tilde{F}_0(Ax)-\tilde{F}_0(A\tilde{x})+\tilde{F}_0(A\tilde{x})-f_0(x))\vert\\
		&\leq\vert D^\alpha(\tilde{f}_n(Ax)-\tilde{F}_0(Ax))\vert+\vert D^\alpha(\tilde{F}_0(Ax)-\tilde{F}_0(A\tilde{x}))\vert+\vert D^\alpha(\tilde{F}_0(A\tilde{x})-f_0(x))\vert\\
		&\le C_{s,d_\delta,A(\mathcal{M}_\rho)} N^{-(s-\vert\alpha\vert_1)}\Vert \tilde{F}_0\Vert_{C^{\vert\alpha\vert_1}}+(1+\delta)\rho \Vert \tilde{F}_0\Vert_{C^{\vert\alpha\vert_1}} + \vert D^\alpha( f_0(\tilde{x})-f_0(x))\vert\\
		&\le \big[C_{s,d_\delta,A(\mathcal{M}_\rho)} N^{-(s-\vert\alpha\vert_1)} +(1+\delta)\rho\big]\Vert \tilde{F}_0\Vert_{C^{\vert\alpha\vert_1}}+ \rho \Vert f_0\Vert_{C^{\vert\alpha\vert_1}}\\
		&\le  C_{s,d_\delta,A(\mathcal{M}_\rho)} (1+\delta)\Vert f_0\Vert_{C^{\vert\alpha\vert_1}} N^{-(s-\vert\alpha\vert_1)}+ 2(1+\delta)^2\rho\Vert f_0\Vert_{C^{\vert\alpha\vert_1}}\\
		&\le  \tilde{C}_{s,d_\delta,A(\mathcal{M}_\rho)} (1+\delta)\Vert f_0\Vert_{C^{\vert\alpha\vert_1}} N^{-(s-\vert\alpha\vert_1)},\\
		&\le  {C}_{p,s,d_\delta,A(\mathcal{M}_\rho)} (1+\delta)\Vert f_0\Vert_{C^{\vert\alpha\vert_1}} \mathcal{U}^{-(s-\vert\alpha\vert_1)/d_\delta},
	\end{align*}
	where ${C}_{p,s,d_\delta,A(\mathcal{M}_\rho)}$ is a constant depending only on $p,s,d_\delta,A(\mathcal{M}_\rho)$. The second last inequality follows from $\rho\leq C_1 N^{-(s-1)}(1+\delta)^{-1}$. Since the number neurons $\mathcal{U}=\mathcal{O}(18pN^d)$ and $(\mathcal{U}/18p)^{1/d}\le N$, the last inequality follows. This completes the proof.
	
	$\hfill \Box$

\subsection*{Proof of Lemma \ref{non-asymp-score}}
For the empirical risk minimizer $\hat{s}_n$ based on the sample $S=\{X_i\}_{i=1}^n$, we consider its excess risk $\mathbb{E}\{J(\hat{s}_n)-J(s_0)\}$. 

 For any $s\in\mathcal{F}_n$, we have
 \begin{align*}
 	J(\hat{s}_n)-J(s_0)&=J(\hat{s}_n)-J_n(\hat{s}_n)+J_n(\hat{s}_n)-J_n(s)+J_n(s)-J(s)+J(s) -J(s_0)\\
 	&\le J(\hat{s}_n)-J_n(\hat{s}_n) +J_n(s)-J(s)+J(s) -J(s_0)\\
 	&\le 2\sup_{s\in\mathcal{F}_n}\vert J(s)-J_n(s)\vert + J(s) -J(s_0),
 \end{align*}
 where the first inequality follows from the definition of empirical risk minimizer $\hat{s}_n$, and the second inequality holds since $\hat{s}_n,s\in\mathcal{F}_n$. Note that above inequality holds for any $s\in\mathcal{F}_n$, then
 \begin{align*}
 	J(\hat{s}_n)-J(s_0)&=\le 2\sup_{s\in\mathcal{F}_n}\vert J(s)-J_n(s)\vert + \inf_{s\in\mathcal{F}_n}[J(s) -J(s_0)],
 \end{align*}
 where we call $2\sup_{s\in\mathcal{F}_n}\vert J(s)-J_n(s)\vert$ the stochastic error and $\inf_{s\in\mathcal{F}_n}[J(s) -J(s_0)]$ the approximation error.

\subsubsection*{Bounding the stochastic error}

Recall that $s,s_0$ are vector-valued functions, we write $s=(s_1,\ldots,s_d)^\top, s_0=(s_{01},\ldots,s_{0d})^\top$ and let $\pj s_j$ denote the $j$-th diagonal entry in $\nabla_x s$ and $\pj s_{0j}$ the $j$-th diagonal entry in $\nabla_x s_0$.
\begin{align*}
	J(s)&=\mathbb{E} \Big[ tr(\nabla_x s(X)) + \frac{1}{2}\Vert s(X)\Vert^2_2 \Big]=\mathbb{E}\Big[ \sum_{j=1}^d \pj s_j(X) + \frac{1}{2} \sum_{j=1}^{d} \vert s_j(X)\vert^2\Big].
\end{align*}
Define
\begin{align*}
	J^{1,j}(s)&=\mathbb{E}\Big[ \pj s_j(X)\Big]\qquad J^{1,j}_n(s)=\sum_{i=1}^{n}\Big[ \pj s_j(X_i)\Big]\\
	J^{2,j}(s)&=\mathbb{E}\Big[ \frac{1}{2} \vert s_j(X)\vert^2\Big]\qquad J^{2,j}_n(s)=\sum_{i=1}^{n}\Big[ \frac{1}{2}\vert s_j(X_i)\vert^2\Big]
\end{align*}
for $j=1,\ldots,d$ and $s\in\mathcal{F}_n$. Then,
\begin{align}\notag
	\sup_{s\in\mathcal{F}_n}\vert J(s)-J_n(s)\vert&\le \sup_{s\in\mathcal{F}_n} \sum_{j=1}^{d} \Big[\vert J^{1,j}(s)-J^{1,j}_n(s)\vert + \vert J^{2,j}(s)-J^{2,j}_n(s)\vert \Big]\\ \label{sto_sup}
	&\le \sum_{j=1}^{d} \sup_{s\in\mathcal{F}_n} \vert J^{1,j}(s)-J^{1,j}_n(s)\vert + \sum_{j=1}^{d} \sup_{s\in\mathcal{F}_n} \vert J^{2,j}(s)-J^{2,j}_n(s)\vert.
\end{align}

Recall that for any $s\in\mathcal{F}_n$, the output $\Vert s_j\Vert_\infty\le \mathcal{B}$ and the partial derivative $\Vert \pj s_j\Vert_\infty\le \mathcal{B}^\prime$ for $j=1,\ldots,d$. Then by Theorem 11.8 in \cite{mohri2018foundations}, for any $\delta>0$, with probability at least $1-\delta$ over the choice of $n$ i.i.d sample $S$,
\begin{gather}\label{sto_j1}
	\sup_{s\in\mathcal{F}_n} \vert J^{1,j}(s)-J^{1,j}_n(s)\vert\le 2\mathcal{B}^\prime\sqrt{\frac{2\pdim(\mathcal{F}^\prime_{jn})\log(en)}{n}}+2\mathcal{B}^\prime\sqrt{\frac{\log(1/\delta)}{2n}}
\end{gather}
for $j=1,\ldots,d$ where $\pdim(\mathcal{F}^\prime_{jn})$ is the Pseudo dimension of $\mathcal{F}^\prime_{jn}$ and $\mathcal{F}^\prime_{jn}=\{\pj s_j: s\in\mathcal{F}_n\}$.
Similarly, 
with probability at least $1-\delta$ over the choice of $n$ i.i.d sample $S$,
\begin{gather}\label{sto_j2}
	\sup_{s\in\mathcal{F}_n} \vert J^{2,j}(s)-J^{2,j}_n(s)\vert\le \mathcal{B}^2\sqrt{\frac{2\pdim(\mathcal{F}_{jn})\log(en)}{n}}+\mathcal{B}^2\sqrt{\frac{\log(1/\delta)}{2n}}
\end{gather}
for $j=1,\ldots,d$ where  $\mathcal{F}_{jn}=\{s_j: s\in\mathcal{F}_n\}$.
Combining (\ref{sto_sup}), (\ref{sto_j1}) and (\ref{sto_j2}), we have proved that for any $\delta>0$, with probability at least $1-2d\delta$,
\begin{align*}
	\sup_{s\in\mathcal{F}_n}\vert J(s)-J_n(s)\vert&\le \sqrt{\frac{2\log(en)}{n}}\sum_{j=1}^{d}\Big[\mathcal{B}^2\sqrt{\pdim(\mathcal{F}_{jn})}+\mathcal{B}^\prime\sqrt{\pdim(\mathcal{F}^\prime_{jn})}\Big]\\
	&\qquad\qquad+d\Big[\mathcal{B}^2+2\mathcal{B}^\prime\Big]\sqrt{\frac{\log(1/\delta)}{2n}}.
\end{align*}

Note that $s=(s_1,\ldots,s_d)^\top$ for $s\in\mathcal{F}_n$ where $\mathcal{F}_n=\mathcal{F}_{\mathcal{D},\mathcal{W},\mathcal{U},\mathcal{S},\mathcal{B},\mathcal{B}^\prime}$ is a class of RePU neural networks with depth $\mathcal{D}$, width $\mathcal{W}$, size $\mathcal{S}$ and number of neurons $\mathcal{U}$. Then for each $j=1,\ldots,d$, the function class $\mathcal{F}_{jn}=\{s_j: s\in\mathcal{F}_n\}$ consists of RePU neural networks with depth $\mathcal{D}$, width $\mathcal{W}$, number of neurons $\mathcal{U}-(d-1)$ and size no more than $\mathcal{S}$. By Lemma \ref{lemmapdim}, we have $\pdim(\mathcal{F}_{1n})=\pdim(\mathcal{F}_{2n})=\ldots=\pdim(\mathcal{F}_{dn})\le 3p\mathcal{D}\mathcal{S}(\mathcal{D}+\log_2\mathcal{U})$. Similarly, by Theorem \ref{thm2} and Lemma \ref{lemmapdim}, we have $\pdim(\mathcal{F}^\prime_{1n})=\pdim(\mathcal{F}^\prime_{2n})=\ldots=\pdim(\mathcal{F}^\prime_{dn})\le 2484p\mathcal{D}\mathcal{S}(\mathcal{D}+\log_2\mathcal{U})$. Then, for any $\delta>0$, with probability at least $1-\delta$
\begin{align*}\label{sto_sup2}
	\sup_{s\in\mathcal{F}_n}\vert J(s)-J_n(s)\vert&\le 50\times d(\mathcal{B}^2+2\mathcal{B}^\prime)\Bigg(50\sqrt{\frac{2p\log(en)\mathcal{D}\mathcal{S}(\mathcal{D}+\log_2\mathcal{U})}{n}}+\sqrt{\frac{\log(2d/\delta)}{2n}}\Bigg).
\end{align*}
If we let $t=50 d(\mathcal{B}^2+2\mathcal{B}^\prime)(50\sqrt{{2p\log(en)\mathcal{D}\mathcal{S}(\mathcal{D}+\log_2\mathcal{U})}/{n}})$, then above inequality implies

\begin{align*}
	\mathbb{P}\Bigg(\sup_{s\in\mathcal{F}_n}\vert J(s)-J_n(s)\vert&\ge  \epsilon\Bigg)\le 2d\exp\left(\frac{-2n(\epsilon-t)^2}{[50d(\mathcal{B}^2+2\mathcal{B}^\prime)]^2}\right),
\end{align*}
 for $\epsilon\ge t$. And

\begin{align*}
&	\mathbb{E}\left[\sup_{s\in\mathcal{F}_n}\vert J(s)-J_n(s)\vert\right]\\
&=\int_0^\infty \mathbb{P}\Bigg(\sup_{s\in\mathcal{F}_n}\vert J(s)-J_n(s)\vert\ge  u\Bigg) du\\
	&=\int_0^t \mathbb{P}\Bigg(\sup_{s\in\mathcal{F}_n}\vert J(s)-J_n(s)\vert\ge  u\Bigg) du+\int_t^\infty \mathbb{P}\Bigg(\sup_{s\in\mathcal{F}_n}\vert J(s)-J_n(s)\vert\ge  u\Bigg) du\\
	&\le \int_0^t 1 du+\int_t^\infty 2d\exp\left(\frac{-2n(u-t)^2}{[50d(\mathcal{B}^2+2\mathcal{B}^\prime)]^2}\right) du\\
	&=t+25\sqrt{2\pi}d^2(\mathcal{B}^2+2\mathcal{B}^\prime)\frac{1}{\sqrt{n}}\\
	&\le 2575d^2(\mathcal{B}^2+2\mathcal{B}^\prime)\sqrt{{2p\log(en)\mathcal{D}\mathcal{S}(\mathcal{D}+\log_2\mathcal{U})}/{n}}.
\end{align*}

\subsubsection*{Bounding the approximation error}

Recall that for $s\in\mathcal{F}_n$,
\begin{align*}
	J(s)&=\mathbb{E} \Big[ tr(\nabla_x s(X)) + \frac{1}{2}\Vert s(X)\Vert^2_2 \Big]=\mathbb{E}\Big[ \sum_{j=1}^d \pj s_j(X) + \frac{1}{2} \sum_{j=1}^{d} \vert s_j(X)\vert^2\Big],
\end{align*}
and the excess risk
\begin{align*}
&	J(s)-J(s_0)\\
&=\mathbb{E}\Big[ \sum_{j=1}^d \pj s_j(X) + \frac{1}{2} \sum_{j=1}^{d} \vert s_j(X)\vert^2\Big]-\mathbb{E}\Big[ \sum_{j=1}^d \pj s_{0j}(X) + \frac{1}{2} \sum_{j=1}^{d} \vert s_{0j}(X)\vert^2\Big]\\
	&=\sum_{j=1}^d \mathbb{E}\Big[ \pj s_j(X)-\pj s_{0j}(X) + \frac{1}{2} \vert s_{j}(X)\vert^2 -\frac{1}{2}\vert s_{0j}(X)\vert^2  \Big].
\end{align*}

Recall that $s=(s_1,\ldots,s_d)^\top$ and $s_0=(s_{01},\ldots,s_{0d})^\top$ are vector-valued functions.  
For each $j=1,\ldots,d$, we let $\mathcal{F}_{jn}$ be a class of RePU neural networks with depth $\mathcal{D}$, width $\mathcal{W}$, size $\mathcal{S}$ and number of neurons $\mathcal{U}$. Define $\tilde{\mathcal{F}}_n=\{s=(s_1,\ldots,s_d)^\top: s_j\in\mathcal{F}_{jn}, j=1,\ldots,d\}$. The neural networks in $\tilde{\mathcal{F}}_n$ has depth $\mathcal{D}$, width $d\mathcal{W}$, size $d\mathcal{S}$ and number of neurons $d\mathcal{U}$, which can be seen as built by paralleling $d$ subnetworks in  $\mathcal{F}_{jn}$ for $j=1,\ldots,d$. Let $\mathcal{F}_n$ be the class of all RePU neural networks with depth $\mathcal{D}$, width $d\mathcal{W}$, size $d\mathcal{S}$ and number of neurons $d\mathcal{U}$. Then $\tilde{\mathcal{F}}_n\subset \mathcal{F}_n$ and

\begin{align*}
&	\inf_{s\in\mathcal{F}_n} \Big[J(s)-J(s_0)\Big] \\
&\le \inf_{s\in\tilde{\mathcal{F}}_n} \Big[J(s)-J(s_0)\Big]\\
&	= \inf_{s=(s_1,\ldots,s_d)^\top, s_j\in\mathcal{F}_{jn}, j=1,\ldots,d} \Big[J(s)-J(s_0)\Big]\\
&	= \sum_{j=1}^{d}  \inf_{s_j\in\mathcal{F}_{jn}} \mathbb{E}\Big[ \pj s_j(X)-\pj s_{0j}(X) + \frac{1}{2} \vert s_{j}(X)\vert^2 -\frac{1}{2}\vert s_{0j}(X)\vert^2 \Big].
\end{align*}
Now we focus on derive upper bound for
$$ \inf_{s_j\in\mathcal{F}_{jn}} \mathbb{E}\Big[ \pj s_j(X)-\pj s_{0j}(X) + \frac{1}{2} \vert s_{j}(X)\vert^2 -\frac{1}{2}\vert s_{0j}(X)\vert^2 \Big]$$ for each $j=1,\ldots,d$. By assumption, $\Vert \pj s_{0j}\Vert_{\infty},\Vert \pj s_{j}\Vert_{\infty}\le \mathcal{B}^\prime$ and $\Vert s_{0j}\Vert_{\infty},\Vert s_{j}\Vert_{\infty}\le \mathcal{B}$ for any $s\in\mathcal{F}_{jn}$. For each $j=1,\ldots,n$, the target $s_{0j}$ defined on domain $\mathcal{X}$ is a real-valued function belonging to class $C^m$ for $1\le m<\infty$. By Theorem \ref{approx2}, for any $N\in\mathbb{N}^+$, there exists a RePU activated neural network $s_{Nj}$ with  $2N-1$ hidden layers,  $(6p+2)(2N^d-N^{d-1}-N)+2p(2N^d-N^{d-1}-N)/(N-1)$ number of neurons, $(30p+2)(2N^d-N^{d-1}-N)+(2p+1)(2N^d-N^{d-1}-N)/(N-1)$ number of parameters and network width $12pN^{d-1}+6p(N^{d-1}-N)/(N-1)$ such that for each multi-index $\alpha\in\mathbb{N}^d_0$, we have $\vert\alpha\vert_1\le1$,
$$\sup_{\mathcal{X}}\vert D^\alpha (s_{0j}-s_{Nj})\vert\le C_{m,d,\mathcal{X}} N^{-(m-\vert\alpha\vert_1)}\Vert s_{0j}\Vert_{C^{\vert \alpha\vert_1}},$$
where $C_{m,d,\mathcal{X}}$  is a positive constant depending only on $d,m$ and the diameter of $\mathcal{X}$.
Then
\begin{align*}
	&\inf_{s_j\in\mathcal{F}_{jn}} \mathbb{E}\Big[ \pj s_j(X)-\pj s_{0j}(X) + \frac{1}{2} \vert s_{j}(X)\vert^2 -\frac{1}{2}\vert s_{0j}(X)\vert^2 \Big]\\
	&\le \mathbb{E}\Big[ \pj s_{Nj}(X)-\pj s_{0j}(X) + \frac{1}{2} \vert s_{Nj}(X)\vert^2 -\frac{1}{2}\vert s_{0j}(X)\vert^2 \Big]\\
	&\le  C_{m,d,\mathcal{X}} N^{-(m-1)}\Vert s_{0j}\Vert_{C^{1}}+ C_{m,d,\mathcal{X}} \mathcal{B} N^{-m}\Vert s_{0j}\Vert_{C^0}\\
	&\le {C}_{m,d,\mathcal{X}} (1+\mathcal{B}) N^{-(m-1)} \Vert s_{0j}\Vert_{C^{1}}
\end{align*}
holds for each $j=1,\ldots,d$. Sum up above inequalities, we have proved that
$$\inf_{s\in\mathcal{F}_n} \Big[J(s)-J(s_0)\Big]\le {C}_{m,d,\mathcal{X}} (1+\mathcal{B}) N^{-(m-1)} \Vert s_{0j}\Vert_{C^{1}}.$$

\subsubsection*{Non-asymptotic error bound}
Based on the obtained stochastic error bound and approximation error bound, we can conclude that with probability at least $1-\delta$, the empirical risk minimizer $\hat{s}_n$ defined in (\ref{erm_score}) satisfies
\begin{align*}
	J(\hat{s}_n)-J(s_0)&\le 100\times d(\mathcal{B}^2+2\mathcal{B}^\prime)\Bigg(50\sqrt{\frac{2p\log(en)\mathcal{D}\mathcal{S}(\mathcal{D}+\log_2\mathcal{U})}{n}}+\sqrt{\frac{\log(2d/\delta)}{2n}}\Bigg)\\
	&\qquad\qquad\qquad\qquad\qquad\qquad\qquad +{C}_{m,d,\mathcal{X}} (1+\mathcal{B}) N^{-(m-1)} \Vert s_{0}\Vert_{C^{1}},
\end{align*}
and
\begin{align*}
	\mathbb{E}\{J(\hat{s}_n)-J(s_0)\}&\le 2575d^2(\mathcal{B}^2+2\mathcal{B}^\prime)\sqrt{{2p\log(en)\mathcal{D}\mathcal{S}(\mathcal{D}+\log_2\mathcal{U})}/{n}}\\
	&\qquad\qquad\qquad\qquad\qquad\qquad\qquad +{C}_{m,d,\mathcal{X}} (1+\mathcal{B}) N^{-(m-1)} \Vert s_{0}\Vert_{C^{1}},
\end{align*}
where $C_{m,d,\mathcal{X}}$  is a positive constant depending only on $d,m$ and the diameter of $\mathcal{X}$.

Note that the network depth $\mathcal{D}=2N-1$ is a positive odd number. Then we let $\mathcal{D}$  be a positive odd number, and let the class of neuron network specified by depth $\mathcal{D}$, width $\mathcal{W}=18pd[(\mathcal{D}+1)/2]^{d-1}$, neurons $\mathcal{U}=18pd[(\mathcal{D}+1)/2]^{d}$ and size $\mathcal{S}=67pd[(\mathcal{D}+1)/2]^{d}$. Then we can further express the stochastic error in term of $\mathcal{U}$:

\begin{align*}
	\mathbb{E}\{J(\hat{s}_n)-J(s_0)\}&\le C p^2d^3(\mathcal{B}^2+2\mathcal{B}^\prime)\mathcal{U}^{(d+2)/{2d}}(\log n)^{1/2}{n}^{-1/2}\\
	&\qquad\qquad\qquad\qquad\qquad\qquad\qquad +{C}_{m,d,\mathcal{X}} (1+\mathcal{B}) \Vert s_{0}\Vert_{C^{1}} \mathcal{U}^{-(m-1)/d} ,
\end{align*}
where $C$ is a universal positive constant and $C_{m,d,\mathcal{X}}$  is a positive constant depending only on $d,m$ and the diameter of $\mathcal{X}$.   $\hfill \Box$

\subsection*{Proof of Lemma \ref{non-asymp-score-low}}
By applying Theorem \ref{approx_lowdim}, we can prove Lemma \ref{non-asymp-score-low} similarly following the proof of Lemma \ref{non-asymp-score}.

\subsection*{Proof of Lemma \ref{decom}}

Recall that $\hat{f}^\lambda_n$ is the empirical risk minimizer. Then, for any $f\in\mathcal{F}_n$ we have
\begin{align*}
	\mathcal{R}^\lambda_n(\hat{f}^\lambda_n)\le\mathcal{R}^\lambda_n(f).
\end{align*}
For any $f\in\mathcal{F}_n$, let
$$\rho^\lambda(f):=\frac{1}{d}\sum_{j=1}^d\lambda_j\mathbb{E}\{\rho(\pj f(X))\}$$
and $$\rho^\lambda_n(f):=\frac{1}{n\times d}\sum_{i=1}^n\sum_{j=1}^d\lambda_j\mathbb{E}\{\rho(\pj f(X_i))\}.$$
Then for any $f\in\mathcal{F}_n$,  we have $\rho^\lambda(f)\ge0$ and  $\rho^\lambda_n(f)\ge0$ since $\rho^\lambda$ and $\rho^\lambda_n$ are nonnegative functions and $\lambda_j$'s are nonnegative numbers. Note that $\rho^\lambda(f_0)=\rho^\lambda_n(f_0)=0$ by the assumption that $f_0$ is coordinate-wisely nondecreasing. Then,
\begin{align*}
	\mathcal{R}(\hat{f}^\lambda_n)-\mathcal{R}(f_0)\le&\mathcal{R}(\hat{f}^\lambda_n)-\mathcal{R}(f_0)+\rho^\lambda(\hat{f}^\lambda_n)-\rho^\lambda(f_0)=\mathcal{R}^\lambda(\hat{f}^\lambda_n)-\mathcal{R}^\lambda(f_0).
\end{align*}
We can then give upper bounds for the excess risk $\mathcal{R}(\hat{f}^\lambda_n)-\mathcal{R}(f_0)$. For any $f\in\mathcal{F}_n$,
\begin{align*}
	&\mathbb{E}\{\mathcal{R}(\hat{f}^\lambda_n)-\mathcal{R}(f_0)\}\\
	&\le\mathbb{E}\{\mathcal{R}^\lambda(\hat{f}^\lambda_n)-\mathcal{R}^\lambda(f_0)\}\\
	&\le\mathbb{E}\{\mathcal{R}^\lambda(\hat{f}^\lambda_n)-\mathcal{R}^\lambda(f_0)\}+2\mathbb{E}\{\mathcal{R}_n^\lambda(f)-\mathcal{R}_n^\lambda(\hat{f}^\lambda_n)\}\\
	&=\mathbb{E}\{\mathcal{R}^\lambda(\hat{f}^\lambda_n)-\mathcal{R}^\lambda(f_0)\}+2\mathbb{E}[\{\mathcal{R}_n^\lambda(f)-\mathcal{R}_n^\lambda(f_0)\}-\{\mathcal{R}_n^\lambda(\hat{f}^\lambda_n)-\mathcal{R}_n^\lambda(f_0)\}]\\
	&=\mathbb{E}\{\mathcal{R}^\lambda(\hat{f}^\lambda_n)-2\mathcal{R}^\lambda_n(\hat{f}^\lambda_n)+\mathcal{R}^\lambda(f_0)\}+2\mathbb{E}\{\mathcal{R}_n^\lambda(f)-\mathcal{R}_n^\lambda(f_0)\}
\end{align*}
where the second inequality holds by the the fact that $\hat{f}^\lambda_n$ satisfies $\mathcal{R}_n^\lambda(f)\ge\mathcal{R}_n^\lambda(\hat{f}^\lambda_n)$ for any $f\in\mathcal{F}_n$. Since the inequality holds for any $f\in\mathcal{F}_n$, we have
\begin{align*}
	\mathbb{E}\{\mathcal{R}(\hat{f}^\lambda_n)-\mathcal{R}(f_0)\}&\le\mathbb{E}\{\mathcal{R}^\lambda(\hat{f}^\lambda_n)-2\mathcal{R}^\lambda_n(\hat{f}^\lambda_n)+\mathcal{R}^\lambda(f_0)\}+2\inf_{f\in\mathcal{F}_n}\{\mathcal{R}^\lambda(f)-\mathcal{R}^\lambda(f_0)\}.
\end{align*}
This completes the proof. $\hfill \Box$

\subsection*{Proof of Lemma \ref{non-asymp}}
Lemma \ref{non-asymp} can be proved by combing Lemma \ref{decom}, Lemma \ref{stoerr} and Lemma \ref{lemma5}.

\subsection*{Proof of Lemma \ref{non-asymp-low}}
Lemma \ref{non-asymp-low} can be proved by combing Lemma \ref{decom}, Lemma \ref{stoerr}, Lemma \ref{lemma5} and Theorem \ref{approx_lowdim}.

{\color{black}

\subsection*{Proof of Lemma \ref{non-asymp-mis}}

Under the misspecified model, the target function $f_0$ may not be monotonic, and the quantity $\sum_{j=1}^{d}\lambda_j[\rho(\pj f^\lambda_n(X_i))-\rho(\pj{f}_0(X_i))]$ is not guaranteed to be positive, which prevents us to use the decomposition technique in proof of Lemma \ref{stoerr} to get a fast rate. Instead, we use the canonical decomposition of the excess risk. Let $S=\{Z_i=(X_i,Y_i)\}_{i=1}^n$ be the sample, and let $S_{X}=\{X_i\}_{i=1}^n$ and $S_{Y}=\{Y_i\}_{i=1}^n$. We notice that
\begin{align*}
	\mathbb{E}[\mathcal{R}(\hat{f}^\lambda_n)-	\mathcal{R}(f_0)]&\le\mathbb{E}\Big[ \mathcal{R}(\hat{f}^\lambda_n)-\mathcal{R}(f_0)+\sum_{j=1}^d\lambda_j\mathbb{E}[\rho(\pj \hat{f}^\lambda_n(X))]\Big]\\
	&=\mathbb{E}\Big[\mathcal{R}^\lambda(\hat{f}^\lambda_n)-\mathcal{R}^\lambda(f_0)\Big]+\sum_{j=1}^d\lambda_j\mathbb{E}[\rho(\pj f_0(X))],
\end{align*}
and
\begin{align*}
	&\mathbb{E}\Big[\mathcal{R}^\lambda(\hat{f}^\lambda_n)-\mathcal{R}^\lambda(f_0)\Big]\\
	&=\mathbb{E}\Big[\mathcal{R}^\lambda(\hat{f}^\lambda_n)-\mathcal{R}^\lambda_n(\hat{f}^\lambda_n)+\mathcal{R}^\lambda_n(\hat{f}^\lambda_n)-\mathcal{R}^\lambda(f^*_n)+\mathcal{R}^\lambda(f^*_n)-\mathcal{R}^\lambda(f_0)\Big]\\
	&\le\mathbb{E}\Big[\mathcal{R}^\lambda(\hat{f}^\lambda_n)-\mathcal{R}^\lambda_n(\hat{f}^\lambda_n)+\mathcal{R}^\lambda_n(f^*_n)-\mathcal{R}^\lambda(f^*_n)+\mathcal{R}^\lambda(f^*_n)-\mathcal{R}^\lambda(f_0)\Big]\\
	&=\mathbb{E}\Big[[\mathcal{R}^\lambda(\hat{f}^\lambda_n)-\mathcal{R}^\lambda(f_0)]-[\mathcal{R}^\lambda_n(\hat{f}^\lambda_n)-\mathcal{R}^\lambda_n(f_0)]\\
	&+[\mathcal{R}^\lambda_n(f^*_n)-\mathcal{R}^\lambda_n(f_0)]-[\mathcal{R}^\lambda(f^*_n)-\mathcal{R}^\lambda(f_0)]+\mathcal{R}^\lambda(f^*_n)-\mathcal{R}^\lambda(f_0)\Big]\\
	&\le \mathbb{E}\Big[2\sup_{f\in\mathcal{F}_n}\Big\vert [\mathcal{R}^\lambda(f)-\mathcal{R}^\lambda(f_0)]-\mathbb{E}[\mathcal{R}^\lambda_n(f)-\mathcal{R}^\lambda_n(f_0)\mid {S_X}]\Big\vert\Big] +\inf_{f\in\mathcal{F}_n} [\mathcal{R}^\lambda(f)-\mathcal{R}^\lambda(f_0)],
\end{align*}
where $\mathcal{R}^\lambda(f_n^*)=\inf_{f\in\mathcal{F}_n}\mathcal{R}^\lambda(f)$, $\mathbb{E}$ denotes the expectation taken with respect to $S$, and $\mathbb{E}[\cdot\mid S_X]$ denotes the conditional expectation given $S_X$. Then we have
\begin{align*}
	\mathbb{E}\Big[\mathcal{R}^\lambda(\hat{f}^\lambda_n)-\mathcal{R}^\lambda(f_0)\Big]&\le \mathbb{E}\Big[2\sup_{f\in\mathcal{F}_n}\Big\vert [\mathcal{R}^\lambda(f)-\mathcal{R}^\lambda(f_0)]-\mathbb{E}[\mathcal{R}^\lambda_n(f)-\mathcal{R}^\lambda_n(f_0)\mid {S_X}]\Big\vert\Big] \\
	&+\inf_{f\in\mathcal{F}_n} [\mathcal{R}^\lambda(f)-\mathcal{R}^\lambda(f_0)]+\sum_{j=1}^d\lambda_j\mathbb{E}[\rho(\pj f_0(X))],
\end{align*}
where the first term is the stochastic error, the second term is the approximation error, and the third term the misspecification error with respect to the penalty. Compared with the decomposition in Lemma \ref{decom}, the approximation error is the same and can be bounded using Lemma \ref{lemma5}. However, the stochastic error is different, and there is an additional misspecification error. We will leave the misspecification error untouched and include it in the final upper bound. Next, we focus on deriving the upper bound for the stochastic error.

 For $f\in\mathcal{F}_n$ and each $Z_i=(X_i,Y_i)$ and $j=1\ldots,d$, let
 \begin{align*}
 	g_1(f,X_i) =\mathbb{E}\Big[\vert Y_i-f(X_i)&\vert^2-\vert Y_i-f_0(X_i)\vert^2 \mid X_i\Big]=\vert f(X_i)-f_0(X_i)\vert^2\\
 	g^j_2(f,X_i) &=\rho(\pj f(X_i))- \rho(\pj f_0(X_i)).
 \end{align*}
Then we have
\begin{align*}
&\sup_{f\in\mathcal{F}_n}\left\vert [\mathcal{R}^\lambda(f)-\mathcal{R}^\lambda(f_0)]-\mathbb{E}[\mathcal{R}^\lambda_n(f)-\mathcal{R}^\lambda_n(f_0)\mid {S_X}]\right\vert\\
&\le\sup_{f\in\mathcal{F}_n}\Big\vert\mathbb{E}[g_1(f,X)]-\frac{1}{n}\sum_{i=1}^{n}g_1(f,X_i)+\frac{1}{d}\sum_{j=1}^{d}\lambda_j\Big[\mathbb{E}[g^j_2(f,X)]-\frac{1}{n}\sum_{i=1}^{n}g^j_2(f,X_i)\Big]\Big\vert\\
	&\le\sup_{f\in\mathcal{F}_n}\Big\vert \mathbb{E}[g_1(f,X)]-\frac{1}{n}\sum_{i=1}^{n}g_1(f,X_i)\Big\vert+ \frac{1}{d}\sum_{j=1}^{d}\lambda_j\sup_{f\in\mathcal{F}_n} \Big\vert  \mathbb{E}[g^j_2(f,X)]-\frac{1}{n}\sum_{i=1}^{n}g^j_2(f,X_i)\Big\vert.
\end{align*}
Recall that for any $f\in\mathcal{F}_n$, the $\Vert f\Vert_\infty\le \mathcal{B}, \Vert \pj f \Vert\le \mathcal{B}^\prime$ and by assumption $\Vert f_0\Vert_\infty\le\mathcal{B},\Vert \pj f_0\Vert_\infty\le\mathcal{B}^\prime$ for $j=1,\ldots,d$. By applying Theorem 11.8 in \cite{mohri2018foundations}, for any $\delta>0$, with probability at least $1-\delta$ over the choice of $n$ i.i.d sample $S$,
\begin{gather*}
\sup_{f\in\mathcal{F}_n}\Big\vert \mathbb{E}[g_1(f,X)]-\frac{1}{n}\sum_{i=1}^{n}g_1(f,X_i)\Big\vert\le 4\mathcal{B}^2\sqrt{\frac{2\pdim(\mathcal{F}_{n})\log(en)}{n}}+4\mathcal{B}^2\sqrt{\frac{\log(1/\delta)}{2n}},
\end{gather*}
and
\begin{gather*}
	\sup_{f\in\mathcal{F}_n} \vert \mathbb{E}[g^j_2(f,X)]-\frac{1}{n}\sum_{i=1}^{n}g^j_2(f,X_i)\vert\le 2\kappa\mathcal{B}^\prime\sqrt{\frac{2\pdim(\mathcal{F}^\prime_{jn})\log(en)}{n}}+2\kappa\mathcal{B}^\prime\sqrt{\frac{\log(1/\delta)}{2n}}
\end{gather*}
for $j=1,\ldots,d$ where  $\mathcal{F}^\prime_{jn}=\{\pj f: f\in\mathcal{F}_n\}$.
Combining above in probability bounds, we know that for any $\delta>0$, with probability at least $1-(d+1)\delta$,
\begin{align*}
	&\sup_{f\in\mathcal{F}_n}\left\vert [\mathcal{R}^\lambda(f)-\mathcal{R}^\lambda(f_0)]-\mathbb{E}[\mathcal{R}^\lambda_n(f)-\mathcal{R}^\lambda_n(f_0)\mid {S_X}]\right\vert\\
	&\le4\sqrt{\frac{2\log(en)}{n}}\Big[\mathcal{B}^2\sqrt{\pdim(\mathcal{F}_{n})}+
\bar{\lambda}\kappa\mathcal{B}^\prime\sqrt{\pdim(\mathcal{F}^\prime_{jn})}\Big]+4\Big[\mathcal{B}^2+\bar{\lambda}\kappa\mathcal{B}^\prime\Big]\sqrt{\frac{\log(1/\delta)}{2n}}.
\end{align*}

Recall that $\mathcal{F}_n=\mathcal{F}_{\mathcal{D},\mathcal{W},\mathcal{U},\mathcal{S},\mathcal{B},\mathcal{B}^\prime}$ is a class of RePU neural networks with depth $\mathcal{D}$, width $\mathcal{W}$, size $\mathcal{S}$ and number of neurons $\mathcal{U}$. By Lemma \ref{lemmapdim}, $\pdim(\mathcal{F}_n)\le 3p\mathcal{D}\mathcal{S}(\mathcal{D}+\log_2\mathcal{U})$ Then for each $j=1,\ldots,d$, the function class $\mathcal{F}_{jn}=\{\pj f: f\in\mathcal{F}_n\}$ consists of RePU neural networks with depth $3\mathcal{D}+3$, width $6\mathcal{W}$, number of neurons $13\mathcal{U}$ and size no more than $23\mathcal{S}$. By Theorem \ref{thm2} and Lemma \ref{lemmapdim}, we have $\pdim(\mathcal{F}^\prime_{1n})=\pdim(\mathcal{F}^\prime_{2n})=\ldots=\pdim(\mathcal{F}^\prime_{dn})\le 2484p\mathcal{D}\mathcal{S}(\mathcal{D}+\log_2\mathcal{U})$. Then, for any $\delta>0$, with probability at least $1-\delta$
\begin{align*}
	&\sup_{f\in\mathcal{F}_n}\left\vert [\mathcal{R}^\lambda(f)-\mathcal{R}^\lambda(f_0)]-\mathbb{E}[\mathcal{R}^\lambda_n(f)-\mathcal{R}^\lambda_n(f_0)\mid {S_X}]\right\vert\\
	&\le 200 d(\mathcal{B}^2+\bar{\lambda}\kappa\mathcal{B}^\prime)\Bigg(\sqrt{\frac{2p\log(en)\mathcal{D}\mathcal{S}(\mathcal{D}+\log_2\mathcal{U})}{n}}+\sqrt{\frac{\log((d+1)/\delta)}{2n}}\Bigg).
\end{align*}
If we let $t=200 d(\mathcal{B}^2+\kappa\bar{\lambda}\mathcal{B}^\prime)(\sqrt{{2p\log(en)\mathcal{D}\mathcal{S}(\mathcal{D}+\log_2\mathcal{U})}/{n}})$, then above inequality implies

\begin{align*}
	&\mathbb{P}\Bigg(\sup_{f\in\mathcal{F}_n}\left\vert [\mathcal{R}^\lambda(f)-\mathcal{R}^\lambda(f_0)]-\mathbb{E}[\mathcal{R}^\lambda_n(f)-\mathcal{R}^\lambda_n(f_0)\mid {S_X}]\right\vert\ge  \epsilon\Bigg)\\
	&
\le (d+1)\exp\left(\frac{-n(\epsilon-t)^2}{[100d(\mathcal{B}^2+\kappa\bar{\lambda}\mathcal{B}^\prime)]^2}\right),
\end{align*}
for $\epsilon\ge t$. And

\begin{align*}
	&\mathbb{E}\left[\sup_{f\in\mathcal{F}_n}\left\vert [\mathcal{R}^\lambda(f)-\mathcal{R}^\lambda(f_0)]-\mathbb{E}[\mathcal{R}^\lambda_n(f)-\mathcal{R}^\lambda_n(f_0)\mid {S_X}]\right\vert\right]\\
	&=\int_0^\infty \mathbb{P}\Bigg(\sup_{f\in\mathcal{F}_n}\left\vert [\mathcal{R}^\lambda(f)-\mathcal{R}^\lambda(f_0)]-\mathbb{E}[\mathcal{R}^\lambda_n(f)-\mathcal{R}^\lambda_n(f_0)\mid {S_X}]\right\vert\ge  u\Bigg) du\\
	&=\int_0^t \mathbb{P}\Bigg(\sup_{f\in\mathcal{F}_n}\left\vert [\mathcal{R}^\lambda(f)-\mathcal{R}^\lambda(f_0)]-\mathbb{E}[\mathcal{R}^\lambda_n(f)-\mathcal{R}^\lambda_n(f_0)\mid {S_X}]\right\vert\ge  u\Bigg) du\\
	&+\int_t^\infty \mathbb{P}\Bigg(\sup_{f\in\mathcal{F}_n}\left\vert [\mathcal{R}^\lambda(f)-\mathcal{R}^\lambda(f_0)]-\mathbb{E}[\mathcal{R}^\lambda_n(f)-\mathcal{R}^\lambda_n(f_0)\mid {S_X}]\right\vert\ge  u\Bigg) du\\
	&\le \int_0^t 1 du+\int_t^\infty (d+1)\exp\left(\frac{-n(\epsilon-t)^2}{[100d(\mathcal{B}^2+\kappa\bar{\lambda}\mathcal{B}^\prime)]^2}\right) du\\
	&=t+200\sqrt{2\pi}d(\mathcal{B}^2+\kappa\bar{\lambda}\mathcal{B}^\prime)/{\sqrt{n}}\\
	&\le 800d^2(\mathcal{B}^2+\kappa\bar{\lambda}\mathcal{B}^\prime)\sqrt{{2p\log(en)\mathcal{D}\mathcal{S}(\mathcal{D}+\log_2\mathcal{U})}/{n}}.
\end{align*}

Note that the network depth $\mathcal{D}$, number of neurons  $\mathcal{U}$ and number of parameters $\mathcal{S}$ satisfies $\mathcal{U}=18pd((\mathcal{D}+1)/2)^d$ and $\mathcal{S}=67pd((\mathcal{D}+1)/2)^d$. By Lemma \ref{lemma5}, combining the error decomposition, we have
\begin{align*}
	\mathbb{E}\Big[\mathcal{R}^\lambda(\hat{f}^\lambda_n)-\mathcal{R}^\lambda(f_0)\Big]&\le C_1p^2d^3(\mathcal{B}^2+\kappa\bar{\lambda}\mathcal{B}^\prime)(\log n)^{1/2} n^{-1/2}\mathcal{U}^{(d+2)/2d} \\
	&+ C_2(1+\kappa\bar{\lambda}) \Vert f_0\Vert^2_{C^s}\mathcal{U}^{-(s-1)/d}+\sum_{j=1}^d\lambda_j\mathbb{E}[\rho(\pj f_0(X))],
\end{align*}
where $C_1>0$ is a universal constant, $C_2>0$ is a constant depending only on $d,s$ and the diameter of the support $\mathcal{X}$. This completes the proof. $\hfill\Box$

}

\subsection*{Proof of Lemma \ref{stoerr}}

	Let $S=\{Z_i=(X_i,Y_i)\}_{i=1}^n$ be the sample used to estimate $\hat{f}^\lambda_n$ from the distribution $Z=(X,Y)$. And let $S^\prime=\{Z^\prime_i=(X^\prime_i,Y^\prime_i)\}_{i=1}^n$ be another sample independent of $S$. Define

	\begin{gather*}
		g_1(f,X_i)=\mathbb{E}\big\{\vert Y_i-f(X_i)\vert^2-\vert Y_i-f_0(X_i)\vert^2\mid X_i\big\}=\mathbb{E}\big\{\vert f(X_i)-f_0(X_i)\vert^2\mid X_i\big\}\\
		g_2(f,X_i)=\mathbb{E}\big[\frac{1}{d}\sum_{j=1}^d\lambda_j\rho(\pj f(X_i))-\frac{1}{d}\sum_{j=1}^d\lambda_j\rho(\pj f_0(X_i))\mid X_i\big]\\
		g(f,X_i)=g_1(f,X_i)+g_2(f,X_i)
	\end{gather*}
	for any (random) $f$ and sample $X_i$. It worth noting that for any $x$ and $f\in\mathcal{F}_n$,
	$$0\le g_1(f,x)=\mathbb{E}\big\{\vert f(X_i)-f_0(X_i)\vert^2\mid X_i=x\big\}\le 4\mathcal{B}^2,$$
	since $\Vert f \Vert_\infty\le \mathcal{B}$ and $\Vert f_0 \Vert_\infty\le \mathcal{B}$ for $f\in\mathcal{F}_n$ by assumption.  For any $x$ and $f\in\mathcal{F}_n$,
	$$0\le g_2(f,x)=\mathbb{E}\big[\frac{1}{d}\sum_{j=1}^d\lambda_j\rho(\pj f(X_i))-\frac{1}{d}\sum_{j=1}^d\lambda_j\rho(\pj f_0(X_i))\mid X_i=x\big]\le 2\mathcal{B}^\prime\kappa\bar{\lambda},$$
	since $\rho(\cdot)$ is a $\kappa$-Lipschitz function and $\Vert \pj f \Vert_\infty\le \mathcal{B}^\prime$ and $\Vert \pj f_0 \Vert_\infty\le \mathcal{B}^\prime$ for $j=1,\ldots,d$ for $f\in\in\mathcal{F}_n$ by assumption
	
	Recall that the the empirical risk minimizer $\hat{f}^\lambda_n$ depends on the sample $S$, and the stochastic error is
	\begin{align} \notag
		\mathbb{E}\{\mathcal{R}^\lambda(\hat{f}^\lambda_n)-2\mathcal{R}^\lambda_n(\hat{f}^\lambda_n)+\mathcal{R}^\lambda(f_0)\}&=\mathbb{E}_S\Big(\frac{1}{n}\sum_{i=1}^{n}\bigg[\mathbb{E}_{S^\prime}\big\{g(\hat{f}^\lambda_n,X^\prime_i)\big\}-2g(\hat{f}^\lambda_n,X_i)\bigg]\Big)\\ \label{sto_g1}
		&=\mathbb{E}_S\Big(\frac{1}{n}\sum_{i=1}^{n}\bigg[\mathbb{E}_{S^\prime}\big\{g_1(\hat{f}^\lambda_n,X^\prime_i)\big\}-2g_1(\hat{f}^\lambda_n,X_i)\bigg]\Big)\\ \label{sto_g2}
		&+\mathbb{E}_S\Big(\frac{1}{n}\sum_{i=1}^{n}\bigg[\mathbb{E}_{S^\prime}\big\{g_2(\hat{f}^\lambda_n,X^\prime_i)\big\}-2g_2(\hat{f}^\lambda_n,X_i)\bigg]\Big).
	\end{align}

In the following,  we derive upper bounds of (\ref{sto_g1}) and (\ref{sto_g2}) respectively. For any random variable $\xi$, it is clear that $\mathbb{E} [\xi]\le \mathbb{E} [\max\{\xi,0\}]=\int_{0}^\infty \mathbb{P}(\xi>t) dt$. In light of this, we aim at giving upper bounds for the tail probabilities
$$\mathbb{P}\left(\frac{1}{n}\sum_{i=1}^{n}\bigg[\mathbb{E}_{S^\prime}\big\{g_k(\hat{f}^\lambda_n,X^\prime_i)\big\}-2g_k(\hat{f}^\lambda_n,X_i)\bigg]>t \right),\qquad k=1,2$$
for $t>0$. Given $\hat{f}^\lambda_n\in\mathcal{F}_n$, for $k=1,2$,  we have
\begin{align}\notag
	&\mathbb{P}\left(\frac{1}{n}\sum_{i=1}^{n}\bigg[\mathbb{E}_{S^\prime}\big\{g_k(\hat{f}^\lambda_n,X^\prime_i)\big\}-2g_k(\hat{f}^\lambda_n,X_i,\xi_i)\bigg]>t \right)\\ \notag
	\le&\mathbb{P}\left(\exists f\in\mathcal{F}_n: \frac{1}{n}\sum_{i=1}^{n}\bigg[\mathbb{E}_{S^\prime}\big\{g_k( f,X^\prime_i)\big\}-2g_k(f,X_i)\bigg]>t \right)\\ \label{sto_1}
	=&\mathbb{P}\left(\exists f\in\mathcal{F}_n: \mathbb{E}\big\{g_k( f,X)\big\}-\frac{1}{n}\sum_{i=1}^{n}\big[g_k(f,X_i)\big]>\frac{1}{2}\bigg(t+\mathbb{E}\big\{g_k( f,X)\big\}\bigg)\right).
\end{align}
The bound the probability (\ref{sto_1}), we apply Lemma 24 in \cite{shen2022estimation}. For completeness of the proof, we present Lemma in the following.

\begin{lemma}[Lemma 24 in \cite{shen2022estimation}]\label{lemma24}
	 Let $\mathcal{H}$ be a set of functions $h:\mathbb{R}^d\to[0,B]$ with $B\ge1$. Let $Z,Z_1,\ldots,Z_n$ be i.i.d. $\mathbb{R}^d$-valued random variables. Then for each $n\ge1$ and any $s>0$ and $0<\epsilon<1$,
	
	\begin{align*}
		\mathbb{P}\left( \sup_{h\in\mathcal{H}}\ \frac{\mathbb{E}\big\{h(Z)\big\}-\frac{1}{n}\sum_{i=1}^{n}\big[h(Z_i)\big]}{s+\mathbb{E}\big\{h(Z)\big\}+\frac{1}{n}\sum_{i=1}^{n}\big[h(Z_i)\big]} >\epsilon\right)\le 4\mathcal{N}_{n}\Big(\frac{s\epsilon}{16},\mathcal{H},\Vert\cdot\Vert_\infty\Big)\exp\Big(-\frac{\epsilon^2sn}{15B}\Big),
	\end{align*}
	where $\mathcal{N}_{n}(\frac{s\epsilon}{16},\mathcal{H},\Vert\cdot\Vert_\infty)$ is the covering number of $\mathcal{H}$ with radius $s\epsilon/16$ under the norm $\Vert\cdot\Vert_\infty$. The definition of the covering number can be found in Appendix \ref{appendix_c}.
\end{lemma}

We apply Lemma \ref{lemma24} with $\epsilon=1/3,s=2t$ to the class of functions $\mathcal{G}_k:=\{g_k(f,\cdot):f\in\mathcal{F}_n\}$ for $k=1,2$ to get
\begin{align}\notag
&	\mathbb{P}\Big(\exists f\in\mathcal{F}_n: \mathbb{E}
\big\{g_1(f,X)\big\}-\frac{1}{n}
\sum_{i=1}^{n}\big[g_1(f,X_i)\big]>\frac{1}{2}\bigg(t+\mathbb{E}
\big\{g_1( f,X)\big\}\bigg)\Big)\\ \label{sto_2}
	&\le4\mathcal{N}_{n}\Big(\frac{t}{24},\mathcal{G}_1,\Vert\cdot\Vert_\infty\Big)\exp\Big(-\frac{tn}{270\mathcal{B}^2}\Big),
\end{align}
and
\begin{align}\notag
&	\mathbb{P}\Big(\exists f\in\mathcal{F}_n: \mathbb{E}\big\{g_2( f,X)\big\}-\frac{1}{n}\sum_{i=1}^{n}\big[g_2(f,X_i)\big]>\frac{1}{2}\bigg(t+\mathbb{E}\big\{g_2( f,X)\big\}\bigg)\Big)\\ \label{sto_3}
&\le 4\mathcal{N}_{n}\Big(\frac{t}{24},\mathcal{G}_2,\Vert\cdot\Vert_\infty\Big)\exp\Big(-\frac{tn}{135\bar{\lambda}\kappa\mathcal{B}^\prime}\Big).
\end{align}

Combining (\ref{sto_1}) and (\ref{sto_2}), for $a_n>1/n$,  we have
\begin{align*}
	&\mathbb{E}_S\Big(\frac{1}{n}\sum_{i=1}^{n}\bigg[\mathbb{E}_{S^\prime}\big\{g_1(\hat{f}^\lambda_n,X^\prime_i)\big\}-2g_1(\hat{f}^\lambda_n,X_i)\bigg]\Big)\\
&	\le  \int_0^\infty \mathbb{P}\Big( \frac{1}{n}\sum_{i=1}^{n}\bigg[\mathbb{E}_{S^\prime}\big\{g_1(\hat{f}^\lambda_n,X^\prime_i)\big\}-2g_1(\hat{f}^\lambda_n,X_i)\bigg]>t\Big) dt\\
&	\le  \int_0^{a_n} 1 dt+ \int_{a_n}^\infty4\mathcal{N}_{n}\Big(\frac{t}{24},\mathcal{G}_1,\Vert\cdot\Vert_\infty\Big)\exp\Big(-\frac{tn}{270\mathcal{B}^2}\Big) dt\\
&	\le a_n + 4\mathcal{N}_{n}\Big(\frac{1}{24n},\mathcal{G}_1,\Vert\cdot\Vert_\infty\Big) \int_{a_n}^\infty\exp\Big(-\frac{tn}{270\mathcal{B}^2}\Big) dt\\
&	= a_n+ 4\mathcal{N}_{n}\Big(\frac{1}{24n},\mathcal{G}_1,\Vert\cdot\Vert_\infty\Big)\exp\Big(-\frac{a_nn}{270\mathcal{B}^2}\Big)\frac{270\mathcal{B}^2}{n}.
\end{align*}
Choosing $a_n=\log \{4\mathcal{N}_n(1/(24n),\mathcal{G}_1,\Vert\cdot\Vert_\infty)\} \cdot 270\mathcal{B}^2/n$, we get
\begin{align*} \notag
	\mathbb{E}_S\Big(\frac{1}{n}\sum_{i=1}^{n}\bigg[\mathbb{E}_{S^\prime}\big\{g_1(\hat{f}^\lambda_n,X^\prime_i)\big\}-2g_1(\hat{f}^\lambda_n,X_i)\bigg]\Big)\le \frac{270\log [4e \mathcal{N}_n(1/(24n),\mathcal{G}_1,\Vert\cdot\Vert_\infty)] \mathcal{B}^2}{n}.
\end{align*}
For any $f_1,f_2\in\mathcal{F}_n$, by the definition of $g_1$, it is easy to show $\Vert g_1(f_1,\cdot)-g_1(f_2,\cdot)\Vert_\infty\le 4\mathcal{B}\Vert f_1-f_2\Vert_\infty$. Then $\mathcal{N}_n(1/(24n),\mathcal{G}_1,\Vert\cdot\Vert_\infty)\le \mathcal{N}_n(1/(96\mathcal{B}n),\mathcal{F}_n,\Vert\cdot\Vert_\infty)$, which leads to
\begin{align} \notag
	&\mathbb{E}_S\Big(\frac{1}{n}\sum_{i=1}^{n}\bigg[\mathbb{E}_{S^\prime}\big\{g_1(\hat{f}^\lambda_n,X^\prime_i,\xi^\prime_i)\big\}-2g_1(\hat{f}^\lambda_n,X_i,\xi_i)\bigg]\Big)\\ \label{sto_4}
	&
\le \frac{270\log [4e \mathcal{N}_n(1/(96\mathcal{B}n),\mathcal{F}_n,\Vert\cdot\Vert_\infty)] \mathcal{B}^2}{n}.
\end{align}
Similarly, combining (\ref{sto_1}) and (\ref{sto_3}), we can obtain
\begin{align*} \notag
	\mathbb{E}_S\Big(\frac{1}{n}\sum_{i=1}^{n}\bigg[\mathbb{E}_{S^\prime}\big\{g_2(\hat{f}^\lambda_n,X^\prime_i)\big\}-2g_2(\hat{f}^\lambda_n,X_i)\bigg]\Big)\le \frac{135\bar{\lambda}\kappa\log [4e \mathcal{N}_n(1/(24n),\mathcal{G}_2,\Vert\cdot\Vert_\infty)] \mathcal{B}^\prime}{n}.
\end{align*}

For any $f_1,f_2\in\mathcal{F}_n$, by the definition of $g_2$, it can be shown $\Vert g_2(f_1,\cdot)-g_2(f_2,\cdot)\Vert_\infty\le \frac{\kappa}{d}\sum_{j=1}^{d} \lambda_j\Vert \pj f_1-\pj f_2\Vert_\infty$. Recall that $\mathcal{F}_{nj}^\prime=\{\pj f:f\in\mathcal{F}_n\}$ for $j=1,\ldots,d$. Then $\mathcal{N}_n(1/(24n),\mathcal{G}_2,\Vert\cdot\Vert_\infty)\le \Pi_{j=1}^d \mathcal{N}_n(1/(24\kappa \lambda_jn),\mathcal{F}_{nj}^\prime,\Vert\cdot\Vert_\infty)$ where we view $11/(24\kappa \lambda_jn)$ as $\infty$ if $\lambda_j=0$. This leads to
\begin{align} \notag
	&\mathbb{E}_S\Big(\frac{1}{n}\sum_{i=1}^{n}\bigg[\mathbb{E}_{S^\prime}\big\{g_2(\hat{f}^\lambda_n,X^\prime_i,\xi^\prime_i)\big\}-2g_2(\hat{f}^\lambda_n,X_i,\xi_i)\bigg]\Big)\\ \label{sto_5}
	&
\le \frac{135\bar{\lambda}\kappa\log [4e \Pi_{j=1}^d \mathcal{N}_n(1/(24\kappa\lambda_jn),\mathcal{F}^\prime_{nj},\Vert\cdot\Vert_\infty)] \mathcal{B}^\prime}{n}.
\end{align}

Then by Lemma \ref{c2p} in Appendix \ref{appendix_c}, we can further bound the covering number by the Pseudo dimension. More exactly, for $n\ge \pdim(\mathcal{F}_n)$ and any $\delta>0$, we have
	\begin{align*}
		\log(\mathcal{N}_{n}(\delta,\mathcal{F}_n,\Vert\cdot\Vert_\infty))\le\pdim(\mathcal{F}_n)\log\Big(\frac{en\mathcal{B}}{\delta\pdim(\mathcal{F}_n)}\Big),
	\end{align*}
	and for $n\ge \pdim(\mathcal{F}^\prime_{nj})$ for $j=1\ldots,d$ and any $\delta>0$, we have
	\begin{align*}
		\log(\mathcal{N}_{n}(\delta,\mathcal{F}^\prime_{nj},\Vert\cdot\Vert_\infty))\le\pdim(\mathcal{F}^\prime_{nj})\log\Big(\frac{en\mathcal{B}^\prime}{\delta\pdim(\mathcal{F}^\prime_{nj})}\Big).
	\end{align*}
By Theorem \ref{thm2} we know $\pdim(\mathcal{F}^\prime_{nj})=\pdim(\mathcal{F}^\prime_{n})$ for $j=1,\ldots,d$. Combining the upper bounds of the covering numbers, we have
	\begin{align}\notag
		&\mathbb{E}\{\mathcal{R}^\lambda(\hat{f}^\lambda_n)-2\mathcal{R}^\lambda_n(\hat{f}^\lambda_n)+\mathcal{R}^\lambda(f_0)\}\le c_0\frac{\big[\mathcal{B}^3\pdim(\mathcal{F}_n)+d(\kappa\bar{\lambda}\mathcal{B}^\prime)^2\pdim(\mathcal{F}_n^\prime)\big]\log(n)}{n},
	\end{align}
	for $n\ge\max\{\pdim(\mathcal{F}_n),\pdim(\mathcal{F}^\prime_n)\}$ and some universal constant $c_0>0$ where $\bar{\lambda}=\sum_{j=1}^d\lambda_j/d$.
	By Lemma \ref{lemmapdim}, for the function class $\mathcal{F}_n$ implemented by Mixed RePU activated multilayer perceptrons with depth no more than $\mathcal{D}$, width no more than $\mathcal{W}$, number of neurons (nodes) no more than $\mathcal{U}$ and size or number of parameters (weights and bias) no more than $\mathcal{S}$, we have
	\begin{align*}
		\pdim(\mathcal{F}_n)\le3p\mathcal{D}\mathcal{S}(\mathcal{D}+\log_2\mathcal{U}),
	\end{align*}
	and by Lemma \ref{lemmapdim}, for any function $f\in\mathcal{F}_n$, its partial derivative $\pj f$ can be implemented by a Mixed RePU activated multilayer perceptron with depth $3\mathcal{D}+3$, width $6\mathcal{W}$, number of neurons $13\mathcal{U}$, number of parameters $23\mathcal{S}$ and bound $\mathcal{B}^\prime$. Then
	\begin{align*}
\pdim(\mathcal{F}^\prime_n)\le2484p\mathcal{D}\mathcal{S}(\mathcal{D}+\log_2\mathcal{U}).
	\end{align*}
	It follows that
	\begin{align}\notag &\mathbb{E}\{\mathcal{R}^\lambda(\hat{f}^\lambda_n)-2\mathcal{R}^\lambda_n(\hat{f}^\lambda_n)+\mathcal{R}^\lambda(f_0)\}\le c_1\big(\mathcal{B}^3+d(\kappa\bar{\lambda}\mathcal{B}^{\prime})^2\big)\frac{\mathcal{D}\mathcal{S}(\mathcal{D}+\log_2\mathcal{U})\log(n)}{n},
	\end{align}
	for $n\ge\max\{\pdim(\mathcal{F}_n),\pdim(\mathcal{F}^\prime_n)\}$ and some universal constant $c_1>0$. This completes the proof. $\hfill \Box$

\subsection*{Proof of Lemma \ref{lemma5}}
Recall that
\begin{align*}
	&\inf_{f\in\mathcal{F}_n}\Big[\mathcal{R}^\lambda(f)-\mathcal{R}^\lambda(f_0)\Big]\\
	&=\inf_{f\in\mathcal{F}_n}\Bigg[\mathbb{E} \vert f(X)-f_0(X)\vert^2+\frac{1}{d}\sum_{j=1}^{d} \lambda_j\{\rho(\pj f(X))-\rho(\pj f_0(X))\}\Bigg]\\
&	\le \inf_{f\in\mathcal{F}_n}\Bigg[\mathbb{E} \vert f(X)-f_0(X)\vert^2+\frac{1}{d}\sum_{j=1}^{d} \lambda_j\kappa\vert\pj f(X)-\pj f_0(X)\vert\Bigg].
\end{align*}
By Theorem \ref{approx2}, for each $N\in\mathbb{N}^+$, there exists a RePU network $\phi_N\in\mathcal{F}_n$ with $2N-1$ hidden layer, no more than $15N^{d}$ neurons, no more than $24N^{d}$ parameters and width no more than $12N^{d-1}$ such that for each multi-index $\alpha\in\mathbb{N}^d_0$ with $\vert\alpha\vert_1\le\min\{s,N\}$ we have
$$\sup_{\mathcal{X}}\vert D^\alpha (f-\phi_N)\vert\le C(s,d,\mathcal{X})\times N^{-(s-\vert\alpha\vert_1)}\Vert f\Vert_{C^{\vert\alpha\vert_1}},$$
where $C(s,d,\mathcal{X})$  is a positive constant depending only on $d,s$ and the diameter of $\mathcal{X}$. This implies
$$\sup_{\mathcal{X}}\vert  f-\phi_N\vert\le C(s,d,\mathcal{X})\times N^{-s}\Vert f\Vert_{C^0},$$
and for $j=1,\ldots,d$
$$\sup_{\mathcal{X}}\Big\vert \pj (f-\phi_N)\Big\vert\le C(s,d,\mathcal{X})\times N^{-(s-1)}\Vert f\Vert_{C^1}.$$
Combine above two uniform bounds, we have
\begin{align*}
	&\inf_{f\in\mathcal{F}_n}\Big[\mathcal{R}^\lambda(f)-\mathcal{R}^\lambda(f_0)\Big]\\
&	\le \Big[\vert \mathbb{E}_{X}\Big\{ \vert \phi_N(X)-f_0(X)\vert^2+\frac{\kappa}{d}\sum_{j=1}^d\lambda_j\vert \pj \phi_N(X)-\pj f_0(X)\vert\Big\}\Big]\\
&	\le C(s,d,\mathcal{X})^2\times N^{-2s}\Vert f\Vert^2_{C^0}+\kappa\bar{\lambda} C(s,d,\mathcal{X})\times N^{-(s-1)}\Vert f\Vert_{C^1}\\
&	\le C_1(s,d,\mathcal{X}) (1+\kappa\bar{\lambda}) N^{-(s-1)}\Vert f\Vert^2_{C^1},
\end{align*}
where $C_1(s,d,\mathcal{X})=\max\{[C(s,d,\mathcal{X})]^2,C(s,d,\mathcal{X}))\}$ is also a constant depending only on $s,d$ and $\mathcal{X}$. By defining the network depth $\mathcal{D}$ to be a positive odd number, and expressing the network width $\mathcal{W}$, neurons $\mathcal{U}$ and size $\mathcal{S}$ in terms of $\mathcal{D}$, one can obtain the approximation error bound in terms of $\mathcal{U}$. This completes the proof. $\hfill \Box$

\section{Definitions and Supporting Lemmas}\label{appendix_c}
\subsection{Definitions}

The following definitions are used in the proofs.

\begin{definition}[Covering number]
	Let $\mathcal{F}$ be a class of function from $\mathcal{X}$ to $\mathbb{R}$. For a given sequence $x=(x_1,\ldots,x_n)\in\mathcal{X}^n,$ let  $\mathcal{F}_n|_x=\{(f(x_1),\ldots,f(x_n):f\in\mathcal{F}_n\}$ be the  subset of $\mathbb{R}^{n}$. For a positive number $\delta$, let $\mathcal{N}(\delta,\mathcal{F}_n|_x,\Vert\cdot\Vert_\infty)$ be the covering number of $\mathcal{F}_n|_x$ under the norm $\Vert\cdot\Vert_\infty$ with radius $\delta$.
	Define the uniform covering number
	$\mathcal{N}_n(\delta,\Vert\cdot\Vert_\infty,\mathcal{F}_n)$ to be the maximum
	over all $x\in\mathcal{X}$ of the covering number $\mathcal{N}(\delta,\mathcal{F}_n|_x,\Vert\cdot\Vert_\infty)$, i.e.,
	\begin{equation}
		\label{ucover}
		\mathcal{N}_n(\delta,\mathcal{F}_n,\Vert\cdot\Vert_\infty)=
		\max\{\mathcal{N}(\delta,\mathcal{F}_n|_x,\Vert\cdot\Vert_\infty):x\in\mathcal{X}^n\}.
	\end{equation}
\end{definition}

\begin{definition}[Shattering]
	Let $\mathcal{F}$ be a family of functions from a set $\mathcal{Z}$ to $\mathbb{R}$. A set $\{z_1,\ldots,Z_n\}\subset\mathcal{Z}$ is said to be shattered by $\mathcal{F}$, if there exists $t_1,\ldots,t_n\in\mathbb{R}$ such that
	\begin{align*}
		\Big\vert\Big\{\Big[
		\begin{array}{lr}
			{\rm sgn}(f(z_1)-t_1)\\
			\ldots\\
			{\rm sgn}(f(z_n)-t_n)\\
		\end{array}\Big]:f\in\mathcal{F}
		\Big\}\Big\vert=2^n,
	\end{align*}
	where ${rm sgn}$ is the sign function returns $+1$ or $-1$ and $\vert\cdot\vert$ denotes the cardinality of a set. When they exist, the threshold values $t_1,\ldots,t_n$ are said to witness the shattering.
\end{definition}

\begin{definition}[Pseudo dimension]
	Let $\mathcal{F}$ be a family of functions mapping from $\mathcal{Z}$ to $\mathbb{R}$. Then, the pseudo dimension of $\mathcal{F}$, denoted by $\pdim(\mathcal{F})$, is the size of the largest set shattered by $\mathcal{F}$.
\end{definition}

\begin{definition}[VC dimension]
	Let $\mathcal{F}$ be a family of functions mapping from $\mathcal{Z}$ to $\mathbb{R}$. Then, the Vapnik–Chervonenkis (VC) dimension of $\mathcal{F}$, denoted by $\vdim(\mathcal{F})$, is the size of the largest set shattered by $\mathcal{F}$ with all threshold values being zero, i.e., $t_1=\ldots,=t_n=0$.
\end{definition}

\subsection{Supporting Lemmas}

{\color{black}
	
	\begin{lemma}[Stochastic error bound]\label{stoerr}
		Suppose Assumption \ref{assump3} and \ref{assump4} hold.
		Let $\mathcal{F}_n=\mathcal{F}_{\mathcal{D},\mathcal{W},\mathcal{U},\mathcal{S},\mathcal{B},\mathcal{B}^\prime}$ be the RePU $\sigma_{p}$ activated multilayer perceptron and let $\mathcal{F}^\prime_n=
		\{\frac{\partial}{\partial x_1} f:f\in\mathcal{F}_{n}\}$ denote the class of the partial derivative of $f\in\mathcal{F}_n$ with respect to its first argument. Then for $n\ge\max\{\pdim(\mathcal{F}_n),\pdim(\mathcal{F}^\prime_n)\}$, the stochastic error satisfies
		\begin{align*} \mathbb{E}\{\mathcal{R}^\lambda(\hat{f}^\lambda_n)-2\mathcal{R}^\lambda_n(\hat{f}^\lambda_n)
			+\mathcal{R}^\lambda(f_0)\}
			\le c_1p\big\{\mathcal{B}^3+d(\kappa\bar{\lambda}\mathcal{B}^\prime)^2\big\}
			\mathcal{D}\mathcal{S}(\mathcal{D}+\log_2\mathcal{U})\frac{\log(n)}{n},
		\end{align*}
		
		for some universal constant $c_1>0,$ where $\bar{\lambda}:=\sum_{j=1}^d\lambda_j/d$.
	\end{lemma}

	\begin{lemma} [Approximation error bound]
		\label{lemma5}
		Suppose that the target function $f_0$ defined in (\ref{reg0}) belongs to $C^s$ for some $s\in\mathbb{N}^+$. For any positive odd number  $\mathcal{D}$, let $\mathcal{F}_n:=\mathcal{F}_{\mathcal{D},\mathcal{W},\mathcal{U},\mathcal{S},\mathcal{B},\mathcal{B}^\prime}$ be the class of RePU activated neural networks $f:\mathcal{X}\to\mathbb{R}^d$ with depth $\mathcal{D}$, width $\mathcal{W}=18pd[(\mathcal{D}+1)/2]^{d-1}$, number of neurons $\mathcal{U}=18pd[(\mathcal{D}+1)/2]^{d}$ and size $\mathcal{S}=67pd[(\mathcal{D}+1)/2]^{d}$, satisfying $\mathcal{B}\ge\Vert f_0\Vert_{C^0}$ and $\mathcal{B}^\prime\ge\Vert f_0\Vert_{C^1}$. Then the approximation error given in Lemma \ref{decom} satisfies
		$$\inf_{f\in\mathcal{F}_n}\Big[\mathcal{R}^\lambda(f)-\mathcal{R}^\lambda(f_0)\Big]\le
		C(1+\kappa\bar{\lambda}) \mathcal{U}^{-(s-1)/d}\Vert f_0\Vert^2_{C^1},$$
		where $\bar{\lambda}:=\sum_{j=1}^d\lambda_j/d$, $\kappa$ is the Lipschitz constant of the panelty function $\rho$ and $C>0$ is a constant depending only on $d,s$ and the diameter of the support $\mathcal{X}$.
	\end{lemma}
	
}

The following lemma gives an upper bound for the covering number in terms of the pseudo-dimension.

\begin{lemma}[Theorem 12.2 in \citet{anthony1999}]\label{c2p}
	Let $\mathcal{F}$ be a set of real functions from domain $\mathcal{Z}$ to the bounded interval $[0,B]$. Let $\delta>0$ and suppose that $\mathcal{F}$ has finite pseudo-dimension $\pdim(\mathcal{F})$ then
	\begin{align*}
		\mathcal{N}_n(\delta,\mathcal{F},\Vert\cdot\Vert_\infty)\le\sum_{i=1}^{\pdim(\mathcal{F})}\binom{n}{i}\Big(\frac{B}{\delta}\Big)^i,
	\end{align*}
	which is less than $\{enB/(\delta\pdim(\mathcal{F}))\}^{\pdim(\mathcal{F})}$ for $n\ge\pdim(\mathcal{F})$.
\end{lemma}

The following lemma presents basic approximation properties of RePU network on monomials.

\begin{lemma}[Lemma 1 in \cite{li2019powernet}]\label{repu_lemma}
			The monomials $x^N,0\le N\le p$ can be exactly represented by RePU ($\sigma_{p},p\ge2$) activated neural network with one hidden layer and no more than $2p$ nodes. More exactly,
			\begin{itemize}
				\item [(i)] If $N=0$, the monomial $x^N$ can be computed by a RePU $\sigma_p$ activated network with one hidden layer and 1 nodes as
				$$1=x^0=\sigma_{p}(0\cdot x+1).$$
				\item [(ii)] If $N=p$, the monomial $x^N$ can be computed by a RePU $\sigma_p$ activated network with one hidden layer and 2 nodes as
				$$x^N=W_1\sigma_{p}(W_0x),\qquad W_1=\left[\begin{array}{c}1\\(-1)^p \end{array}\right], W_0=\left[\begin{array}{c}1\\-1 \end{array}\right].$$
				\item [(iii)] If $1\le N\le p$, the monomial $x^N$ can be computed by a RePU $\sigma_p$ activated network with one hidden layer and no more than $2p$ nodes. More generally, a polynomial of degree no more than $p$, i.e. $\sum_{k=0}^{p}a_k x^k$, can also be computed by a RePU $\sigma_p$ activated network with one hidden layer and no more than $2p$ nodes as

				$$x^N=W_1^\top\sigma_{p}(W_0^\top x+b_0)+u_0,$$
				where
				$$W_0=\left[\begin{array}{c} 1\\-1\\ \vdots\\ 1\\-1\end{array}\right]\in\mathbb{R}^{2p\times1},\ \  b_0=\left[\begin{array}{c}t_1\\-t_1\\ \vdots\\ t_p\\-t_p\end{array}\right]\in\mathbb{R}^{2p\times1},\ \
 W_1=\left[\begin{array}{c}u_1\\(-1)^pu_1\\ \vdots\\ u_p\\(-1)^pu_p\end{array}\right]\in\mathbb{R}^{2p\times1}.$$
				Here $t_1,\ldots,t_p$ are distinct values in $\mathbb{R}$ and values of $u_0,\ldots,u_p$ satisfy
				the linear system
				$$\left[\begin{array}{ccccc}
					1 & 1 &\cdots& 1 & 0\\
					\vdots &\vdots & & \vdots& \vdots\\
					t_1^{p-i} & t_2^{p-i}& \cdots & t_p^{p-i}& 0\\
					\vdots &\vdots & & \vdots& \vdots\\
					t_1^{p-1} & t_2^{p-1}& \cdots & t_p^{p-1}& 0\\
					t_1^{p} & t_2^{p}& \cdots & t_p^{p}& 1
				\end{array}\right]\left[\begin{array}{c}u_1\\\vdots\\u_i\\\vdots\\u_p\\u_0\end{array}\right]=\left[\begin{array}{c}a_p(C^p_p)^{-1}\\\vdots\\a_i(C^i_p)^{-1}\\\vdots\\a_1(C^1_p)^{-1}\\a_0(C^0_p)^{-1}\end{array}\right],$$
				where $C^i_p,i=0,\ldots,p$ are binomial coefficients. Note that the top-left $p\times p$ sub-matrix of the $(p+1)\times(p+1)$ matrix above is a Vandermonde matrix, which is invertible as long as $t_1,\ldots,t_p$ are distinct.
			\end{itemize}
		\end{lemma}

\bibliography{dir.bib}
\end{document}